\documentclass{bmvc2k}

\usepackage[normalem]{ulem}
\usepackage[ruled,vlined]{algorithm2e}
\SetKw{KwBy}{by}

\makeatletter
\newcommand\notsotiny{\@setfontsize\notsotiny\@vipt\@viipt}
\makeatother

\usepackage{color, colortbl}
\definecolor{myyellow}{rgb}{0.87,0.72,0.53}
\definecolor{myred}{rgb}{0.96,0.76,0.76}
\definecolor{mygray}{gray}{0.32}

\title{Towards Fast and Light-Weight Restoration of Dark Images}

\addauthor{Mohit Lamba*}{ee18d009@smail.iitm.ac.in}{1}
\addauthor{Atul Balaji}{ee16b002@smail.iitm.ac.in}{1}
\addauthor{Kaushik Mitra}{kmitra@ee.iitm.ac.in}{1}

\addinstitution{
 Computational Imaging Lab \\
 Dept. of Electrical Engineering\\
 Indian Institute of Technology Madras\\
 Chennai, India
}

\runninghead{Lamba, Balaji, Mitra}{Extremely Fast Low-Light Restoration}

\def\etal{\emph{et al}\bmvaOneDot}

\usepackage{multirow}

\begin{document}

\maketitle

\begin{abstract}
The ability to capture good quality images in the dark and \textit{near-zero lux} conditions has been a long-standing pursuit of the computer vision community. The seminal work by Chen \etal \cite{chen2018learning} has especially caused renewed interest in this area, resulting in methods that build on top of their work in a bid to improve the reconstruction. However, for practical utility and deployment of low-light enhancement algorithms on edge devices such as embedded systems, surveillance cameras, autonomous robots and smartphones, the solution must respect additional constraints such as limited GPU memory and processing power. With this in mind, we propose a deep neural network architecture that aims to strike a balance between the network latency, memory utilization, model parameters, and reconstruction quality. 
The key idea is to forbid computations in the High-Resolution (HR) space and limit them to a Low-Resolution (LR) space. However, doing the bulk of computations in the LR space causes artifacts in the restored image. 
We thus propose \textit{Pack} and \textit{UnPack} operations, which allow us to effectively transit between the HR and LR spaces without incurring much artifacts in the restored image.
State-of-the-art algorithms on dark image enhancement need to pre-amplify the image before processing it. However, they generally use ground truth information to find the amplification factor even during inference, restricting their applicability for unknown scenes. In contrast, we propose a simple yet effective light-weight mechanism for automatically determining the amplification factor from the input image.
We show that we can enhance a full resolution, $2848 \times 4256$, extremely dark single-image in the ballpark of $3$ seconds even on a CPU. We achieve this with $2-7\times$ fewer model parameters, $2-3\times$ lower memory utilization, $5-20\times$ speed up and yet maintain a competitive image reconstruction quality compared to the state-of-the-art algorithms.
\end{abstract}

\section{Introduction}
\label{sec:introduction}
The ability to swiftly capture high quality images with modest computations has led to the widespread proliferation of digital images. These advantages are, however, limited to good lighting conditions. Achieving similar results under low light is still a significant challenge. While much of the work in this direction has focused on enhancing weakly illuminated images \cite{automaticTenHElegacy,automaticNineHElegacy,automaticEllevenHElegacy,TIP2018structurePreserving,automaticThree2011naturalness,LOL,2016lime}, enhancement of extremely dark images has received comparatively lesser attention. 

Recently, however, a landmark paper by Chen \etal \cite{chen2018learning} has shown that it is possible to restore extremely dark images captured under near-zero lux conditions.
Following this work, several modifications have been proposed in a bid to improve the reconstruction quality. This includes the incorporation of attention units \cite{unetmodi1}, recurrent units \cite{unetmodi}, the adoption of a multi-scale approach \cite{gu2019self, iisc2019lowlight} and the usage of deeper networks \cite{ICME19}.
With these added complexities, these methods are constrained to run on desktop GPUs such as NVIDIA RTX 2080Ti with 12GB storage. But, real-world applications require image enhancement algorithms to run on embedded systems and edge devices with limited CPU RAM or minimal GPU capacity. 
One possible solution is to process the images in VGA resolution \cite{TIP2018structurePreserving,2016lime,LOL,vga2019,2017llnet,automaticFour2013naturalness}. But, this is in contrast to the current trend of capturing and processing high-definition images. 
Consequently, we aim to design a deep network that can restore an extreme low-light high-definition single-image with minimal CPU latency and low memory footprint, but at the same time has a competitive image restoration quality. 
\begin{figure}[t!]
	\centering
 	\scriptsize	
    \begin{tabular}{cc}
    \includegraphics[width=0.35\linewidth]{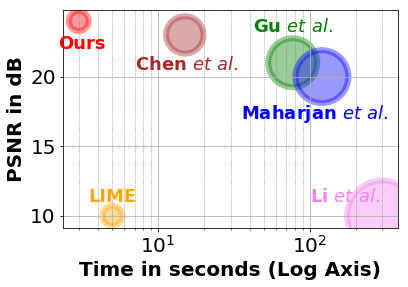} & \includegraphics[width=0.35\linewidth]{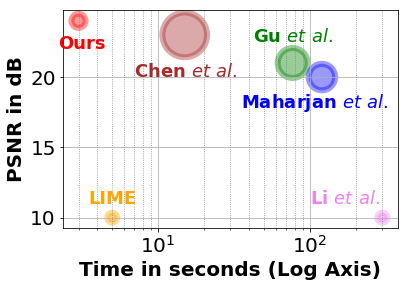} \vspace{-0.18cm}\\
    $\quad ~{}~{} ~{} Marker~{}Size$ $\propto$ $Memory$ $Utilization$ & $\quad ~{}~{} Marker~{}Size$ $\propto$ $Model$ $Parameters$
    \end{tabular}
    \caption{Performance comparison of the proposed method with state-of-the-art methods Chen \etal \cite{chen2018learning}, Gu \etal \cite{gu2019self}, Maharjan \etal \cite{ICME19} and traditional methods LIME \cite{2016lime}, and Li \etal \cite{TIP2018structurePreserving} for extreme low-light single-image enhancement. Refer to Table \ref{tab:main_comparison} for more details.}
\label{fig:abstract}
\end{figure}

We propose a deep neural network, called Low-Light Packing Network (LLPackNet), which is faster and computationally cheaper than the existing solutions. 
Recognizing the fact that a neural network's complexity increases quadratically with spatial dimensions \cite{CVPR2017factorized}, we perform the bulk of computations in a much lower resolution by performing aggressive down/up sampling operation. This is in contrast with much of the existing literature that down/up sample the feature maps in gradations  \cite{RDN2018,CVPR2017EDSR,chen2018learning,2019cvprlfresidual,rethinking_inception_2016}, which increases network latency and memory utilization.
For performing large downsampling operations, popular choices such as max-pooling and strided convolution \cite{2016DeepLearningGuide} cannot be used as they would cause much loss in information. We therefore, propose \textit{Pack} $\alpha \times$ downsampling operation, which rearranges the pixels in such a manner that it reduces the spatial dimension by a factor of $\alpha$, while increasing the number of channels by a factor of $\alpha^2$, see Fig. \ref{fig:architecture}. We show that the Pack operation bestows LLPackNet with an enormous receptive field which is not trivially possible by directly operating in the HR space. We also propose \textit{UnPack} $\alpha \times$ operation, which complements the \textit{Pack} $\alpha \times$ operation to do large upsampling. This operation is much faster than the usual transposed convolution layer \cite{2016DeepLearningGuide} and has no learnable parameters. For upsampling, PixelShuffle \cite{pixelshuffle} is another viable option but it lacks proper correlation between the color channels and hence results in heavy color cast in the restored image as shown in Fig. \ref{fig:packing_visual_comparison}. Altogether, the proposed \textit{Pack} and \textit{UnPack} operations allow us to operate in a much lower resolution space for computational advantages, without significantly affecting the restoration quality. See Fig. \ref{fig:abstract} for a qualitative comparison with state-of-the-art algorithms.


State-of-the-art deep learning solutions on extreme low-light image enhancement need to pre-amplify dark images before processing them \cite{chen2018learning,ICME19,unetmodi1,gu2019self}. However, these methods use ground-truth knowledge for predicting the amplification factor. In a real-world setting, because of lack of ground-truth (GT) knowledge, the amplification factor cannot be estimated properly and hence this would lead to degradation in performance. We therefore, equip the proposed LLPackNet with an amplifier module, which will estimate the amplification factor directly from the input image histogram. 

To summarize, the main contributions of this paper are as follows --- 1) We propose a deep neural network architecture, called \textit{LLPackNet}, that enhances an extremely dark single-image at high resolution even on a CPU with very low latency and computational resources. 2) We propose \textit{Pack} and \textit{UnPack} operations for better color restoration. 3) LLPackNet can estimate the amplification factor directly from the input image, without relying on ground-truth information, making it practical for real world applications. 4) Our experiments show that compared to existing solutions, we are able to restore high definition, extreme low-light RAW images with 2--7$\times$ fewer model parameters, 2--3$\times$ lower memory and 5--20$\times$ speed up, with a competitive restoration quality. 
Our code is available at \url{https://github.com/MohitLamba94/LLPackNet}.

\vspace{-0.3cm}
\section{Related Work}
\label{sec:Relatedwork}
Low-light enhancement methods are chiefly comprised of histogram equalization \cite{automaticTenHElegacy,automaticNineHElegacy,automaticEllevenHElegacy}, Retinex based decomposition \cite{2016lime,TIP2018structurePreserving, automaticTwo2017,SIMULATE2017,automaticThree2011naturalness,vga2019fast} and Deep learning based methods \cite{automaticSix2020unsupervised,2019simulateAdobe,SIMULATE2019,naturalnessillposed,2017llnet, wang2018gladnet,LOL, shen2017msr,automaticFiveSIMULATE2018,icassp2020}.
Most of them however, do not target extreme low-light conditions or high resolution images.
More recently, Chen \etal \cite{chen2018learning} proposed an end-to-end pipeline to restore extreme low-light high-definition RAW images, which has spurred several other works in this direction \cite{ICME19,unetmodi1,gu2019self, iisc2019lowlight, unetmodi,iccv2019nofearofdark}. Most of these methods, however, have significantly large processing time and memory utilization. As noted in Sec. \ref{sec:introduction} many of them also require GT information for image pre-amplification. However, other image amplification techniques that involve the use of CRF \cite{automaticOneCRF2018,automaticEightCRF}, image histogram \cite{automaticTenHElegacy,automaticNineHElegacy} or other assumptions \cite{TIP2020biological,2016lime} have been used in traditional image enhancement methods to estimate amplification, using only the input image. Borrowing from these ideas, we develop an amplifier module that uses the histogram of the input dark image to predict the amplification factor automatically, without relying on GT information. To the best of our knowledge, this has not been attempted before for deep learning based dark image enhancement.

Fast and efficient CNN models have been explored in other areas, especially image classification, but is mostly achieved by either approximating \cite{eccv2016xnor, CVPR2016quantized} or pruning the learned weights \cite{pruning}. In contrast, we propose a network that is inherently fast and efficient without using such weight-approximation or pruning approaches.


\vspace{-0.5cm}
\section{Low-Light Packing Network (LLPackNet)}

We propose Low-Light Packing Network (LLPackNet) for enhancing extremely dark high resolution single-images with low time--memory complexity.
We first describe the network architecture, shown Fig. \ref{fig:architecture}, in Sec. \ref{sec:archi} and then analyze the important components of our network, the Pack and UnPack operations, in Sec. \ref{sec:unpack}.





 

\vspace{-0.3cm}
\subsection{Network architecture}
\label{sec:archi}

\begin{figure}[t!]
        \notsotiny
    \centering
    \setlength\tabcolsep{0.5pt}
    \bgroup
\def\arraystretch{1}
\begin{tabular}{cc}
    \includegraphics[width=0.8\linewidth]{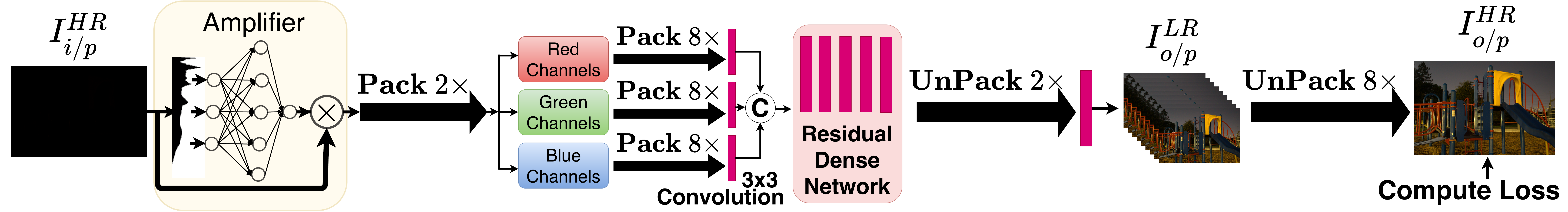} &
    \includegraphics[width=0.17\linewidth]{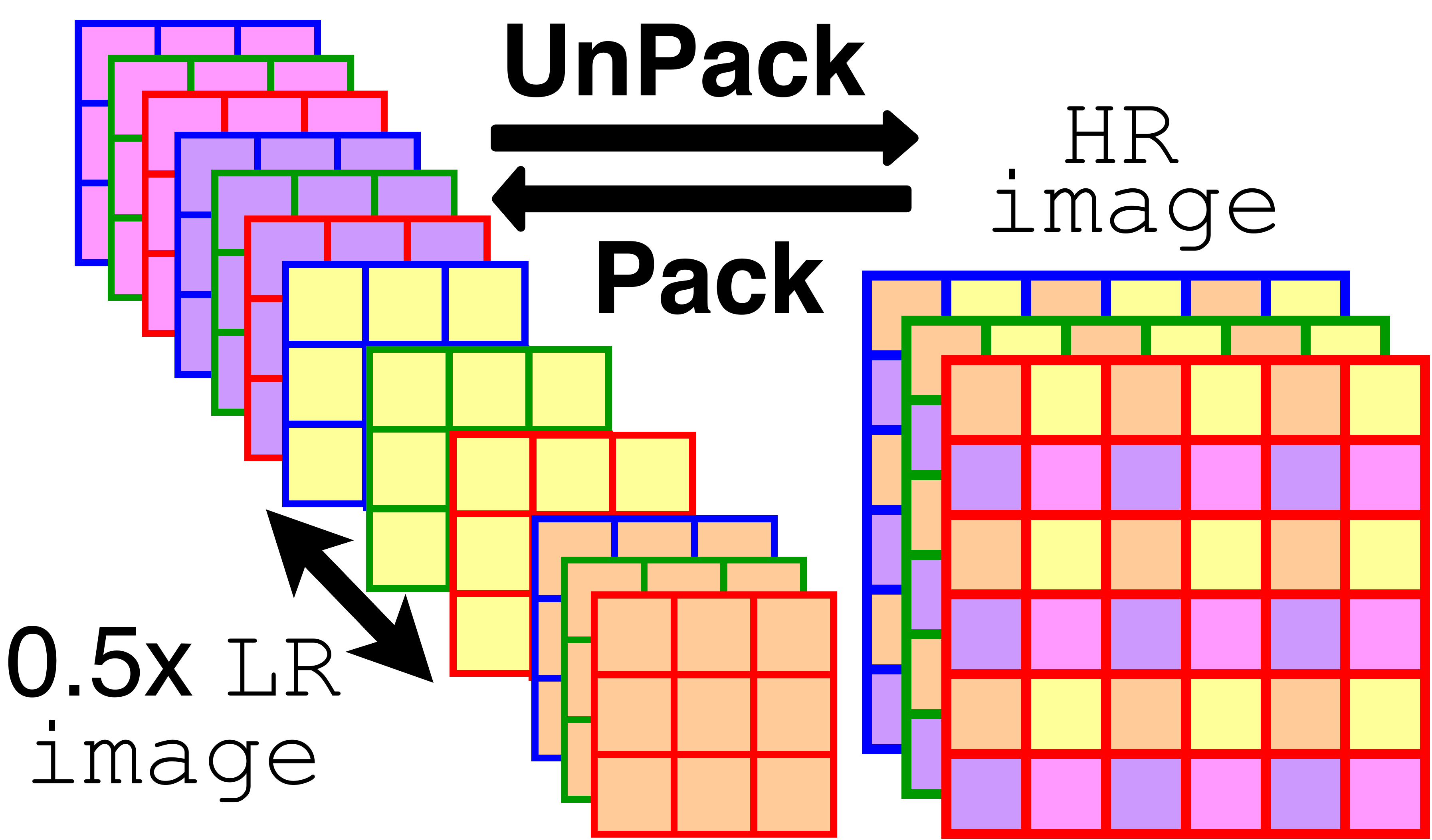} \\
    \textbf{(a) LLPackNet Architecture} & \textbf{(b) Pack/UnPack Operations}
    \end{tabular}
    \egroup
    \caption{Our proposed network, LLPackNet, learns the required amplification factor directly from the dark input image and uses the novel \textit{Pack} and \textit{UnPack} operations to perform aggressive down/up sampling with minimal color distortions. The network has low latency and low memory footprint.}
    \label{fig:architecture}
\end{figure}


\quad~ \textbf{Image amplification:} In general, dark images need to be pre-amplified before enhancing them.
We estimate the amplification factor using the incoming RAW image $I^{HR}_{i/p}$  by constructing a 64 bin histogram, with the histogram bins being equidistant in the log domain. This provides a finer binning resolution for lower intensities and a coarser binning resolution for higher intensities. 
The histogram is used by a multilayer perceptron, having just one hidden layer, to estimate the amplification factor.

\textbf{Fast and light-weight enhancement:} As discussed in Sec \ref{sec:introduction}, we want to perform most of the processing in LR space. Hence, our first step is to downsample the input image, without losing any information. For this purpose, we propose \textit{Pack} $\alpha \times$ operation, that downsamples the image by a factor of $\alpha$ along the spatial dimensions while increasing the number of channels by a factor of $\alpha^2$. This is shown in Fig. \ref{fig:architecture} (b) for $\alpha=2$. A pseudo code is also provided in Algorithm 1. Our goal is to perform 16$\times$ downsampling, which we do in two stages. In the first stage, the Pack 2$\times$ operation separates out the red, green and blue color components lying in the 2$\times$2 Bayer pattern \cite{2005bayerpattern} of the amplified image $I_{i/p}^{HR}$. This reduces the spatial dimension by half and increases the channels from 1 to 4 ($2^2$). Once the colors are separated into these channels, a subsequent Pack 8$\times$ operation is applied individually on each color channel, further reducing the spatial dimension from $2\times$ to $16\times$ lower resolution but increasing the number of channels from 1 to 64 ($8^2$). Now, using a 3$\times$3 convolution kernel, the channel dimension of each color component is reduced such that on concatenation, the resulting feature map has only 60 channels. The channel reduction at this stage is essential to prevent parameter and memory explosion in the downstream operations. 
This downsampled representation is then processed by a series of convolution operations. For this purpose, we use the Residual Dense Network \cite{RDN2018} (RDN) --- which consists of 3 residual dense blocks each with 6 convolutional layers and a growth rate of 32. RDN does not perform any down/up sampling operation or cause any change in channel dimension in its output. 
The output of the RDN now needs to be upsampled and for this we use the proposed \textit{UnPack} $2 \times$ operation, which is the inverse of \textit{Pack} $2 \times$. \textit{UnPack} $2 \times$, however, reduces the number of channels from 60 to 15 ($60/2^2$) and this needs to be increased to 192 ($8^2\times3$) to allow the final $8\times$ upsampling using UnPack $8\times$.
For this we use another set of $3\times3$ convolutions. Except for this operation, all the computations are done in the $16 \times$ lower resolution. We finally perform UnPack $8\times$ operation to get the restored image, $I^{HR}_{o/p}$. 

\textbf{Loss function: }Similar to Ignatov \etal \cite{ICCV2017dslr}, we compute the color loss, content loss and total variation (TV) loss on the restored image to train the network. Specifically, we use,
\begin{eqnarray}
Loss & = & \lambda_1 \cdot ||GT - I^{HR}_{o/p}||_1 + \lambda_2 \cdot ||\Psi(GT) - \Psi(I^{HR}_{o/p})||_1 + \lambda_3 \cdot ||\delta(GT) - \delta(I^{HR}_{o/p})||_1 \nonumber \\
&& + \lambda_4 \cdot  TV(I^{HR}_{o/p}) + \lambda_5 \cdot ||w||_1
\label{eq:loss}
\end{eqnarray}
where $\Psi$ is a feature map of VGG-19, $\delta$ performs Gaussian smoothing and $w$ denotes the network weights. VGG-19 features are obtained right after the final 3 max-pool layers.

\subsection{Pack/UnPack operation for better color restoration}
\label{sec:unpack}
The last section discussed LLPackNet from the vantage point of network complexity. In this section we analyze the network from the standpoint of reconstruction quality.

\begin{algorithm}[H]
\SetAlgoLined
\textbf{\textit{Pack}} $\alpha\times$ \textbf{\textit{operation}} \\ \qquad \textbf{Input: }An RGB image $I_{HR}$ of dimensions $H \times W \times 3$.\\
\qquad \textbf{Output: }$I_{LR}$ of dimension $\frac{H}{\alpha} \times \frac{W}{\alpha} \times 3\alpha^2$.\\

{\color{mygray}
$count = 0$

\textbf{for} $row$ \textbf{in} \textbf{range}($\alpha$) :\\
\qquad \textbf{for} $col$ \textbf{in} \textbf{range}($\alpha$) :\\
\qquad \qquad $I_{LR}[:,~:,~count:count+3]~=~I_{HR}[row:H:\alpha,~col:W:\alpha,~:]$\\
\qquad \qquad $count~=~count+3$} ~\\~\\
\textbf{\textit{UnPack}} $\alpha\times$ \textbf{\textit{operation}} \\ \qquad \textbf{Input: }$I_{LR}$ of dimension $\frac{H}{\alpha} \times \frac{W}{\alpha} \times 3\alpha^2$.\\
\qquad \textbf{Output: }An RGB image $I_{HR}$ of dimensions $H \times W \times 3$.\\

{\color{mygray}

$count = 0$

\textbf{for} $row$ \textbf{in} \textbf{range}($\alpha$) :\\
\qquad \textbf{for} $col$ \textbf{in} \textbf{range}($\alpha$) :\\
\qquad \qquad $I_{HR}[row:H:\alpha,~col:W:\alpha,~:]~=~I_{LR}[:,~:,~count:count+3]$\\
\qquad \qquad $count~=~count+3$}
 
 \caption{Python code for performing Pack $\alpha \times$ and UnPack $\alpha \times$ operation.}
\end{algorithm}

\vspace{0.1cm}
\textbf{Improving color correlation with UnPack $\alpha \times$: }
Making abrupt transitions between LR and HR spaces introduces several distortions in the restored image. To minimize these, we propose the novel Pack $\alpha \times$ and UnPack $\alpha \times$ operations. To understand these operations, it is crucial to analyze PixelShuffle \cite{pixelshuffle} - a fast and effective upsampling method, based on which they are formulated.  Using an analysis similar to \cite{2017checkerboardfreePixelshuffle,pixelshuffleequivalence2016,pixelshuffle}, we will show that Pack/UnPack operations lead to better color correlation than PixelShuffle.

\begin{figure}[t!]
    \notsotiny
    \centering
    \setlength\tabcolsep{-1.5
    pt}
    \bgroup
\def\arraystretch{1.05}
\begin{tabular}{ccc}
 \includegraphics[width = 0.32\linewidth]{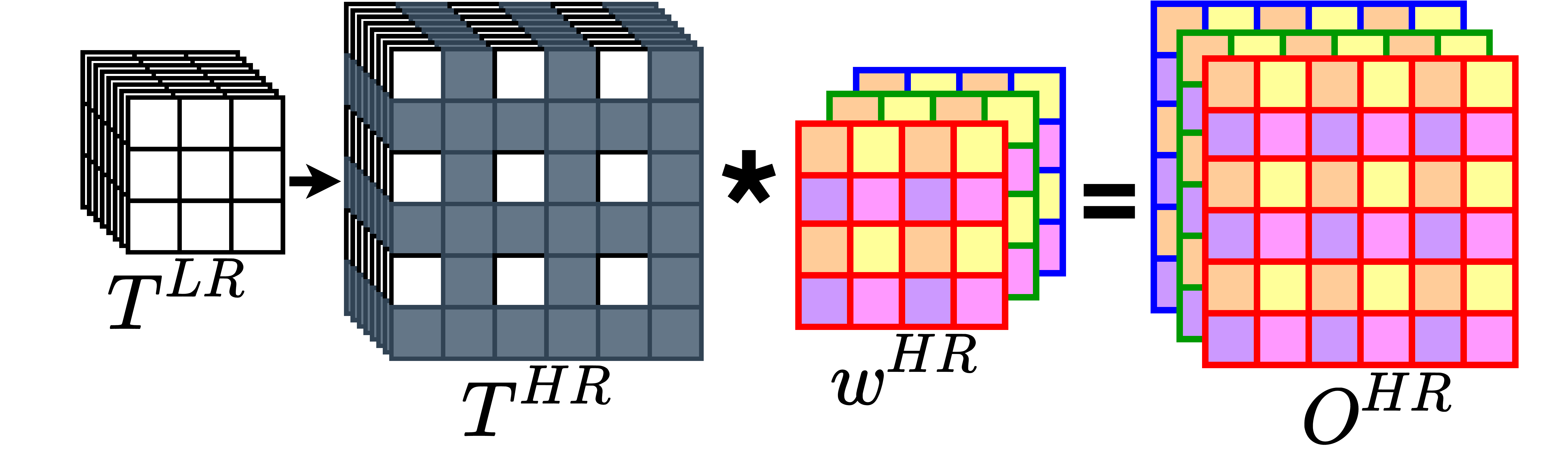} &  \includegraphics[width = 0.30\linewidth]{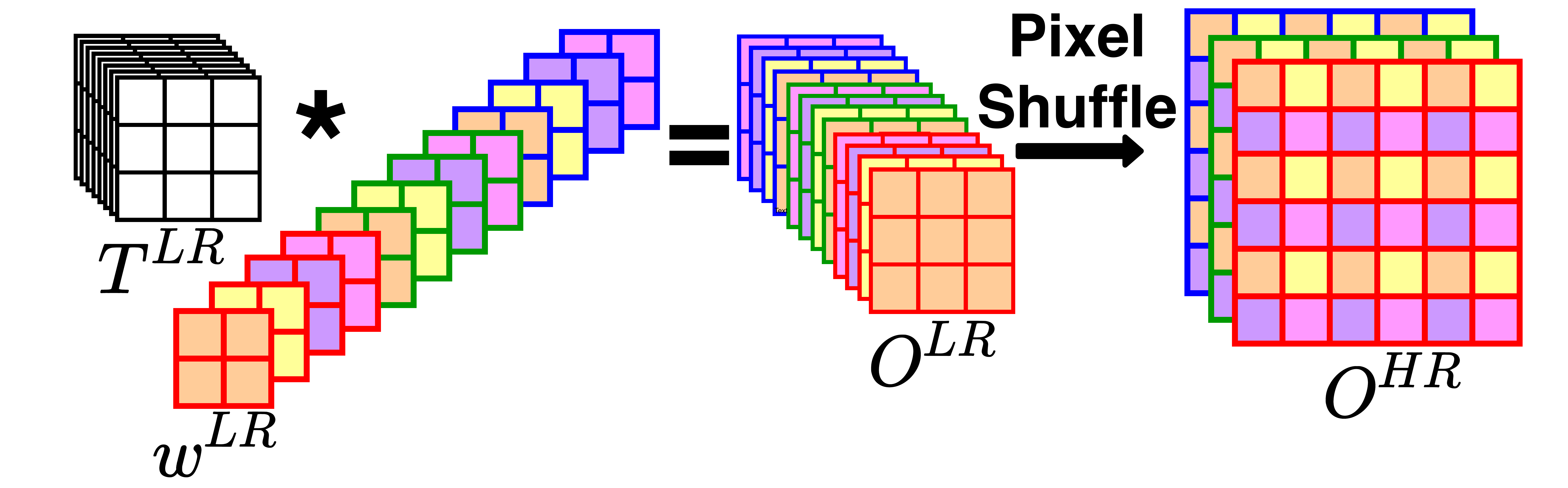} &  \includegraphics[width = 0.18\linewidth]{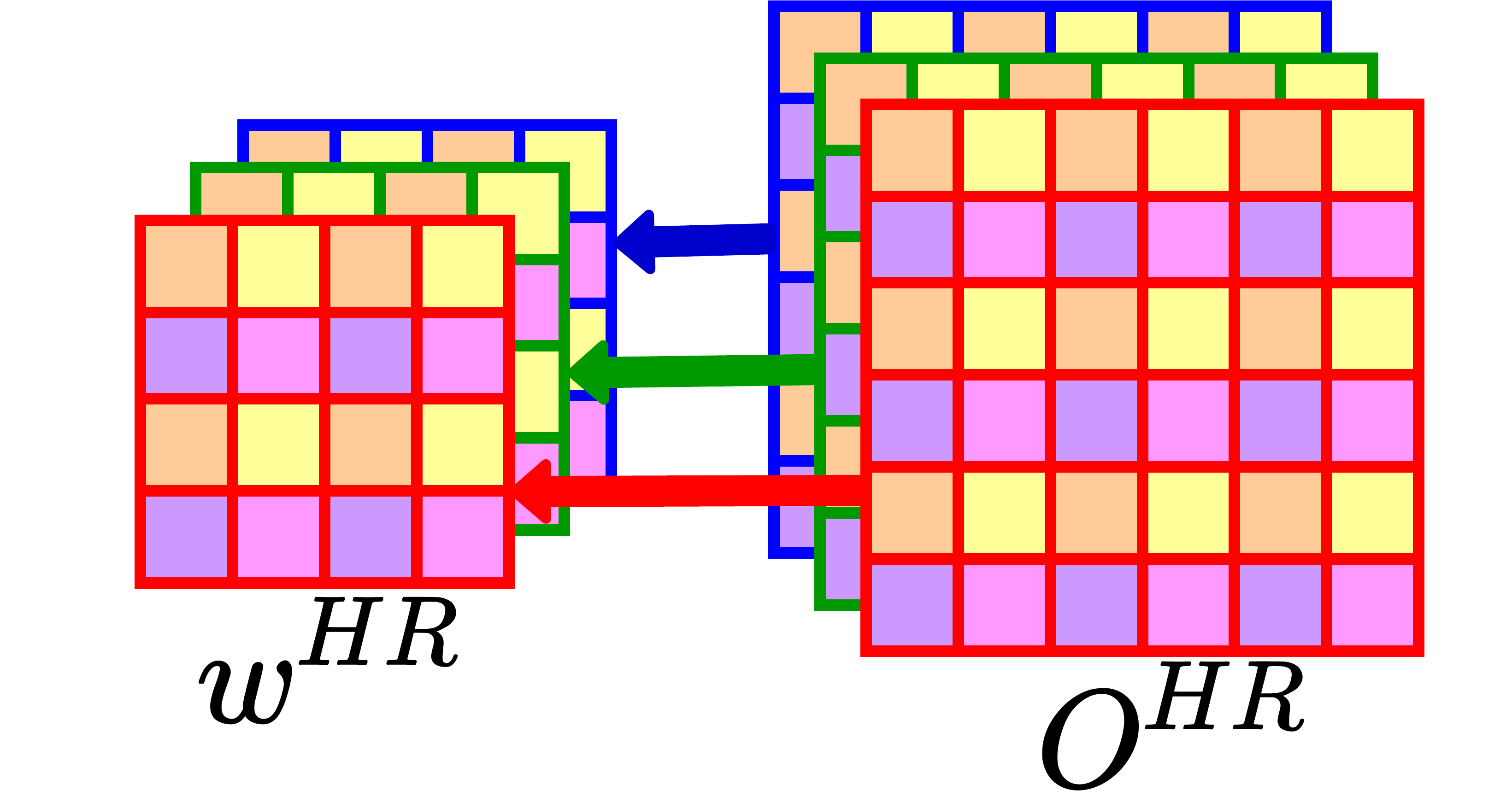}\\
 a) The usual upsampling operation in HR. & b) Implementing (a) in LR using PixelShuffle. & c) Less color correlation.\\&&with PixelShuffle \\ 
 \includegraphics[width = 0.50\linewidth]{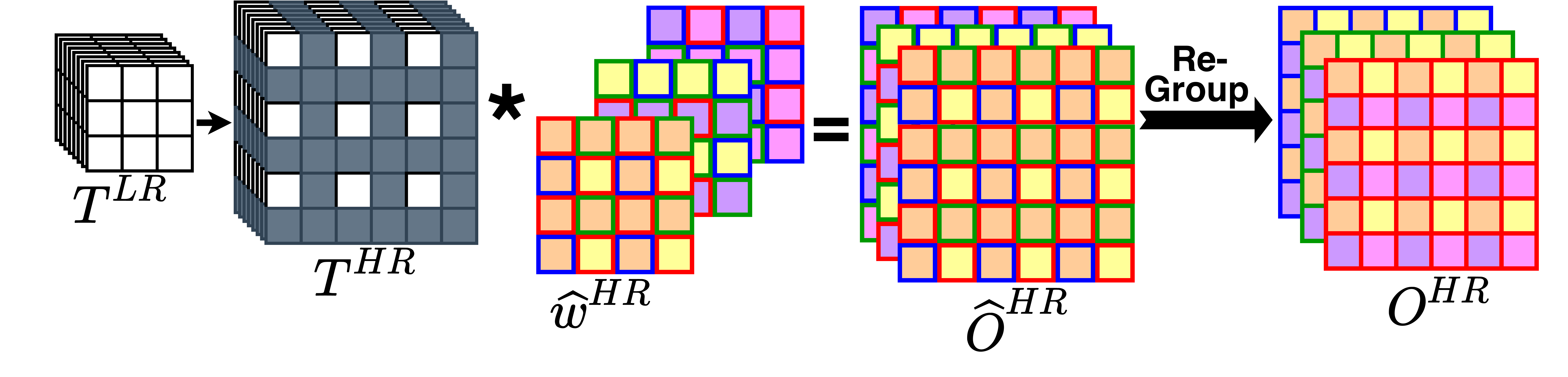} &  \includegraphics[width = 0.30\linewidth]{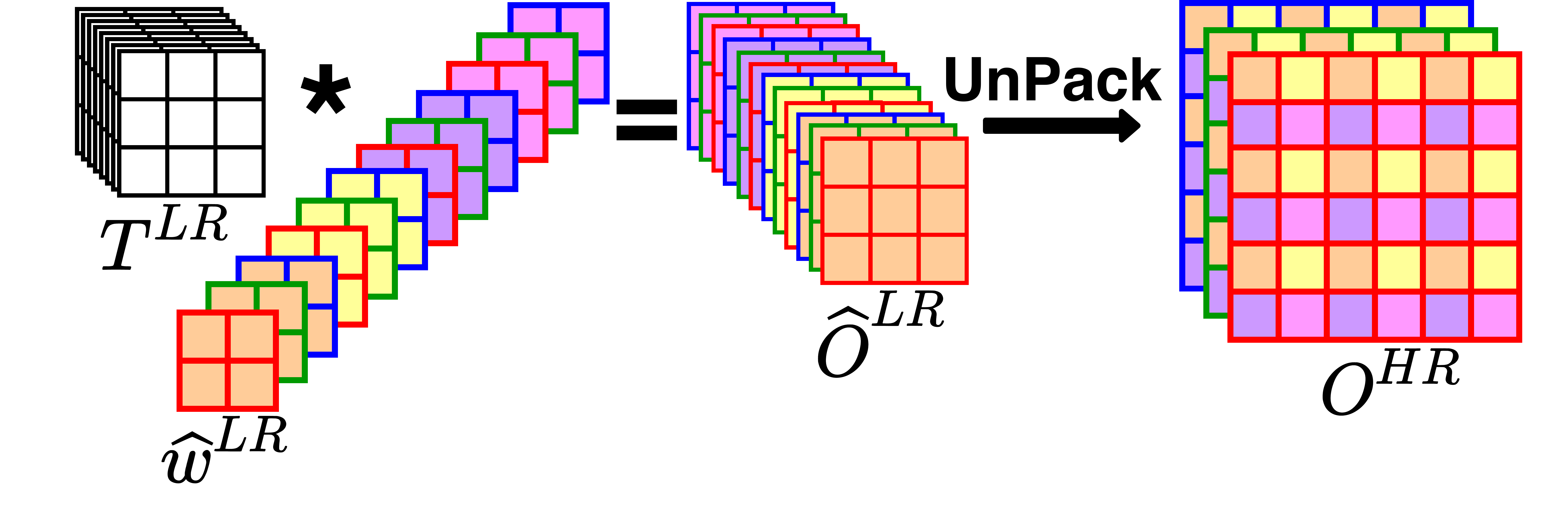} &  \includegraphics[width = 0.20\linewidth]{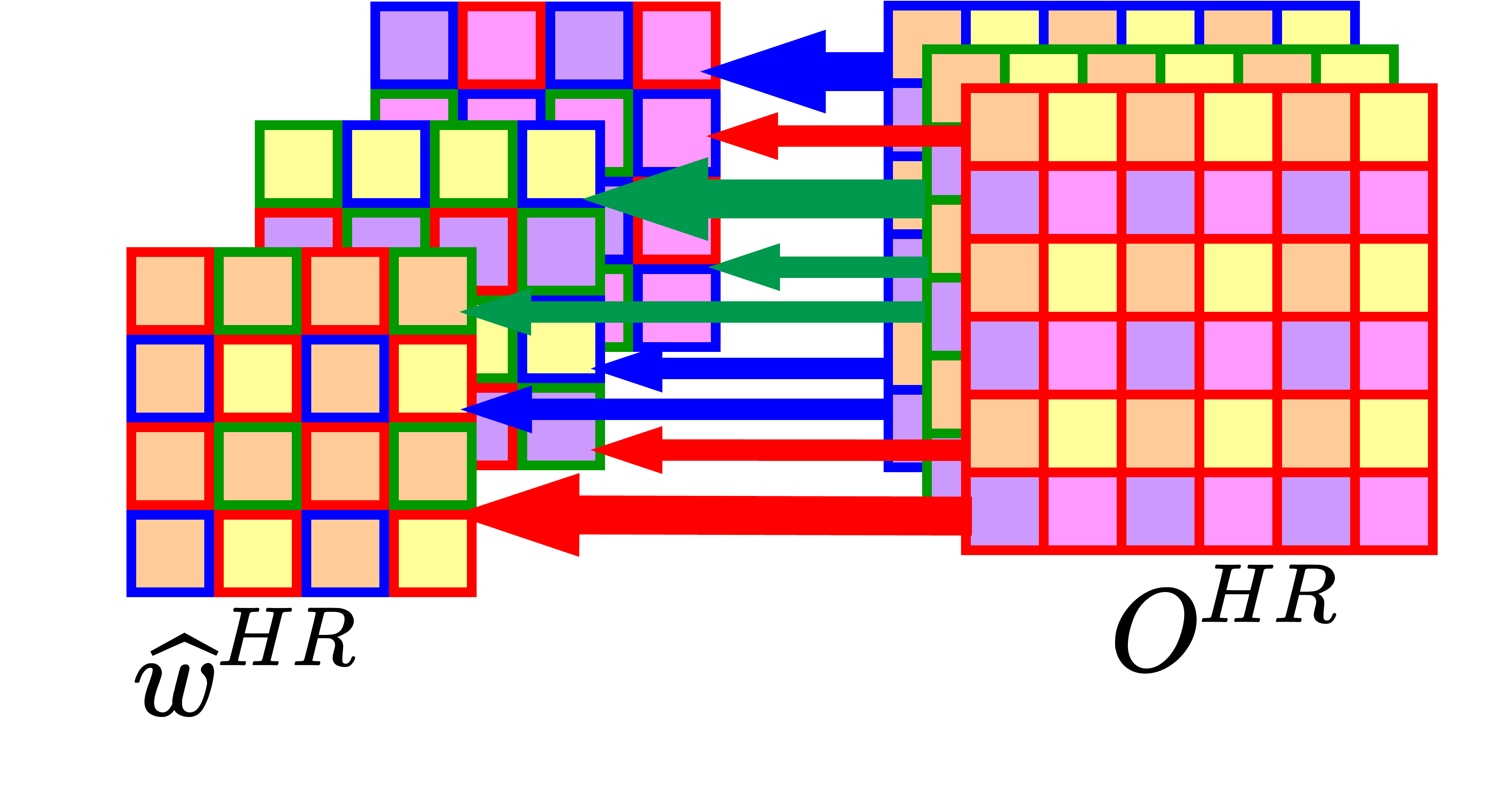} \\
 d) Upsampling in HR followed by regrouping for better color correlation. & e) Implementing (d) in LR using UnPack. & f) Better color correlation\\ && with UnPack.
\end{tabular}
\egroup
    \caption{Performing aggressive down/up-sampling causes several color distortions in the restored image. The proposed UnPack operation limits this by improving the color correlation in the restored image and simultaneously performing quick upsampling from LR to HR space. The effectiveness of the proposed solution is demonstrated in Fig. \ref{fig:packing_visual_comparison}.}
    \label{fig:packing}
 \vspace{-0.5cm}   
\end{figure}

First we analyze the PixelShuffle operation. In Fig. \ref{fig:packing} a), $T^{LR}$ refers to the penultimate feature map, which is upsampled with zero padding and then convolved with
$w^{HR}$
to obtain the restored image $O^{HR}$. 
We now explain the color coding used in the figure. When $w^{HR}$ convolves with $T^{HR}$, for each shifted position of $w^{HR}$, only the weights in one set of colors in $w^{HR}$ contribute to an output pixel in $T^{HR}$. We label the output pixel with the same color.
Doing convolution in HR is computationally expensive. However, an equivalent operation in the LR space can be performed as shown in Fig. \ref{fig:packing} b). This involves decomposing $w^{HR}$ into smaller kernels of $w^{LR}$ which are then convolved with $T^{LR}$ to produce $O^{LR}$. Using PixelShuffle, $O^{HR}$ can then be obtained from $O^{LR}$ . 
However, in this scheme, each kernel in $w^{HR}$ maintains a monopoly on one of the red, green or blue color channels in the restored image $O^{HR}$, see Fig. \ref{fig:packing} c). Thus, restoring images using PixelShuffle causes weak correlation among the color channels of $O^{HR}$, leading to color artifacts as shown in Fig. \ref{fig:packing_visual_comparison}. 

The goal of UnPack operation is to enhance the correlation among the color channels of $O^{HR}$. For this purpose, along with the upsampling, zero-padding and convolution operations, we introduce a re-grouping step as shown in Fig. \ref{fig:packing} d). This may appear to be a complicated two-stage operation, but using our UnPack operation we can easily perform an equivalent operation in the LR space, as shown in Fig. \ref{fig:packing} e). Note that this operation has the same time complexity as the operation shown in Fig. \ref{fig:packing} b).  For this operation, we decompose $\widehat w^{HR}$ into $\widehat w^{LR}$ and then apply UnPack $\alpha \times$. From Fig. \ref{fig:packing} f), we see that all the kernels of $\widehat w^{HR}$ are collectively responsible for all the colors in $O^{HR}$. Thus, UnPack operation leads to better color correlation than PixelShuffle.   

\textcolor{black}{The effectiveness of the proposed UnPack operation can also be intuitively understood in the LR space by comparing Fig. \ref{fig:packing} b) and Fig. \ref{fig:packing} e).} UnPack preserves the RGB ordering in the LR, whereas, PixelShuffle breaks this ordering, especially for large upsampling factors. For example, for a given spatial location in HR, PixelShuffle separates the Red and Blue pixels by $2\text{x}8^2=128$ channels in LR for $8\times$ upsampling. UnPack, however, always separates them by only 1 Green pixel for any upsampling factor. This is crucial because for CNNs it is well known that nearby features correlate more than spaced out ones \cite{rethinking_inception_2016}. Thus even though, UnPack does not introduce any new parametrization, its arrangement favors better color restoration. Therefore, in Fig. \ref{fig:packing_visual_comparison}, PixelShuffle's restored image is heavily affected by color cast, but no such distortion is observed in the case of UnPack.

\textbf{Increasing receptive field with Pack $\alpha \times$: }
Having a large receptive field is essential for capturing the contextual information in an image. 
Downsampling the incoming feature map using the novel Pack operation equips LLPackNet with a large receptive field. To illustrate this fact, let us consider a large feature map $I^{HR}$ which is downsampled to $I^{LR}$ using Pack 10$\times$ operation. Note that the neighboring pixels in $I^{LR}$ are actually $10$ pixels apart in $I^{HR}$. Also, the pixels along the channel dimension of $I^{LR}$ are in a $10\times 10$ neighborhood in $I^{HR}$. Thus, even using a $3\times 3$ convolution kernel on $I^{LR}$ with a stride of 1 leads to a receptive field of 900 pixels in $I^{HR}$. 
In contrast, to do a similar operation directly on $I^{HR}$, requires a $30\times 30$ kernel with a stride of $10$, which is impractical.

\label{fig:unpack}

\section{Experiments}
\begin{table}[t!]
\centering
\scriptsize
\setlength\tabcolsep{3pt}
\begin{tabular}{c|c|c|c|c|c}
\hline \hline \textbf{Model} & \textbf{Processing Time} & \textbf{Memory} & \textbf{Parameters} & \multicolumn{2}{c}{\textbf{PSNR(dB) / SSIM}} \\
& (in seconds) & ( in GB) & (in million) &\textbf{w/o GT exposure} & \textbf{using GT exposure}\\
\hline \hline
\textbf{Maharjan \etal \cite{ICME19}} &  $120$ & $10$ & $2.5$  & $20.98$ / $0.49$ & 28.41 / \textbf{0.81}\\
\textbf{Gu \etal \cite{gu2019self}} &  $77$ & $8$ & $3.5$  & $21.90$ / 0.59 & \textbf{28.53} / \textbf{0.81} \\   
\textbf{Chen \etal \cite{chen2018learning}} & $17$ & $5$ & $7.75$  & {22.93} / 0.70 & 28.30 / 0.79 \\ \hline
\textbf{Chen \etal \cite{chen2018learning} + Our Amplifier} & $17$ & $5$ & $7.76$  & {22.98} / \textbf{0.71} & 28.30 / 0.79 \\
\textbf{LLPackNet (Ours)} & \textbf{3} & \textbf{3} & \textbf{1.1} & \textbf{23.27} / 0.69 & 27.83 / 0.75 \\ \hline \hline
\end{tabular}
\caption{Results on the SID dataset \cite{chen2018learning} for extremely dark $2848 \times 4256$ RAW images. Compared to existing approaches, we have 2--7$\times$ fewer model parameters, 2--3$\times$ lower memory, 5--20$\times$ speed up with competitive restoration quality.}
\label{tab:main_comparison}
\end{table}
\subsection{Experimental settings}
For extreme low-light single-image enhancement, we compare with Chen \etal \cite{chen2018learning}
, Gu \etal \cite{gu2019self} and Maharjan \etal \cite{ICME19}. In addition, we also tried conventional techniques such as LIME \cite{2016lime} and Li \etal \cite{TIP2018structurePreserving} but they did not work well for dark images. The publicly available training and test codes of these methods have been used for the comparisons.
For experiments on dark images, we use See-in-the-Dark (SID) dataset \cite{chen2018learning} captured with high definition full-frame Sony $\alpha$7S II Bayer sensor. Unlike some methods that collect their dataset by simulating pairs of low-light and GT images \cite{2017llnet,SIMULATE2017,SIMULATE2019, 2019simulateAdobe,automaticFiveSIMULATE2018,LOL}, SID provides physically captured extreme low-light RAW images of resolution 2848$\times$4256.
We additionally show comparisons on the LOL dataset \cite{LOL} to evaluate the performance of LLPackNet on a notably distinct test set-up. In contrast to SID, LOL has weakly illuminated VGA resolution PNG compressed images. Additionally, SID comes with GT and low-light exposure information, which can be used for estimating the pre-amplification factor, but LOL has no such information. 

We use the train/test split as given in the datasets. For LLPackNet, patches of size $512 \times 512$ are used for training and full resolution for testing. For benchmarking, we use the PyTorch \cite{pytorch} framework on Intel Xeon E5-1620V4 @ 3.50 GHz CPU with 64 GB RAM. We use the default Adam optimizer of PyTorch with fixed learning rate of $10^{-4}$. All convolutions use kernels of size $3\times3$ with \textit{He initialization} \cite{he2015delving}. Our network was allowed to train for 400,000 iterations. We use, $\lambda_1=1$, $\lambda_2=3$, $\lambda_3=1$, $\lambda_4=400$ and $\lambda_5=10^{-6}$.  

\subsection{Restoration results for extreme low-light images}
\begin{figure*}[t!]
\centering
	\scriptsize
	\bgroup
\def\arraystretch{0.5}
\setlength\tabcolsep{1pt}
	\begin{tabular}{c}
	
    \begin{tabular}{cccccc}
    &
    {\footnotesize \textbf{Maharjan \etal}} & {\footnotesize \textbf{Gu \etal}} & {\footnotesize \textbf{Chen \etal}} & {\footnotesize \textbf{Ours}} & {\footnotesize \textbf{GT}}
    
    \\
    
    \multirow{8}{*}[8.1ex]{\rotatebox[origin=c]{90}{\textbf{\small (A) Amplification using GT exposure}}} & 
    \includegraphics[width=0.15 \linewidth]{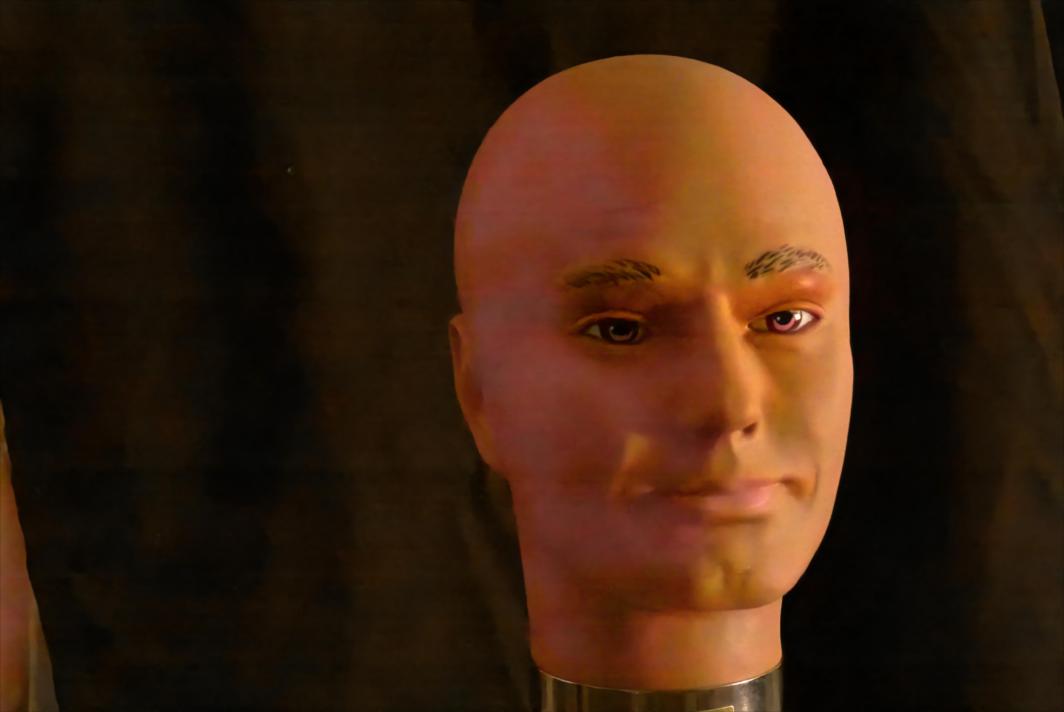}   &
    \includegraphics[width=0.15 \linewidth]{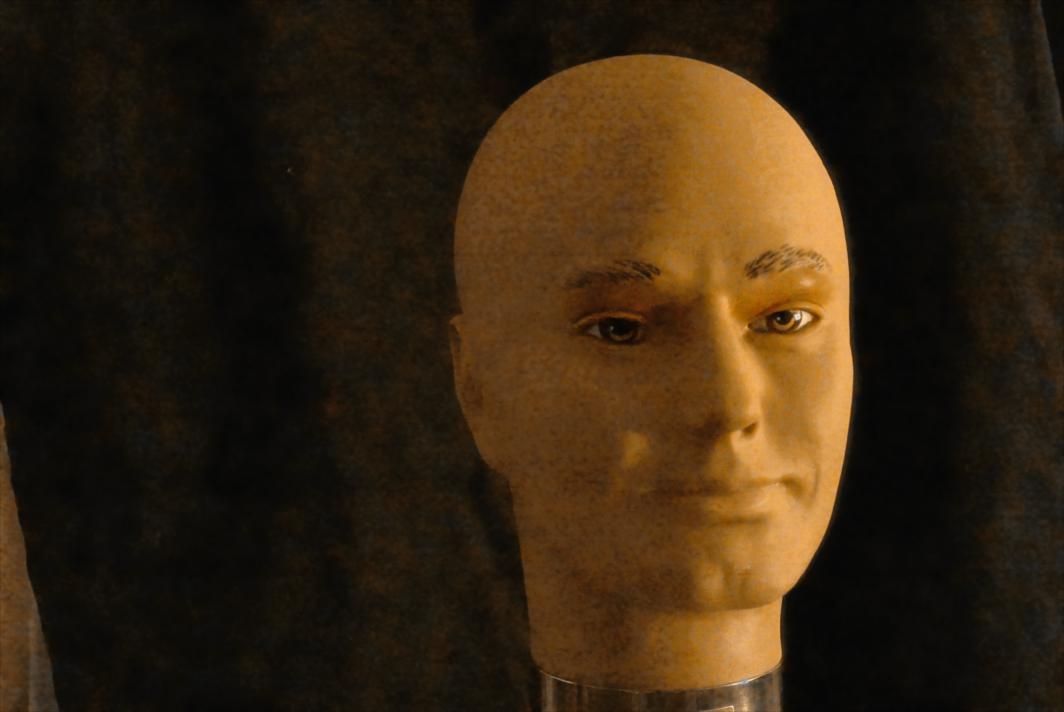} &
    \includegraphics[width=0.15 \linewidth]{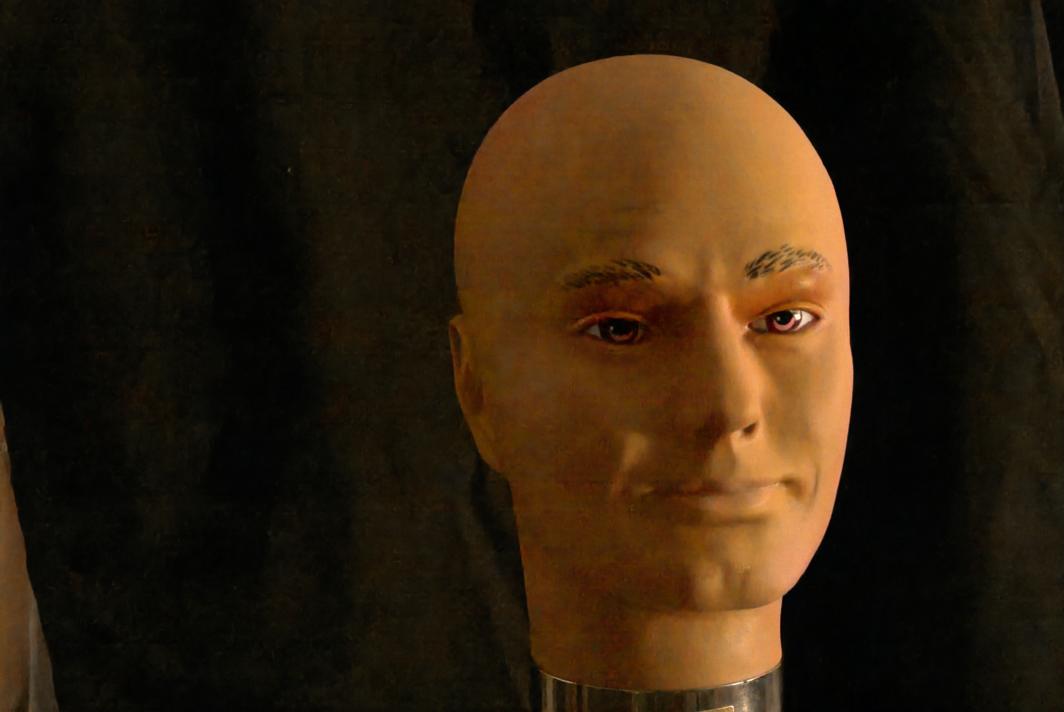}   & 
     \includegraphics[width=0.15 \linewidth]{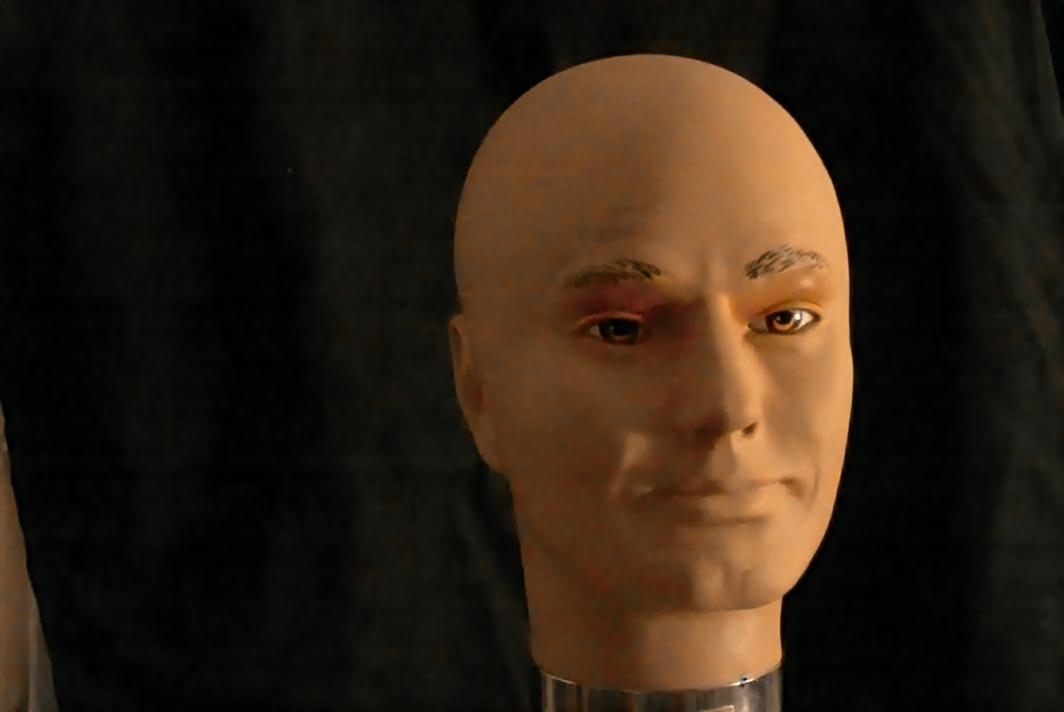} &  
     \includegraphics[width=0.15 \linewidth]{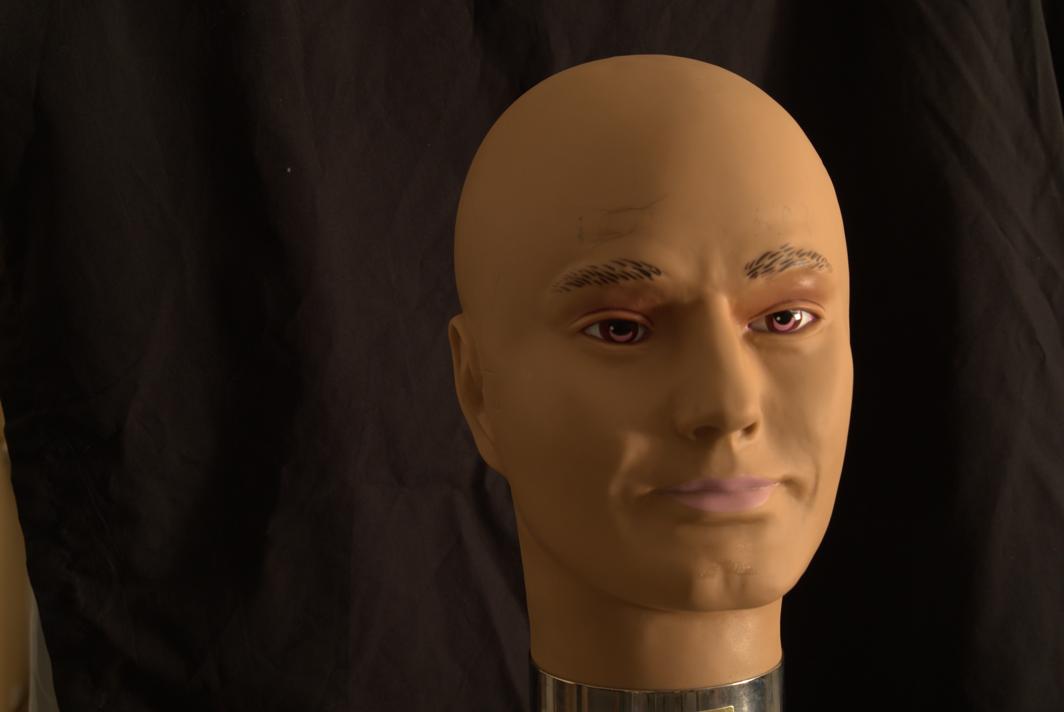} \\
       &  28.01/0.78 & 28.50/\textbf{0.82} & 28.65/\textbf{0.82} &  \textbf{29.00}/0.78 & \\
     
        & 
     \includegraphics[width=0.15 \linewidth]{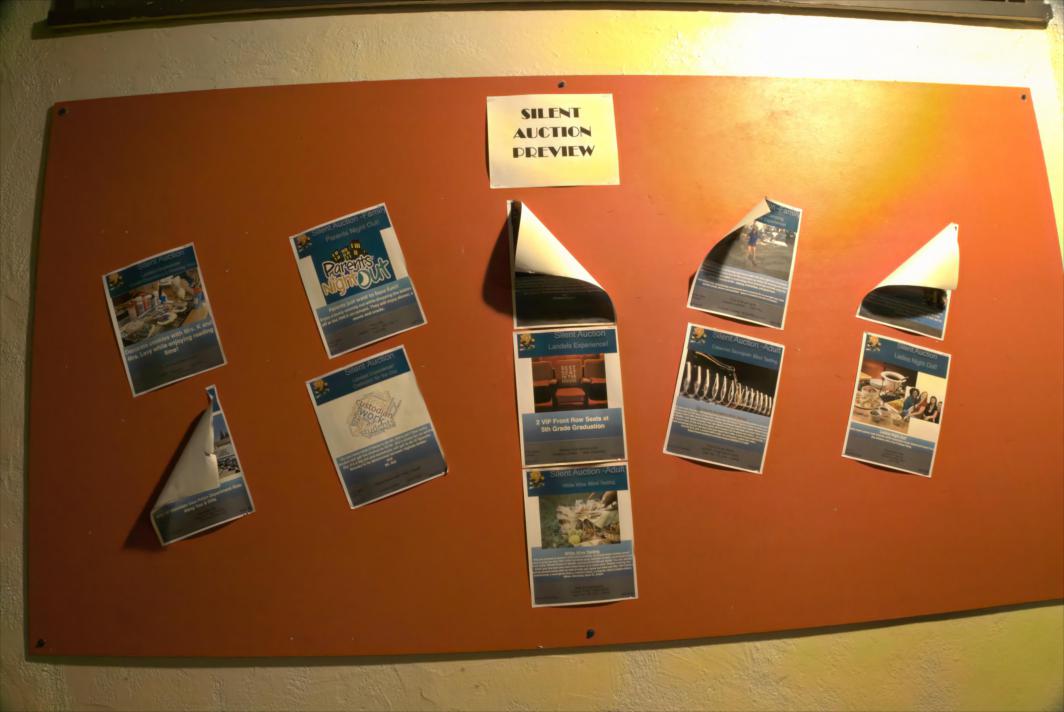}   &
    \includegraphics[width=0.15 \linewidth]{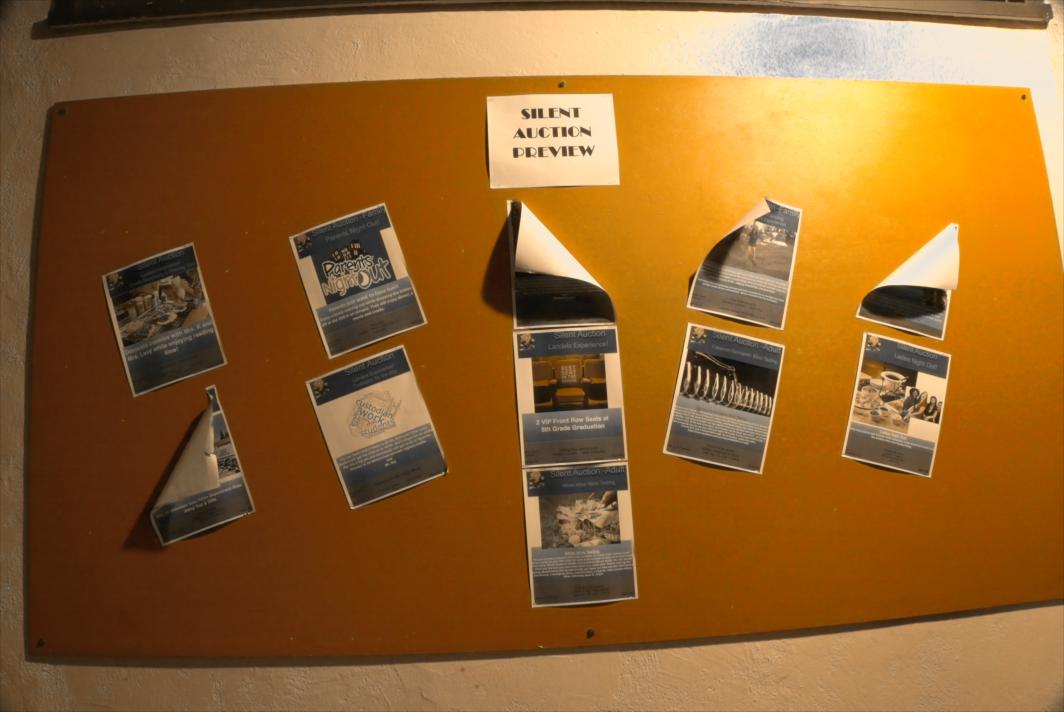} &
    \includegraphics[width=0.15 \linewidth]{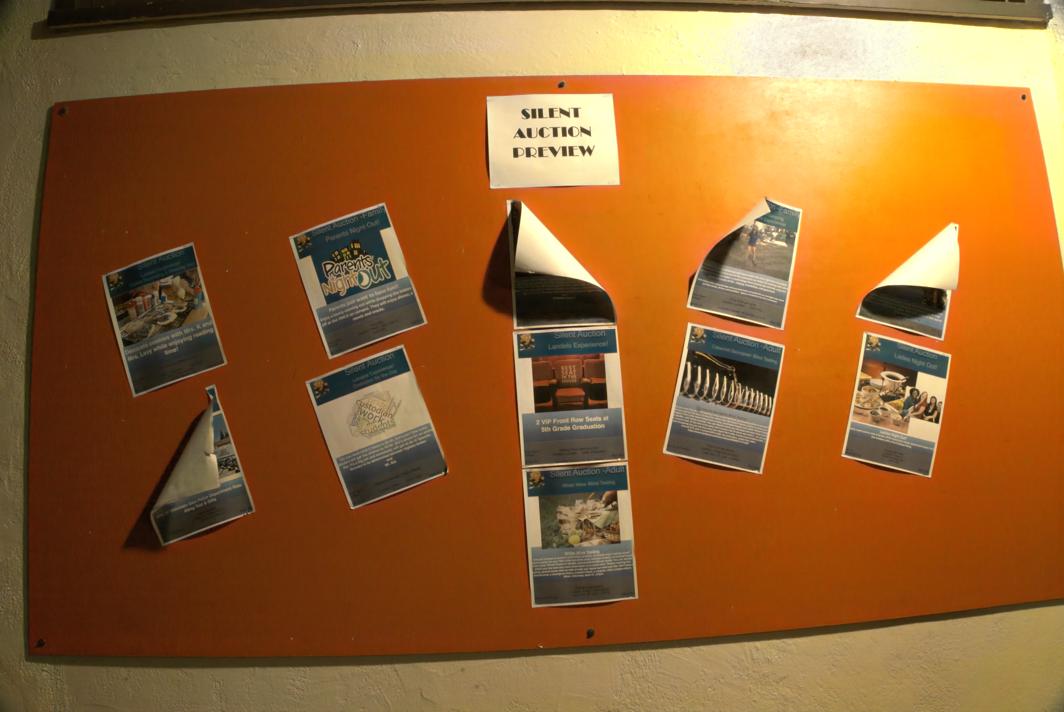}   & 
     \includegraphics[width=0.15 \linewidth]{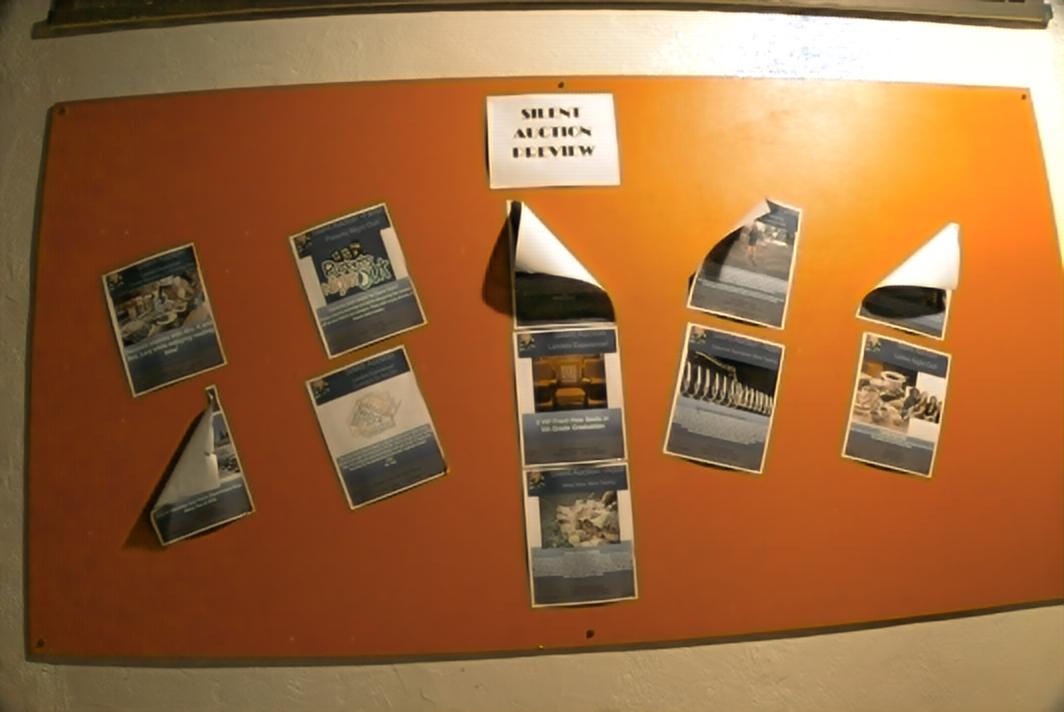} &  
     \includegraphics[width=0.15 \linewidth]{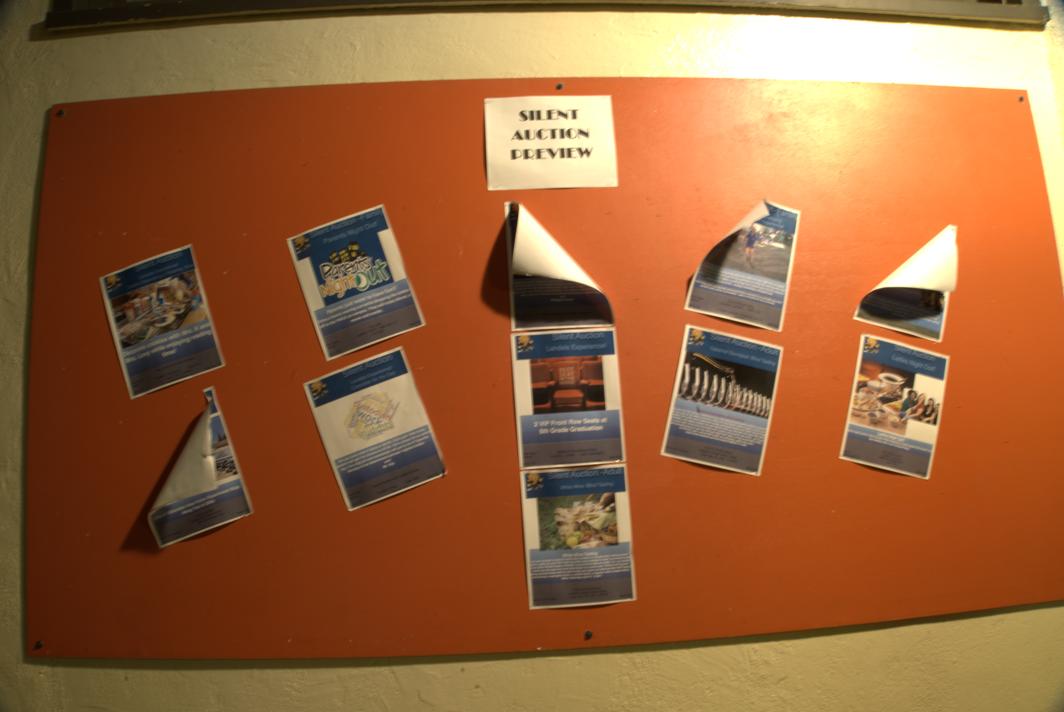} \\   & 
     22.15/0.89 & 21.01/0.88 & 21.09/0.89 & \textbf{22.23/0.90} & \\
     
        & 
     
     \includegraphics[width=0.15 \linewidth]{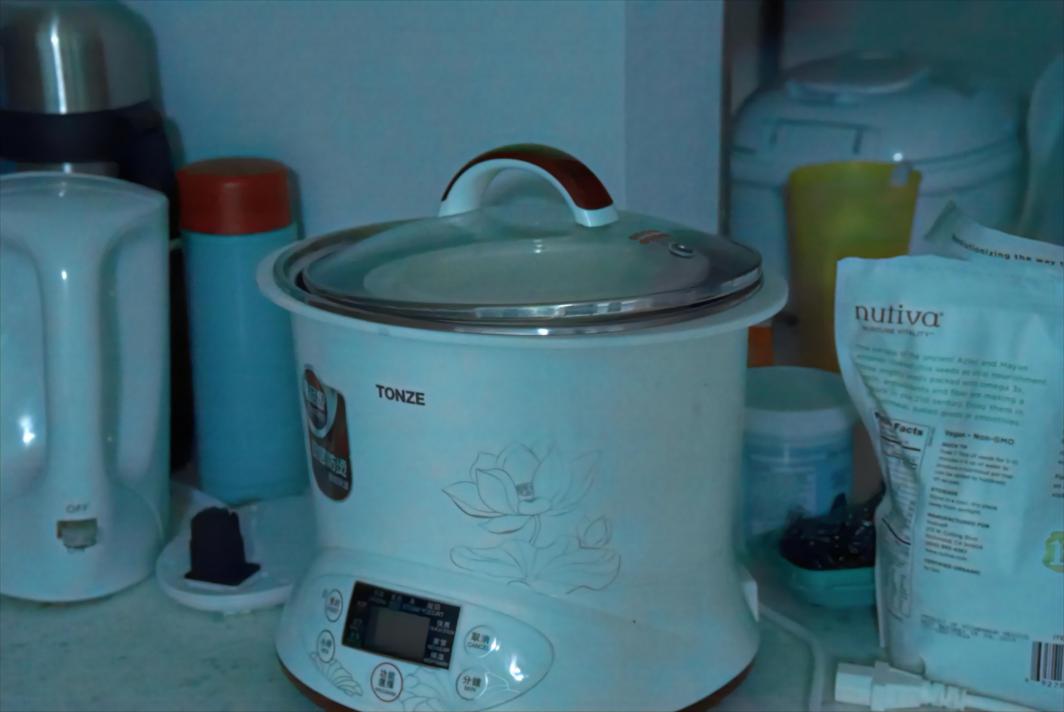}   &
    \includegraphics[width=0.15 \linewidth]{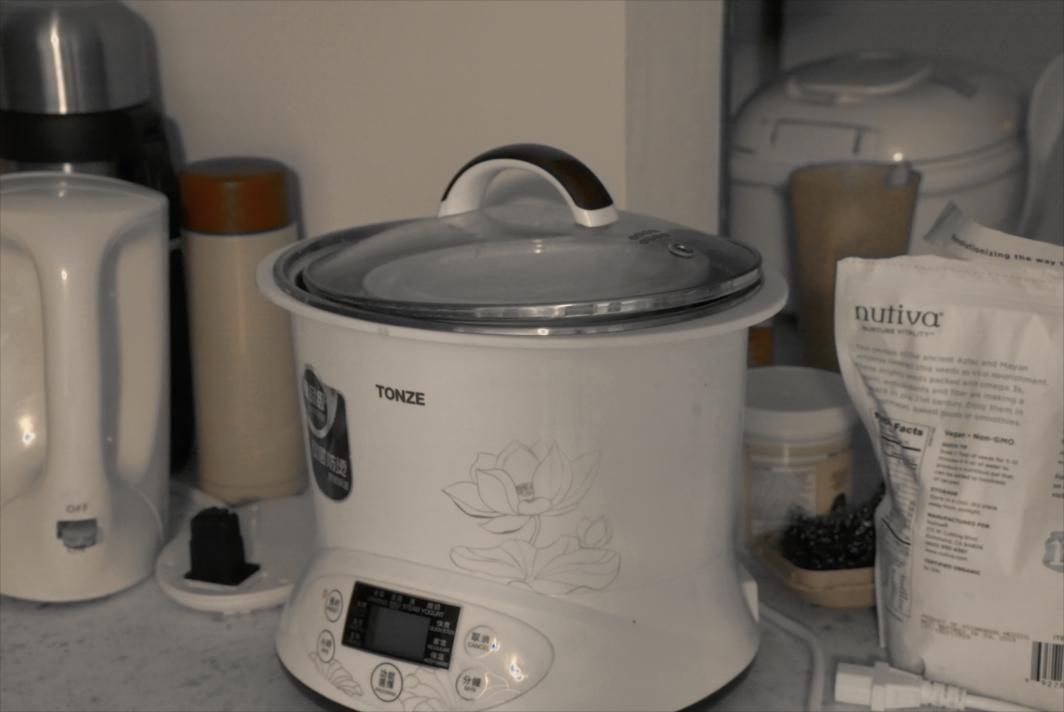} &
    \includegraphics[width=0.15 \linewidth]{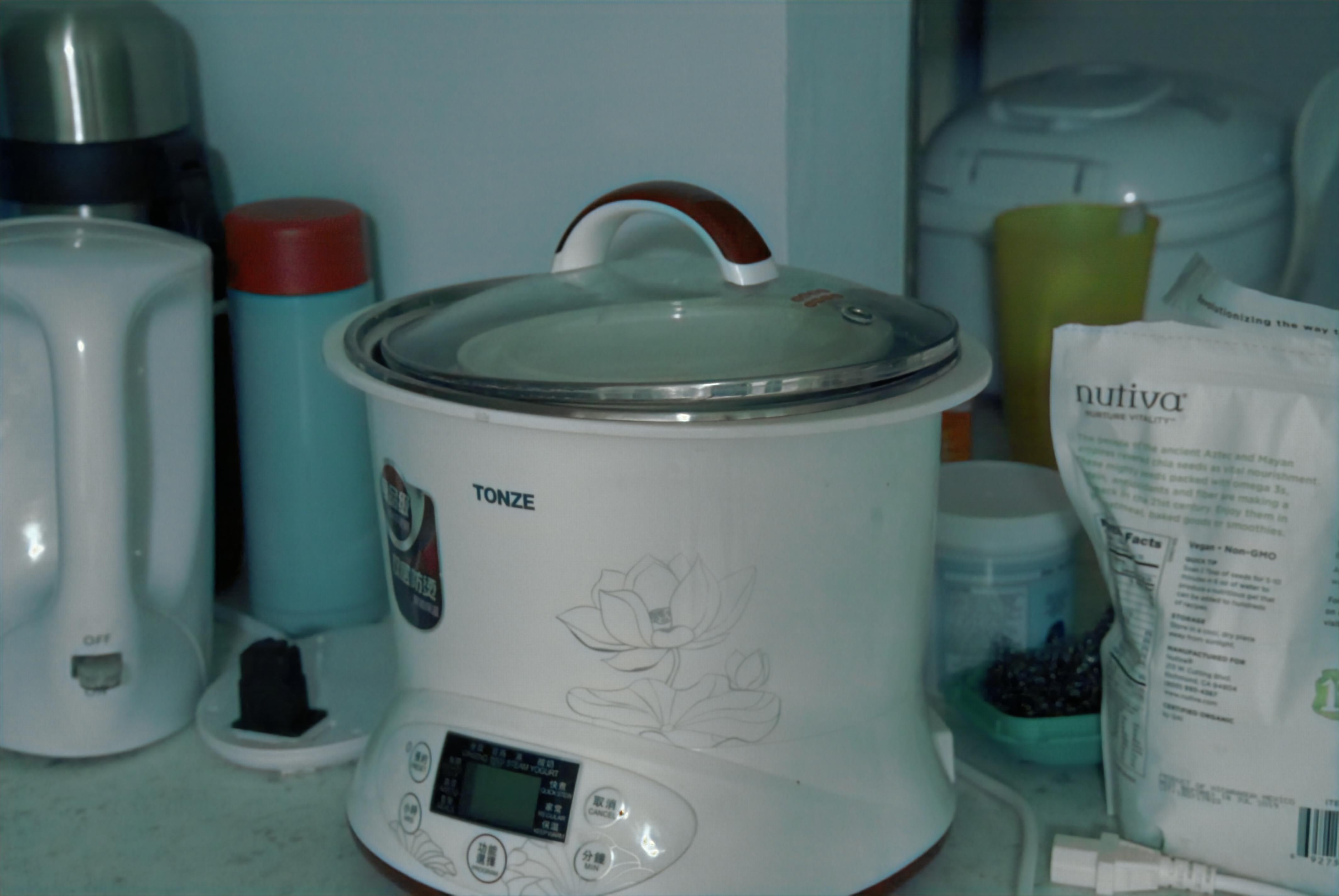}   & 
     \includegraphics[width=0.15 \linewidth]{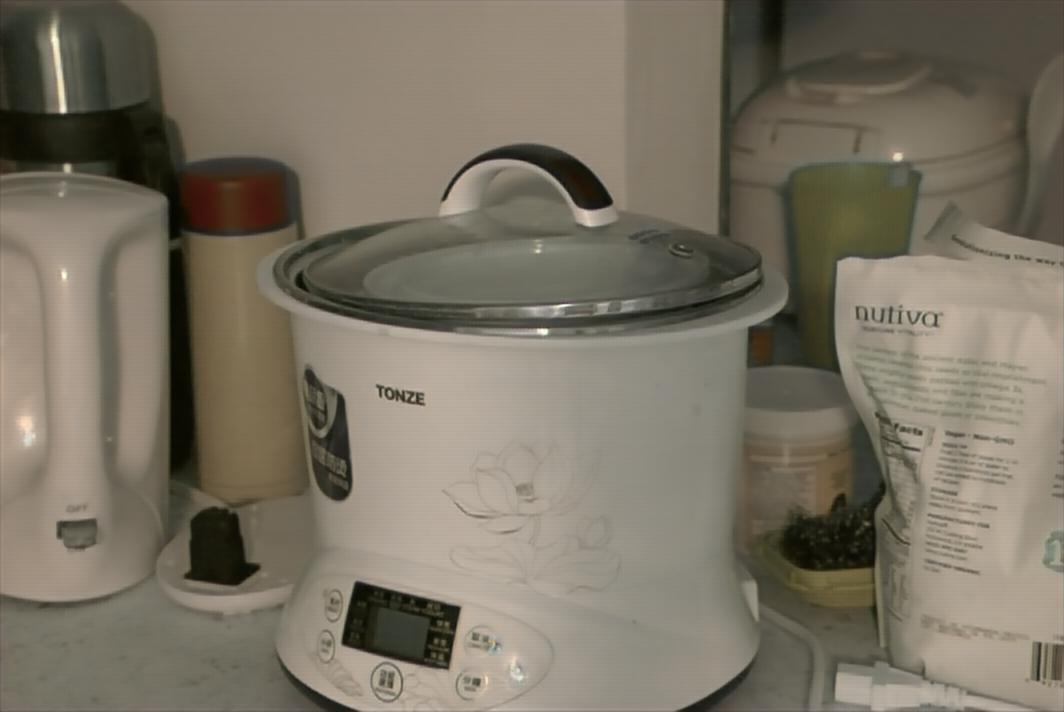} &  
     \includegraphics[width=0.15 \linewidth]{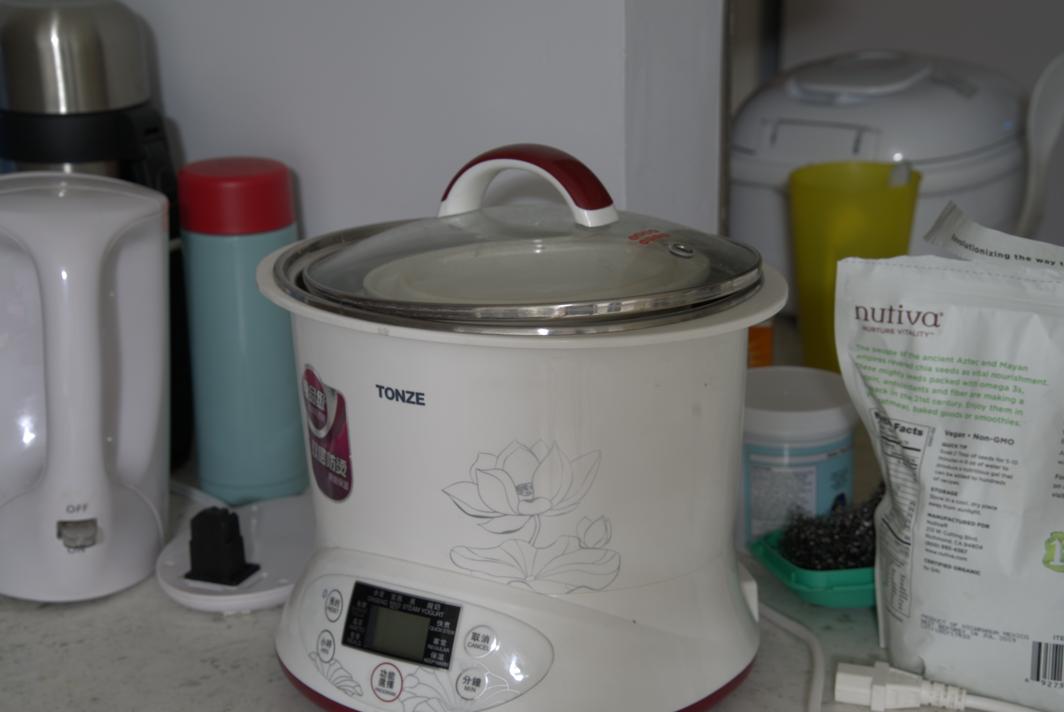} \\   &  25.60/0.59 & \textbf{27.88/0.70} & 25.66/0.60 & 27.87/0.62 & \\
      & 
     \includegraphics[width=0.15 \linewidth]{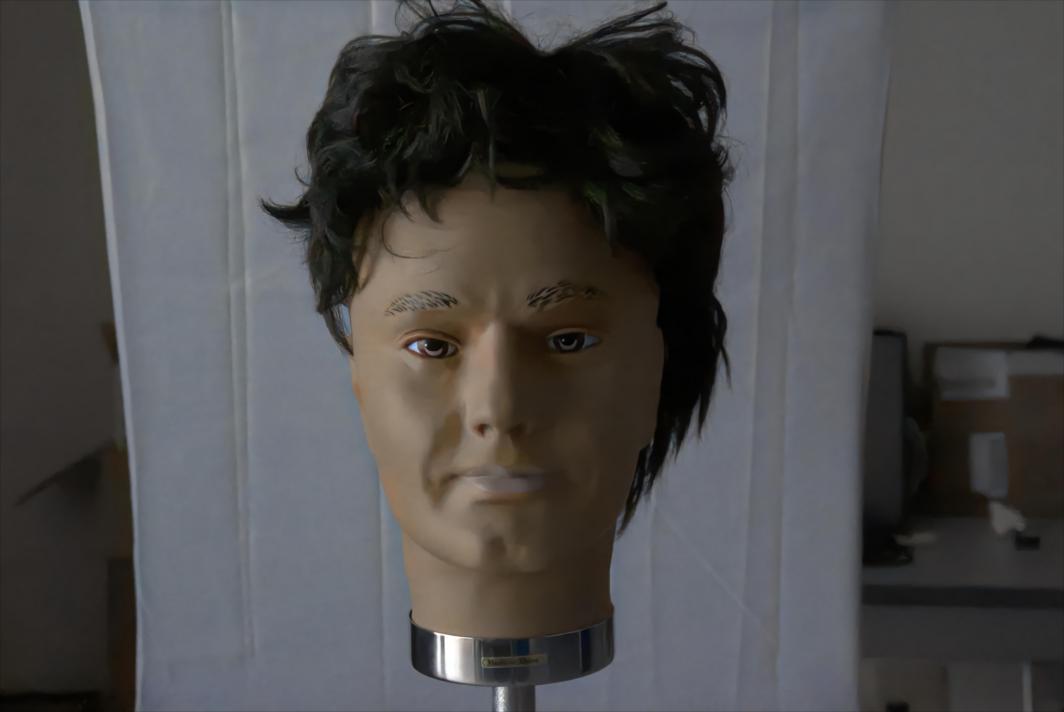}   &
    \includegraphics[width=0.15 \linewidth]{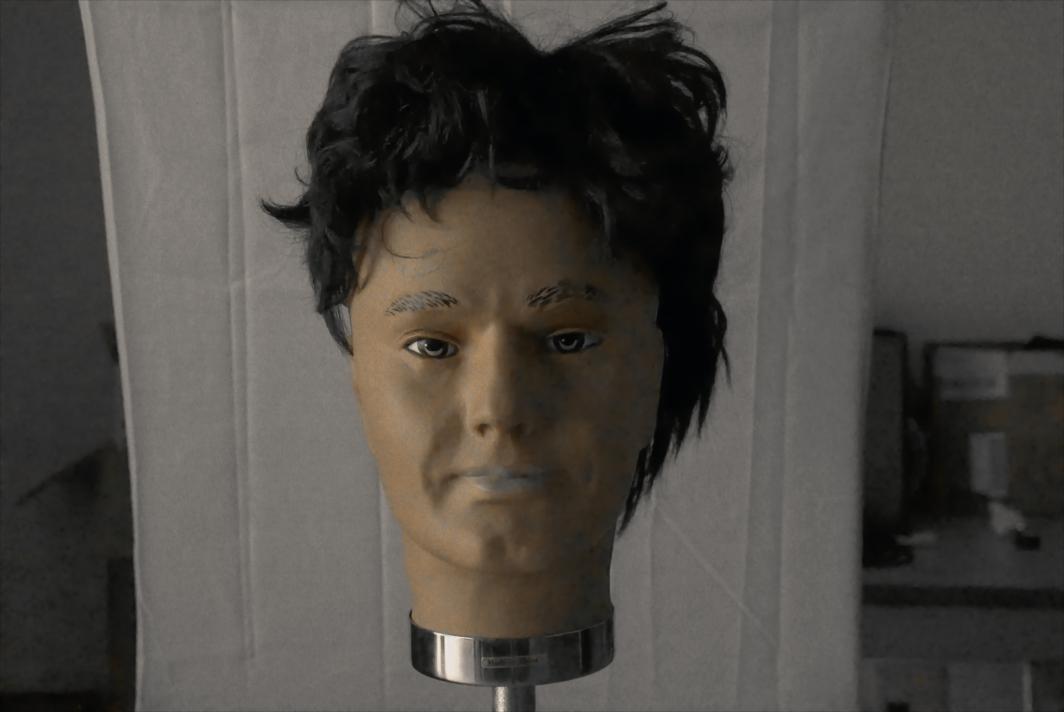} &
    \includegraphics[width=0.15 \linewidth]{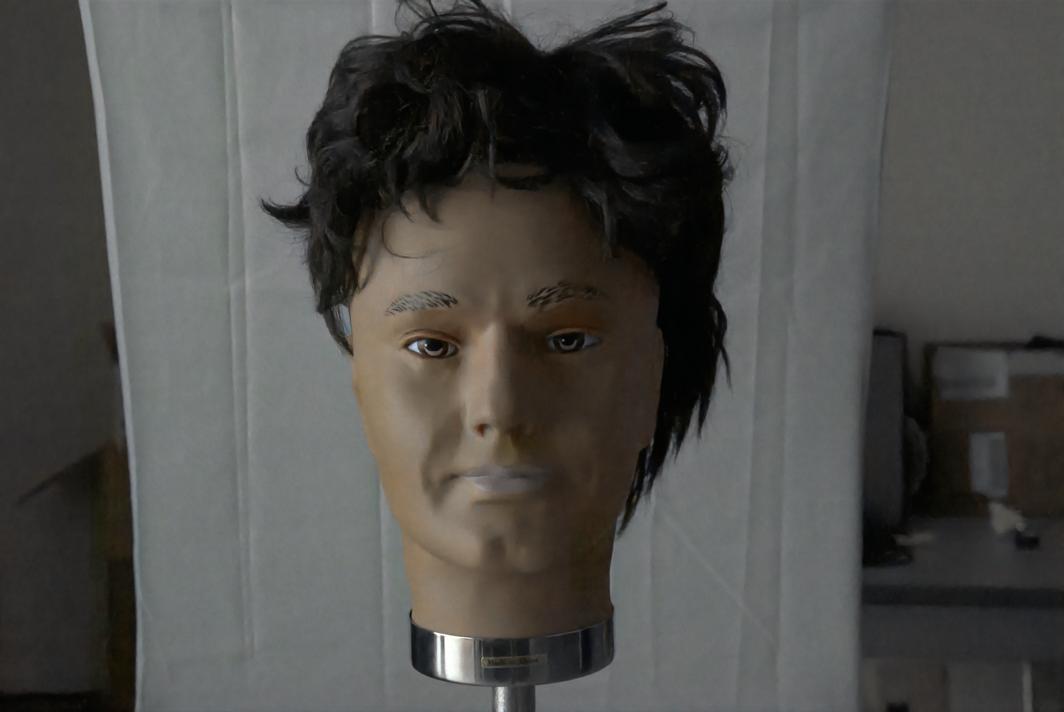}   & 
     \includegraphics[width=0.15 \linewidth]{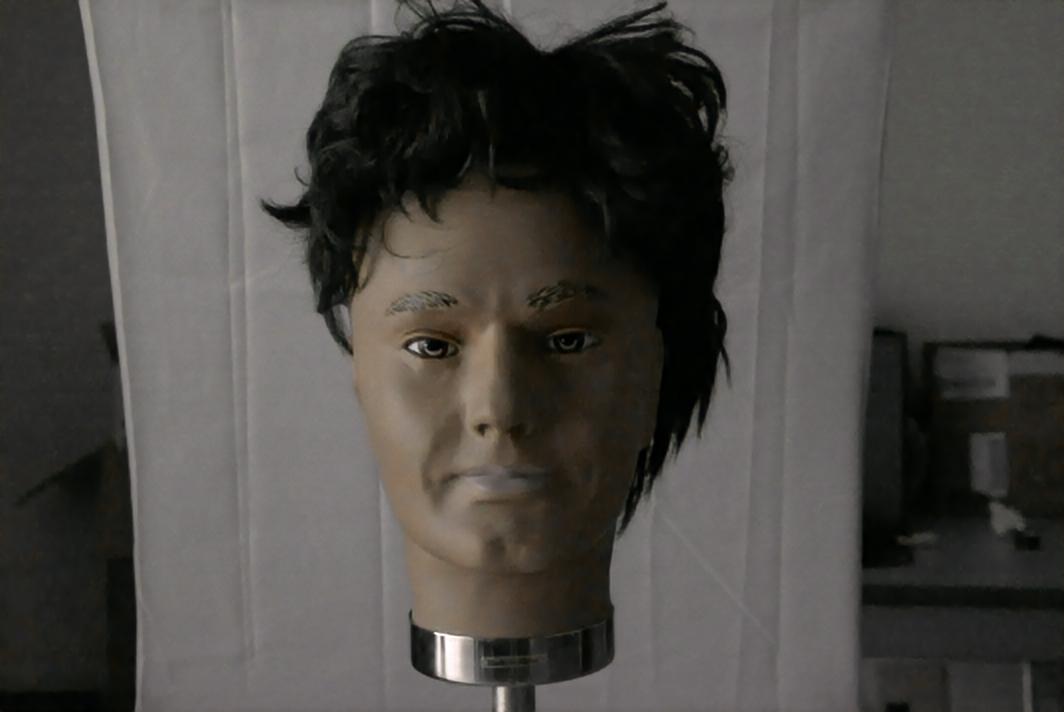} &
     \includegraphics[width=0.15 \linewidth]{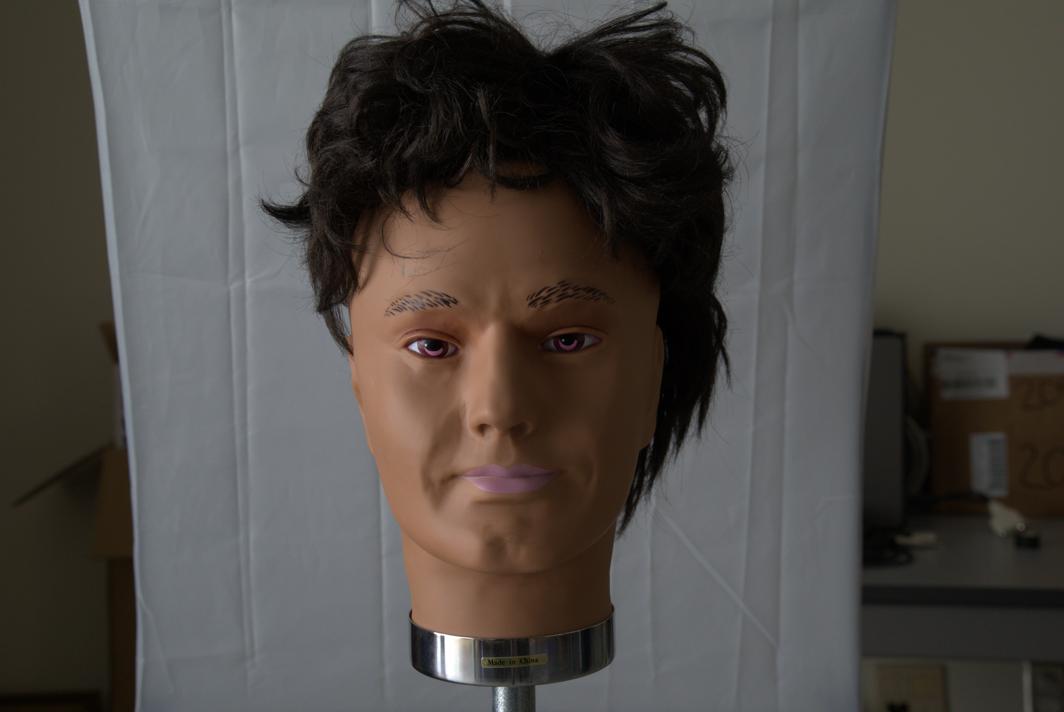} \\   &  31.50/\textbf{0.85} & \textbf{31.85}/0.81 & 31.20/0.79 & 29.77/0.68 &  \vspace{2pt}
     \end{tabular} \\
     \hline \hline \vspace{-2pt} \\
     
     \begin{tabular}{ccccccc}
     & {\notsotiny \textbf{Maharjan \etal}} & {\notsotiny \textbf{Gu \etal}} & {\notsotiny \textbf{Chen \etal}} & 
     {\tiny \textbf{Chen \etal + Our Amplifier}} &
     {\notsotiny \textbf{Ours}} & {\notsotiny \textbf{GT}} \\
      \multirow{8}{*}[2.3ex]{\rotatebox[origin=c]{90}{\textbf{\small (B) w/o using GT exposure}}}
      &  \includegraphics[width=0.15 \linewidth]{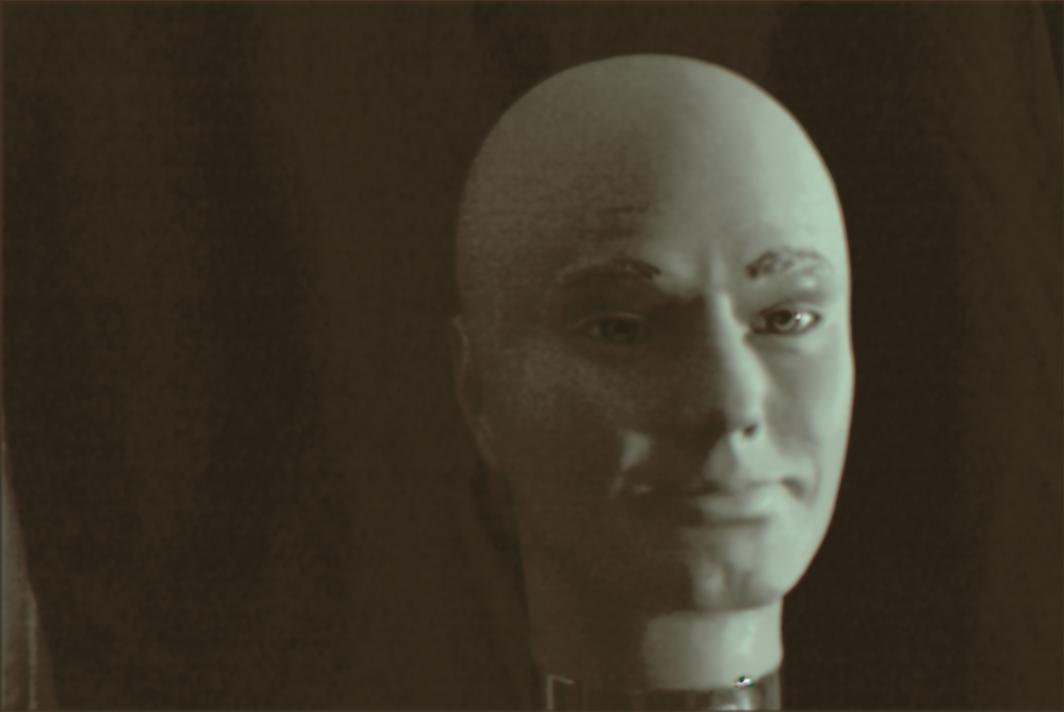}   &
    \includegraphics[width=0.15 \linewidth]{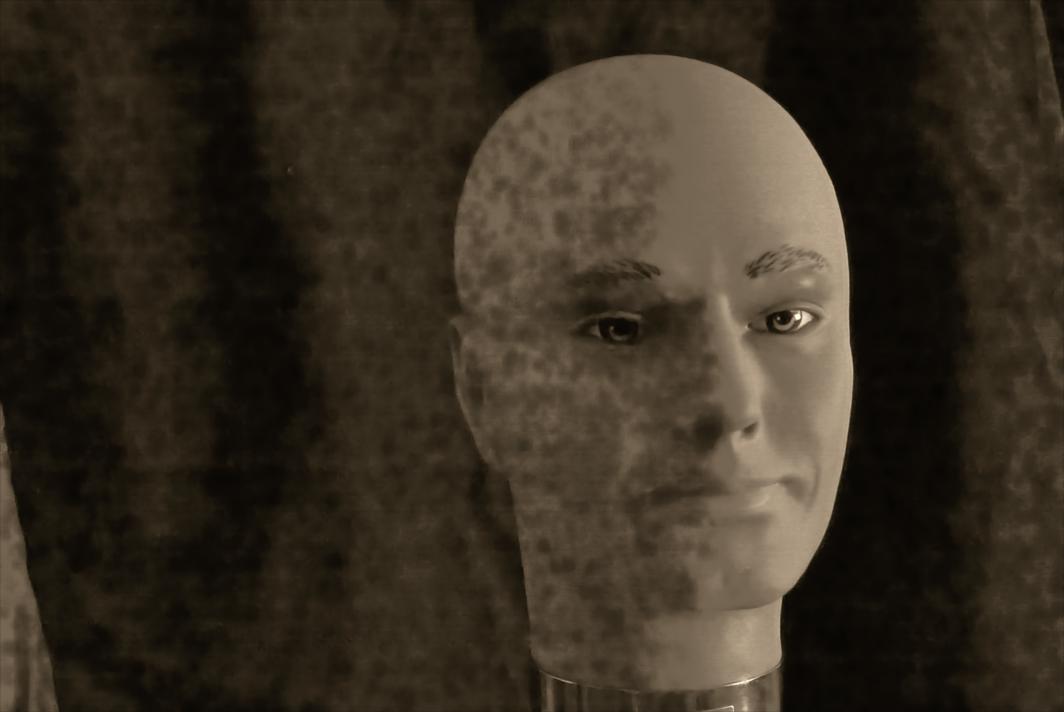} &
    \includegraphics[width=0.15 \linewidth]{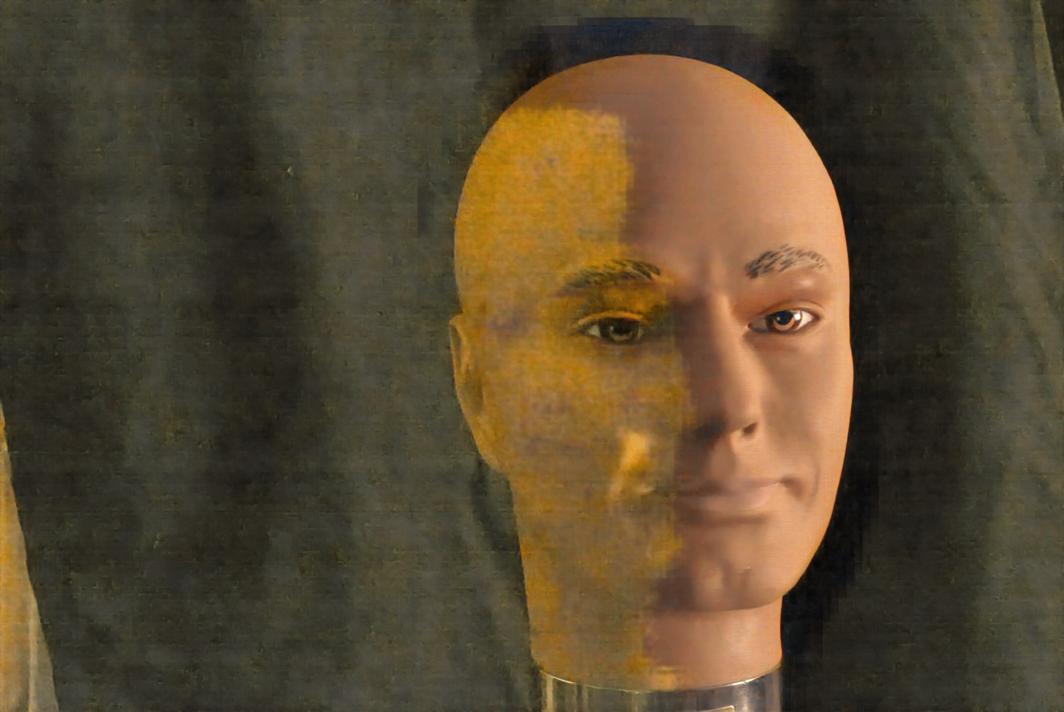}  
    &
    \includegraphics[width=0.15 \linewidth]{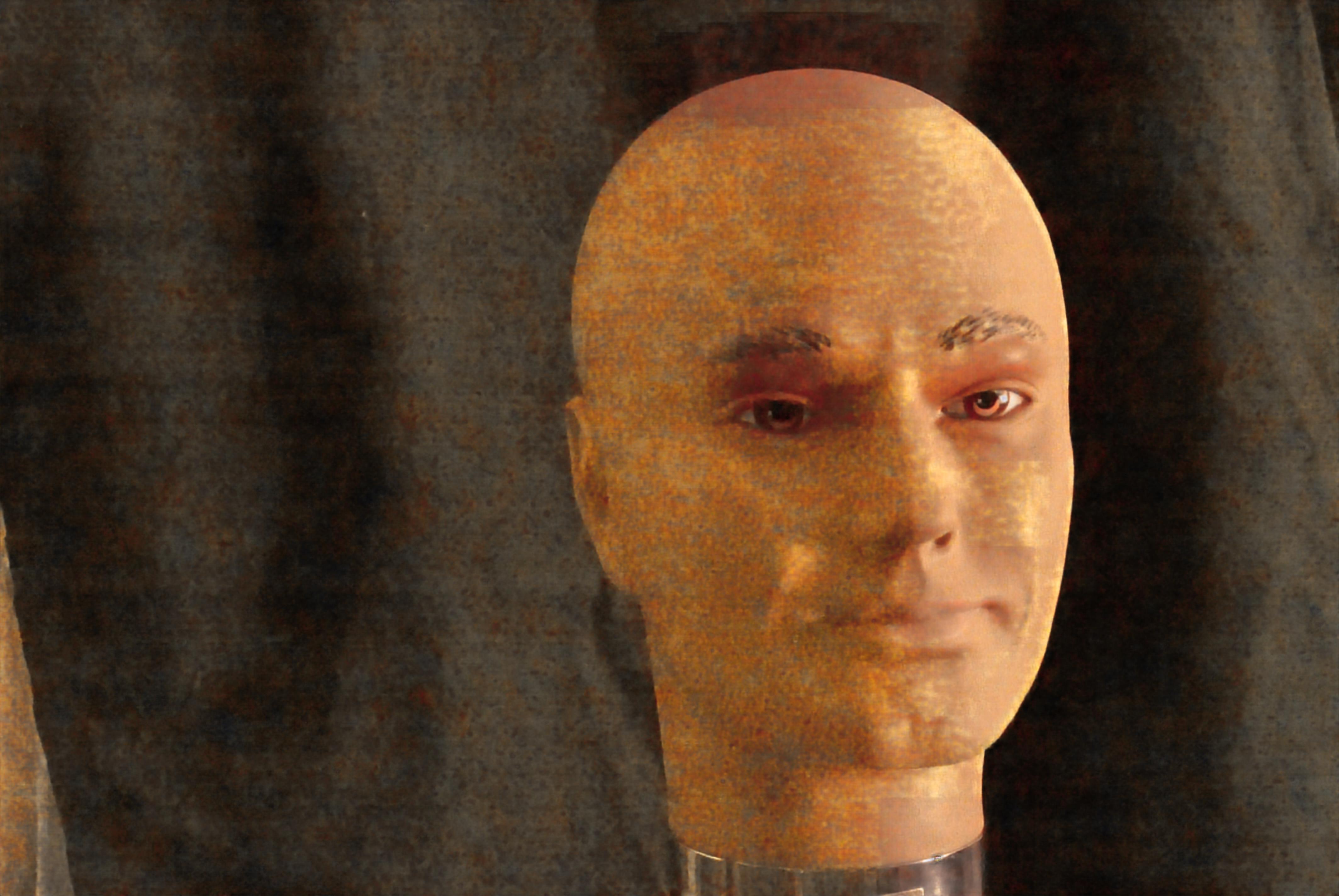} & 
     \includegraphics[width=0.15 \linewidth]{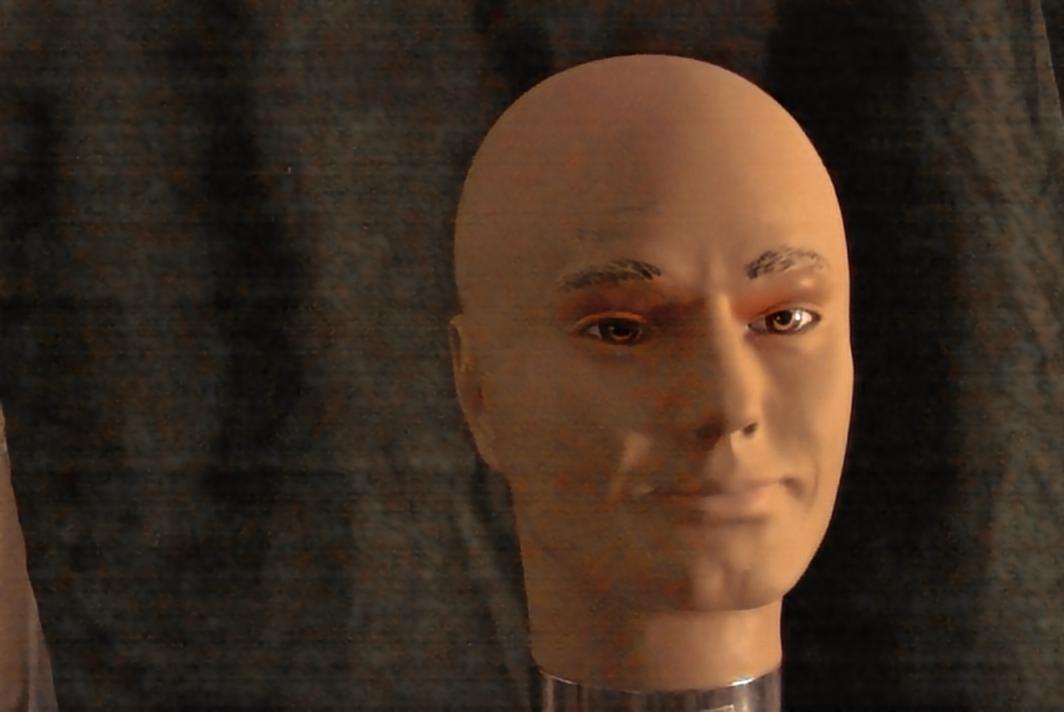} & 
     \includegraphics[width=0.15 \linewidth]{ratio_boosting/14_IMG_GT.jpg}  \\   &  20.58/0.54 & 23.86/\textbf{0.78} & 18.84/0.56 & 18.60/0.64 & \textbf{24.69/0.78} & \\
     
       & 
     \includegraphics[width=0.15 \linewidth]{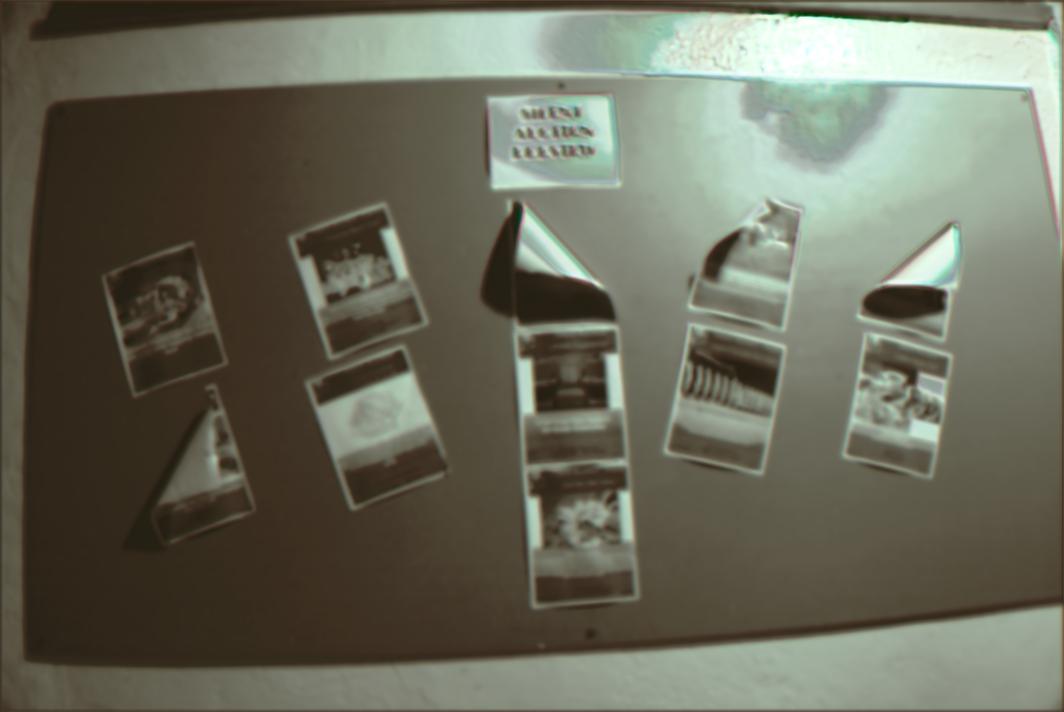}   &
    \includegraphics[width=0.15 \linewidth]{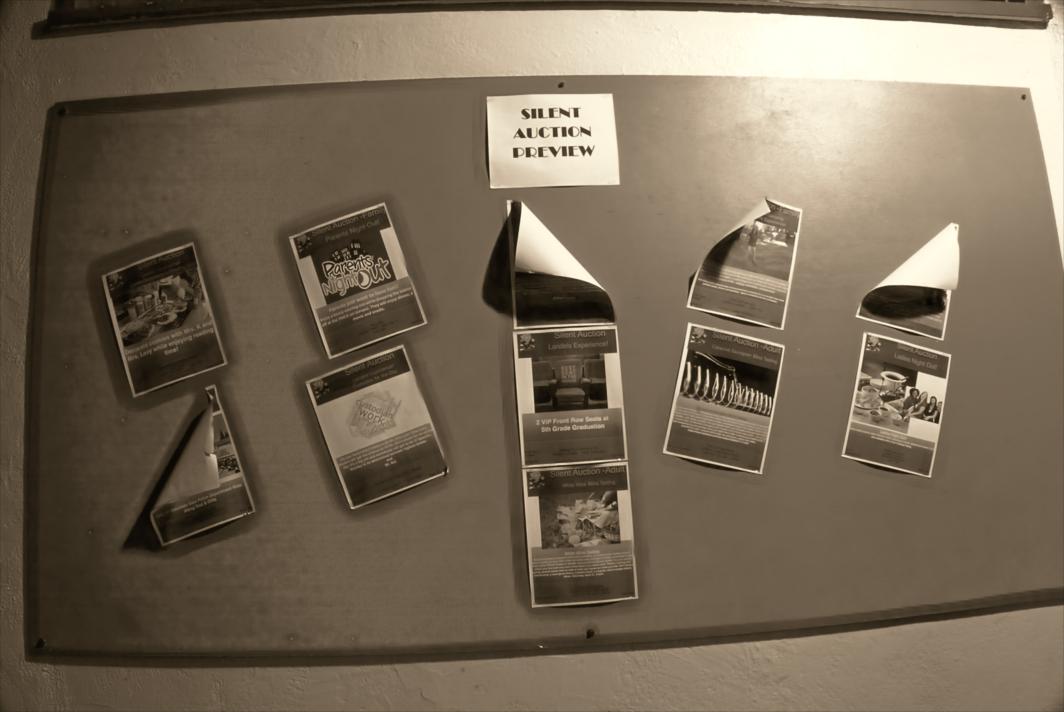} &
    \includegraphics[width=0.15 \linewidth]{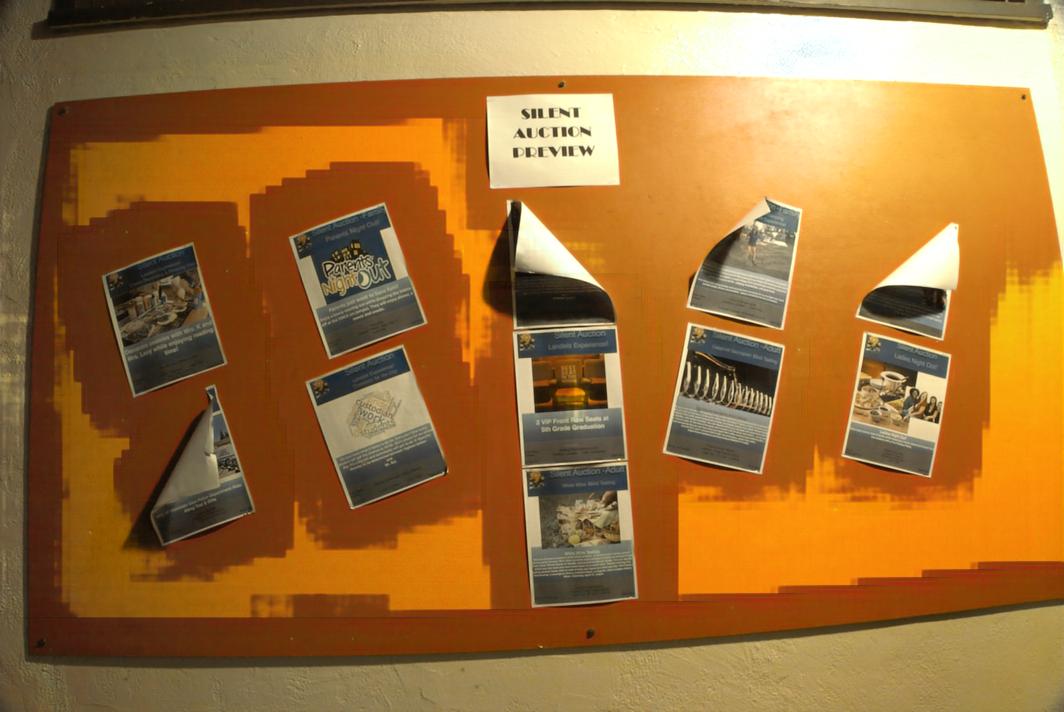} &
    \includegraphics[width=0.15 \linewidth]{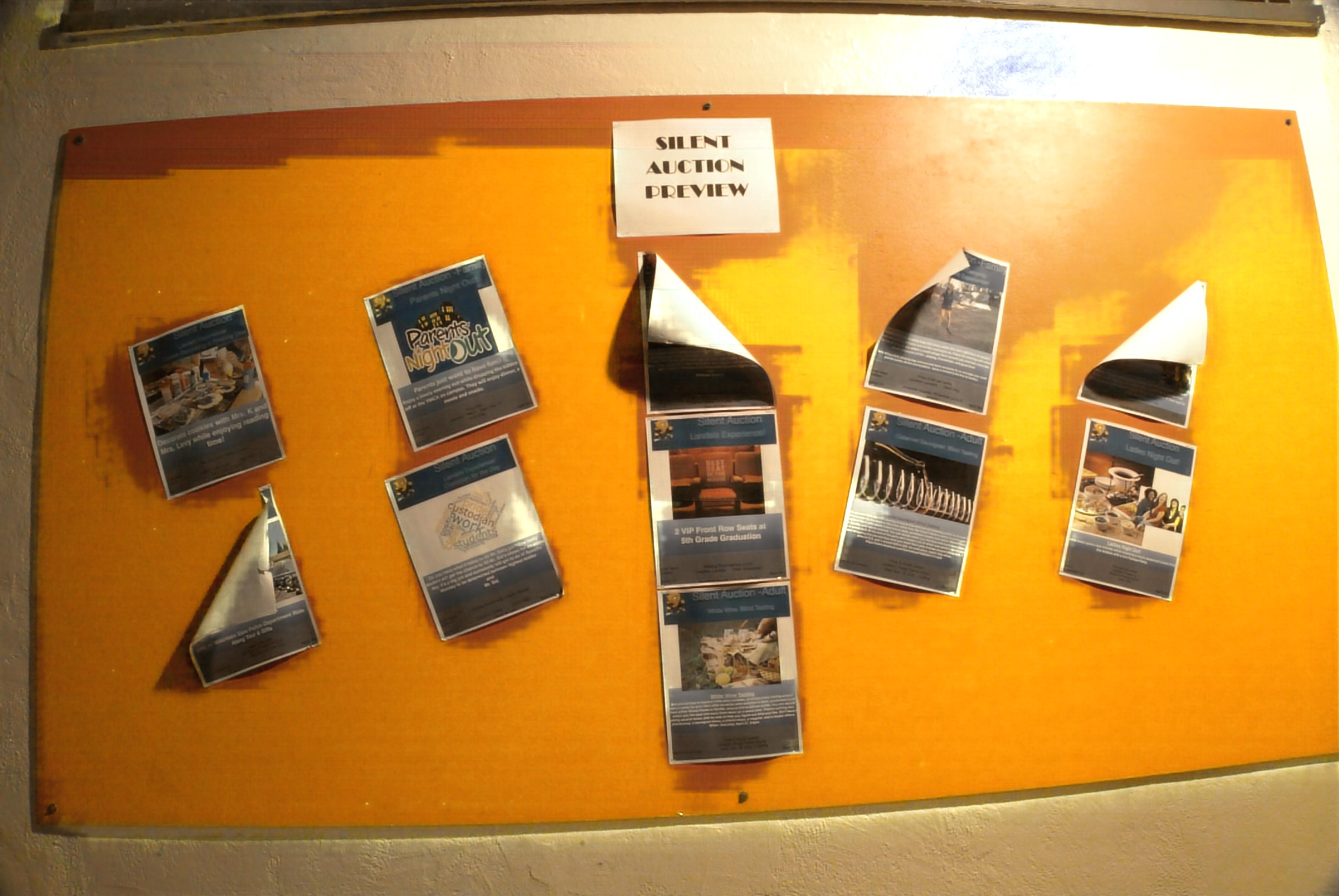}  & 
     \includegraphics[width=0.15 \linewidth]{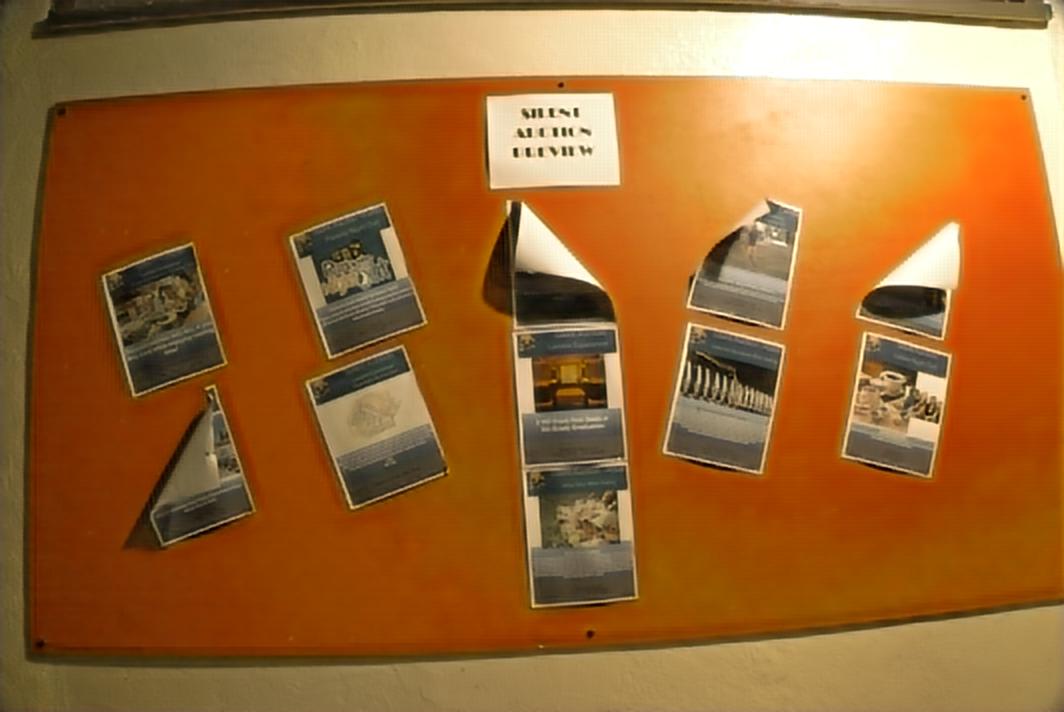} & \includegraphics[width=0.15 \linewidth]{ratio_boosting/48_IMG_GT.jpg}  \\   &  14.89/0.31 & 16.12/0.53 & 18.74/\textbf{0.88} & 14.94/0.71& \textbf{21.01/0.88} &   \\ 
     
       & 
     \includegraphics[width=0.15 \linewidth]{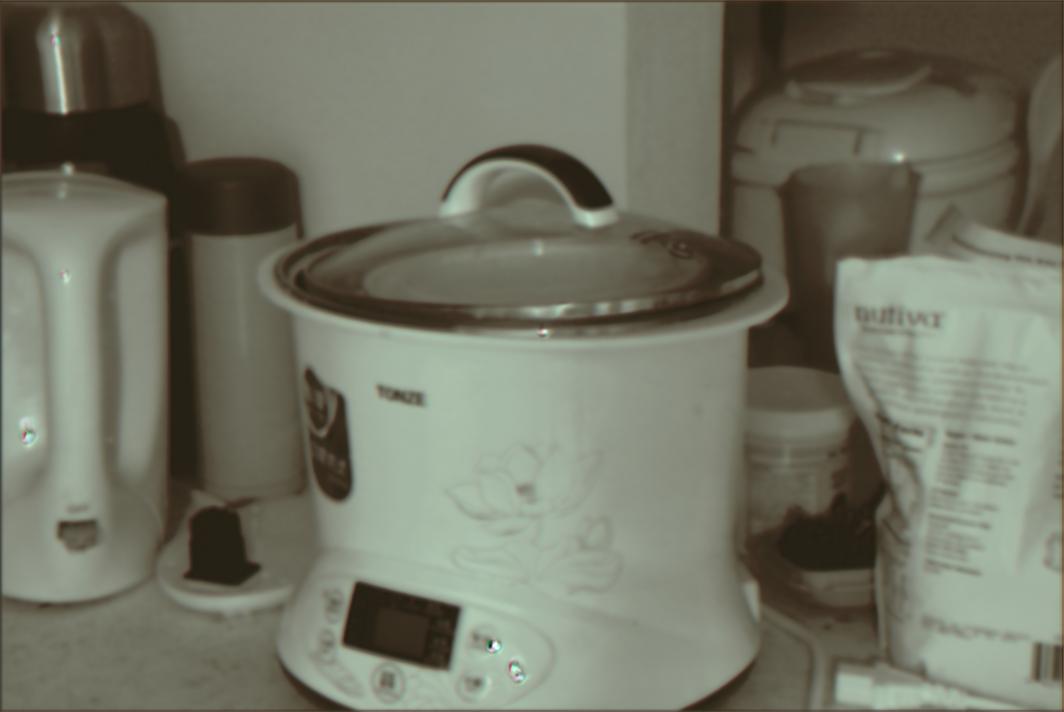}   &
    \includegraphics[width=0.15 \linewidth]{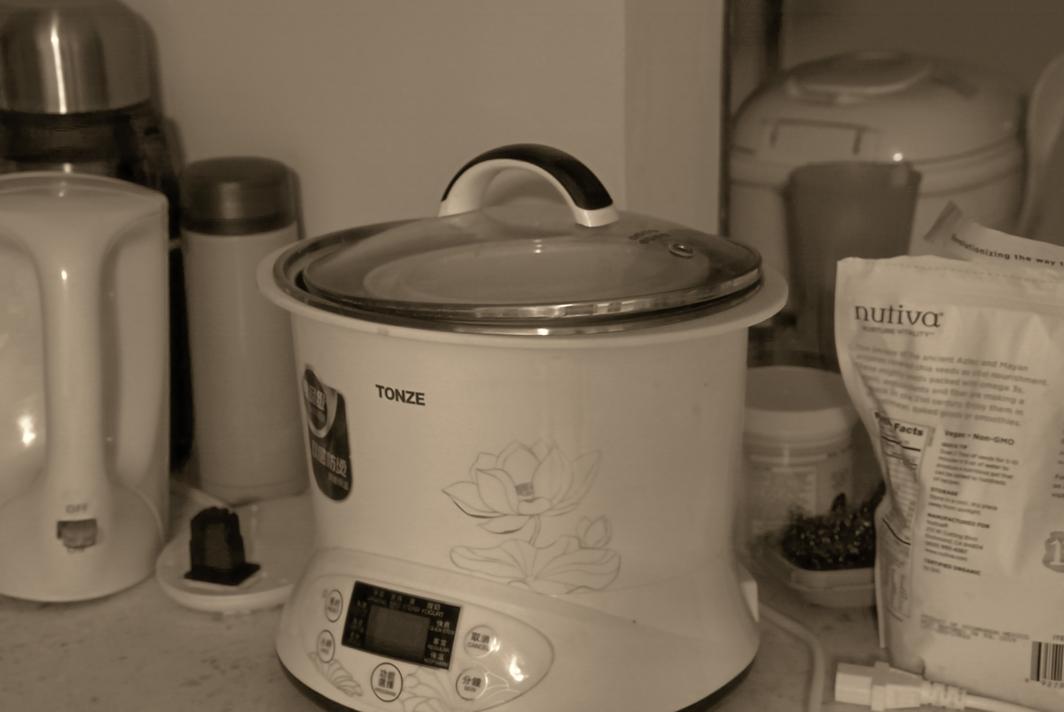} &
    \includegraphics[width=0.15 \linewidth]{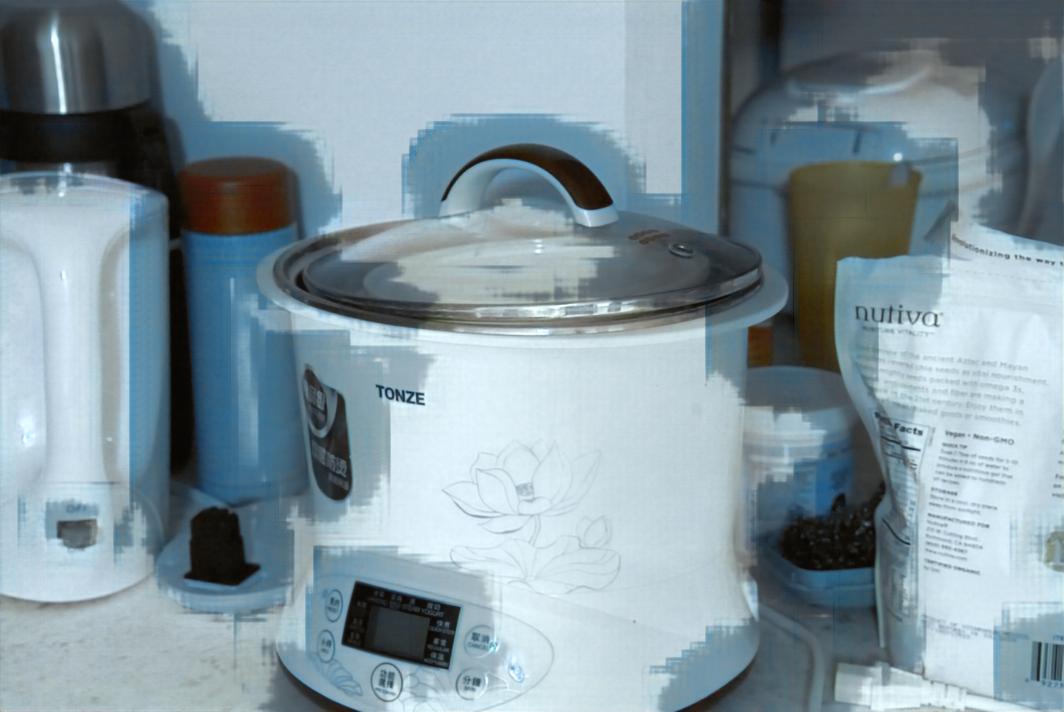} &
    \includegraphics[width=0.15 \linewidth]{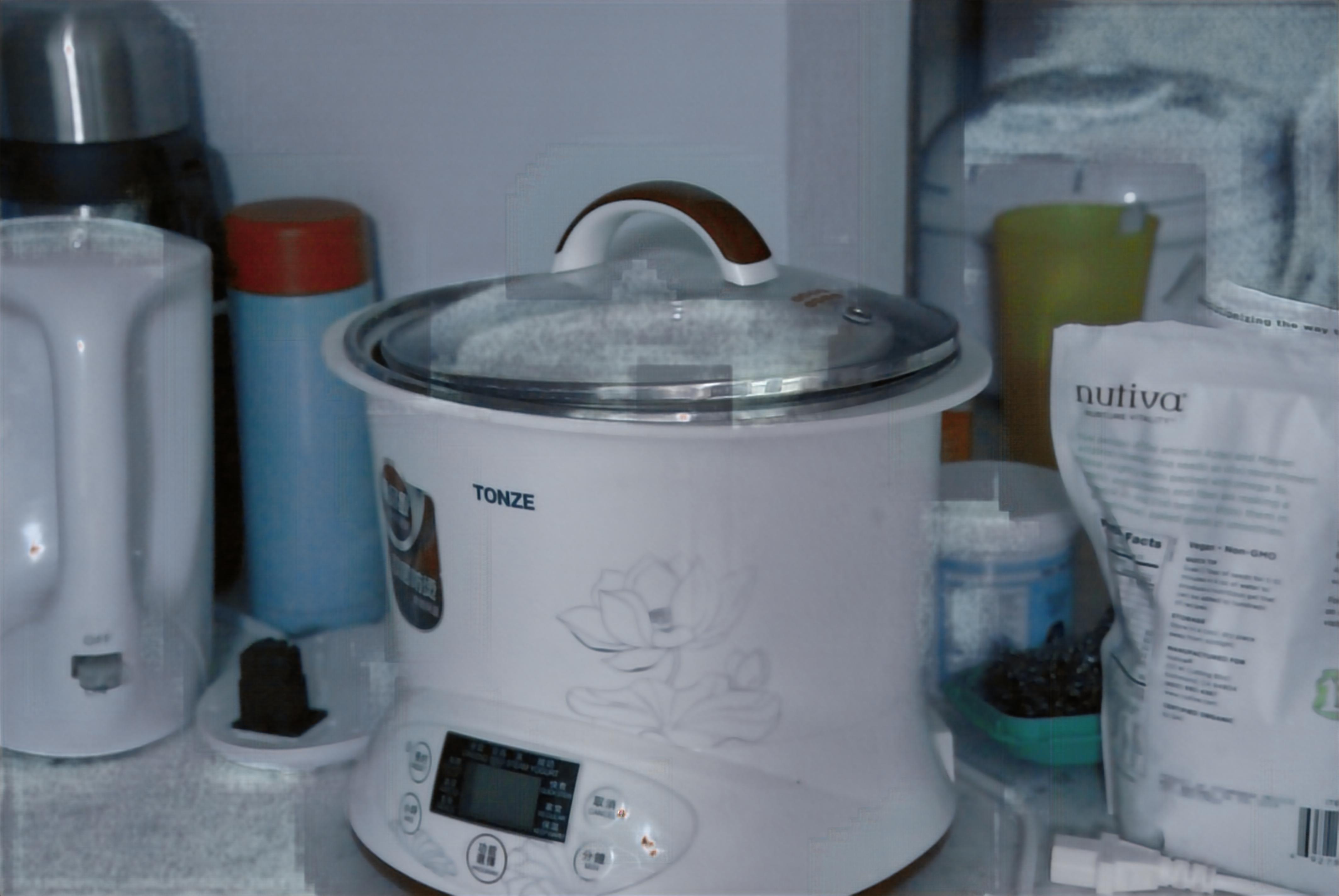}  & 
     \includegraphics[width=0.15 \linewidth]{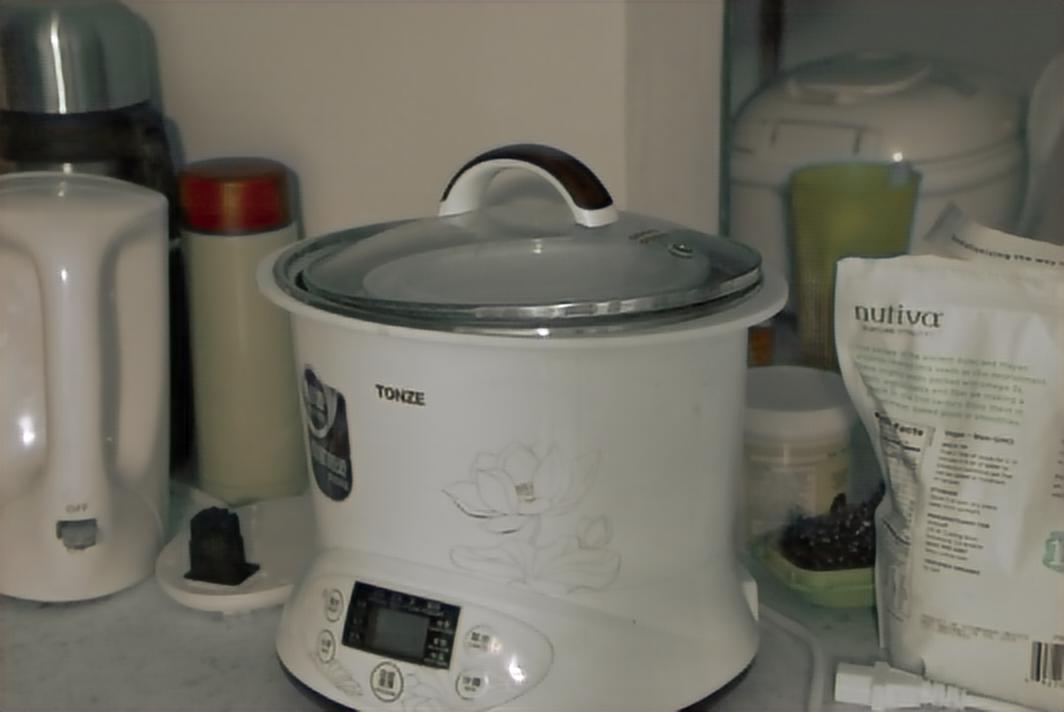} &  
     \includegraphics[width=0.15 \linewidth]{main/ours/31_IMG_GT.jpg} \\
       &  25.39/0.45 & 22.77/0.26 & 13.96/0.34 & 23.22/\textbf{0.59} & \textbf{27.87}/0.58 &  \\  
     
     
       & 
     \includegraphics[width=0.15 \linewidth]{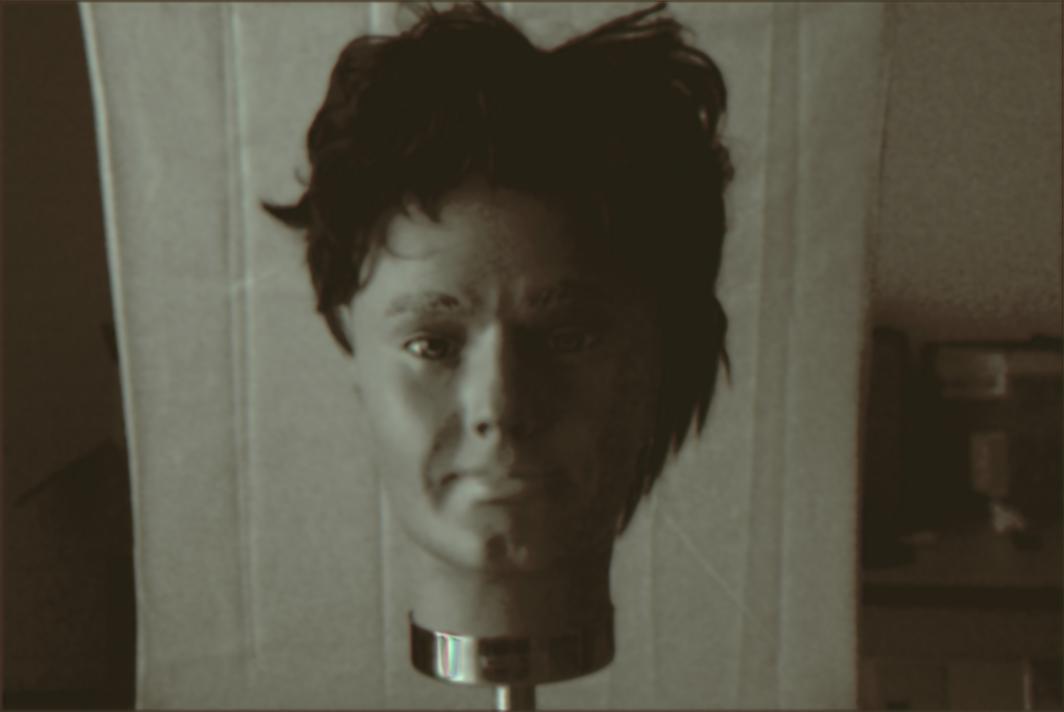}   &
    \includegraphics[width=0.15 \linewidth]{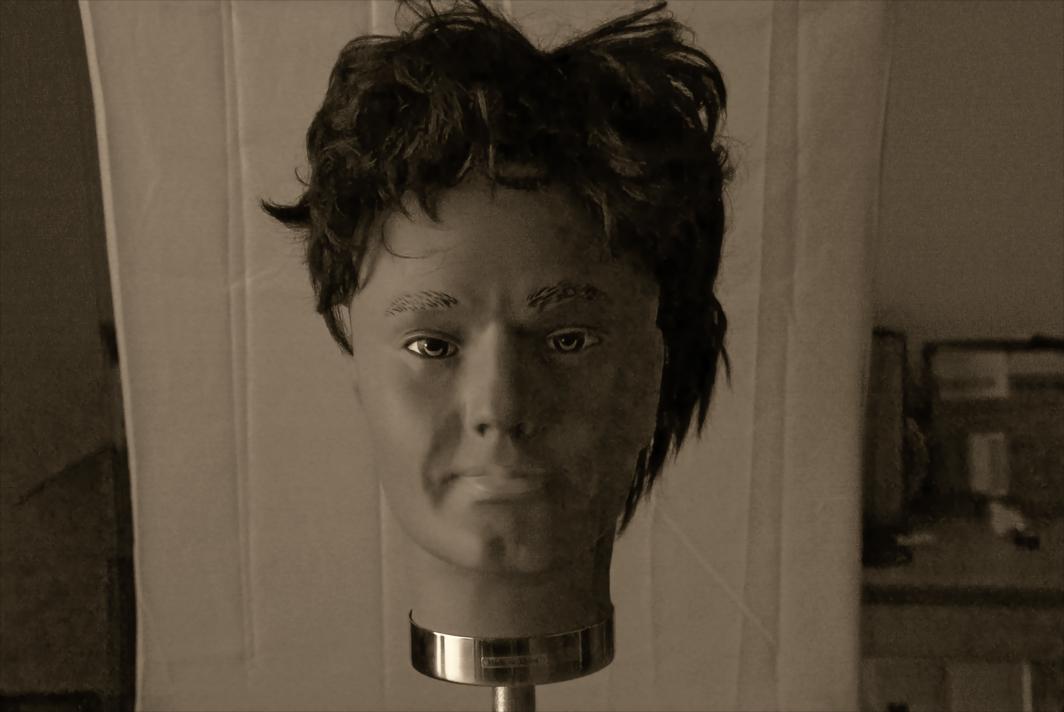} &
    \includegraphics[width=0.15 \linewidth]{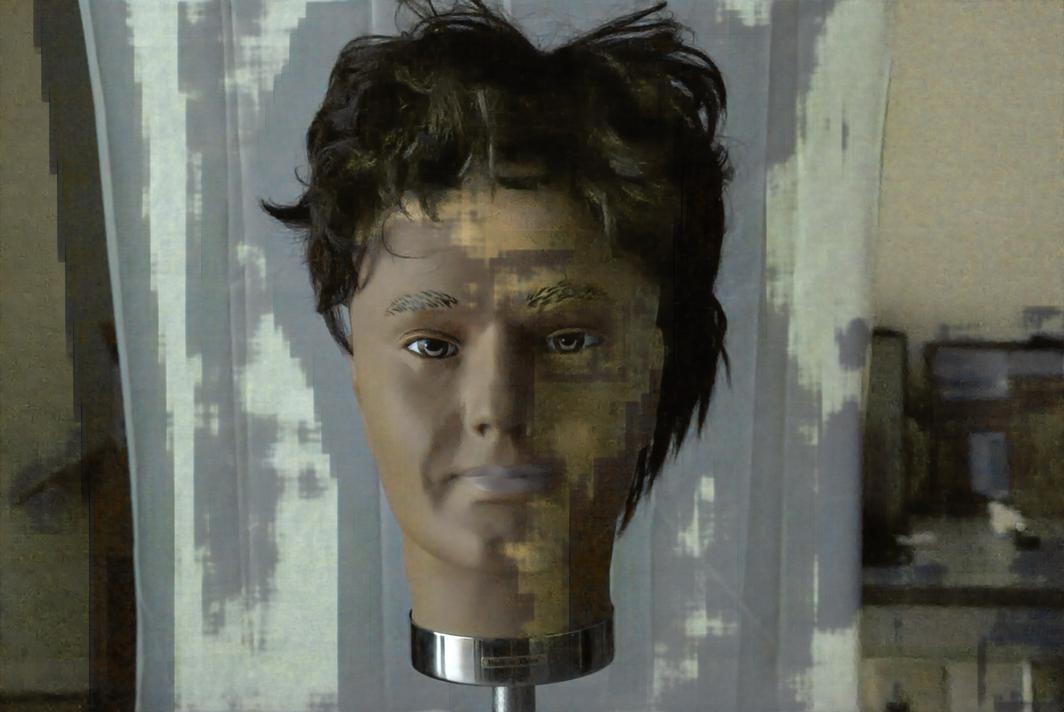} &
    \includegraphics[width=0.15 \linewidth]{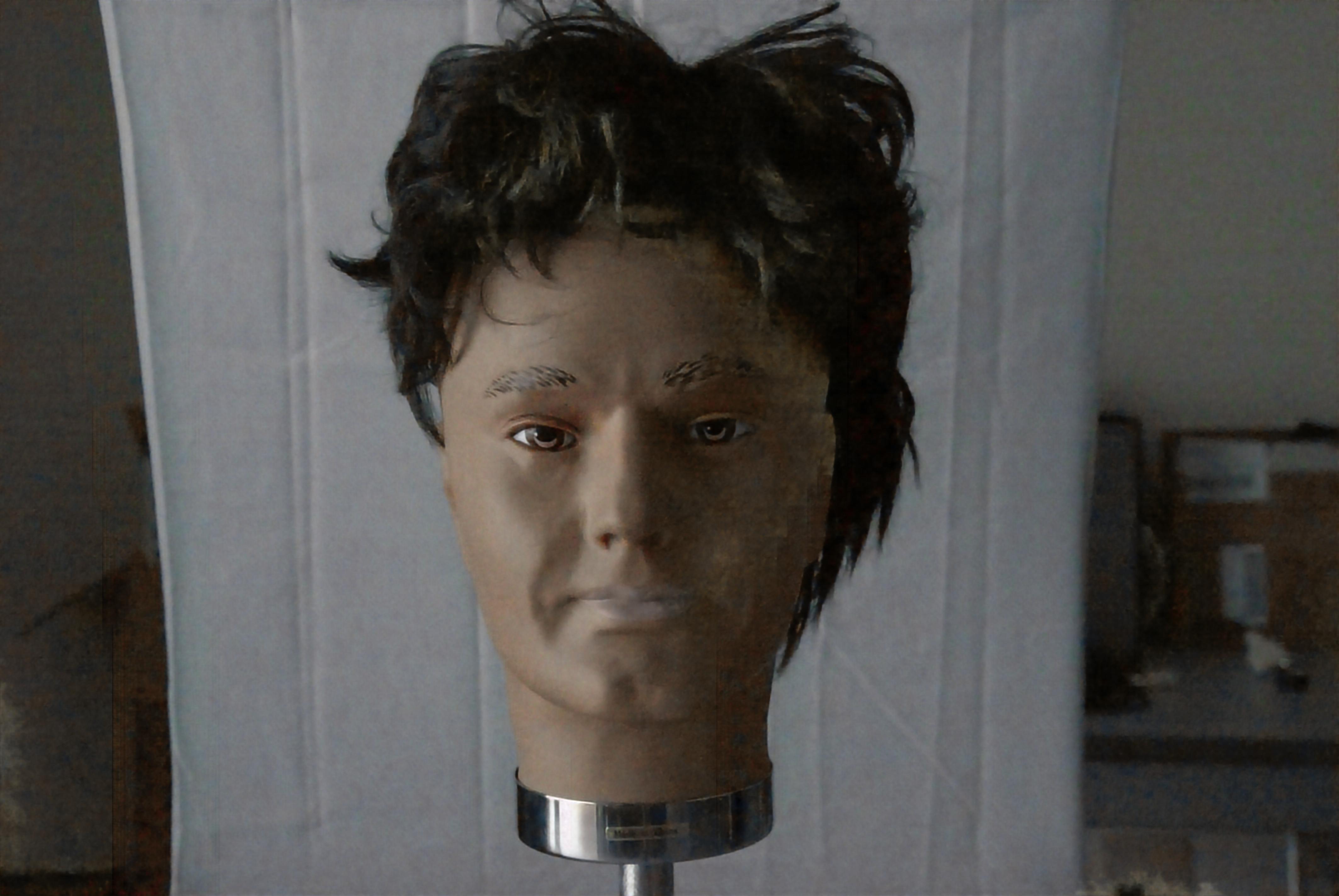}  & 
     \includegraphics[width=0.15 \linewidth]{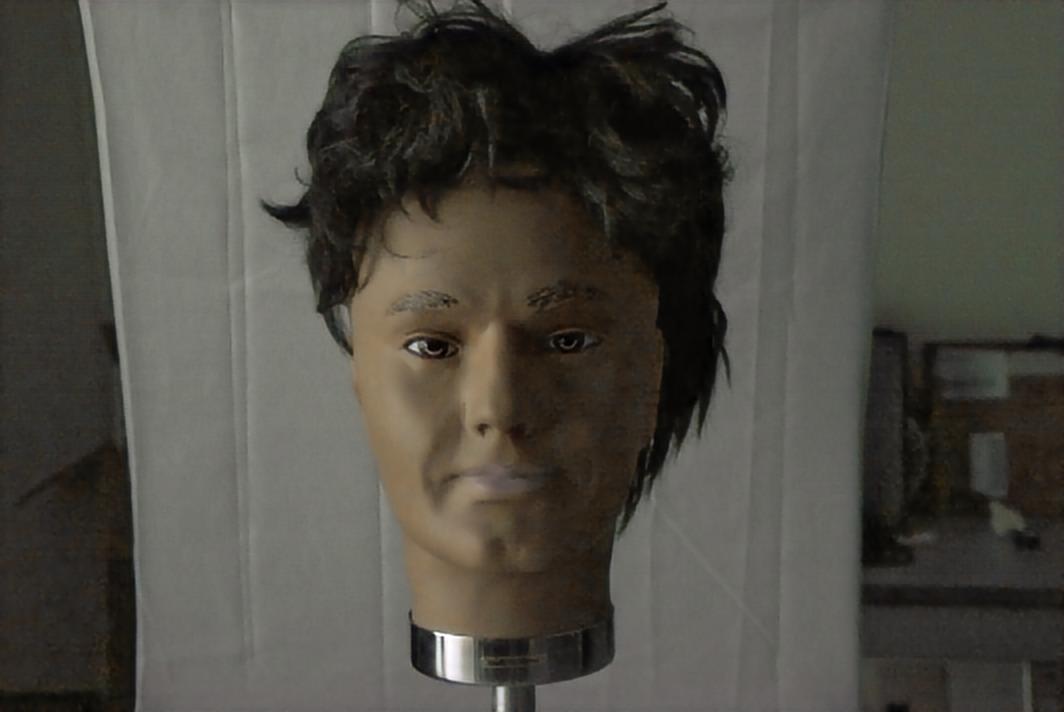} &  
     \includegraphics[width=0.15 \linewidth]{ratio_boosting/13_IMG_GT.jpg}  \\   &  25.13/0.46 & 24.11/0.41 & 18.36/\textbf{0.71} & \textbf{30.55/0.78} & 29.37/0.66 & 
    \end{tabular}  
    \end{tabular}
    \egroup
    \caption{[View with good screen brightness] Performance comparison of the proposed LLPackNet with state-of-the-art algorithms corresponding to Table \ref{tab:main_comparison}. \textbf{(A):} All the methods work well with GT exposure. \textbf{(B):} But in a realistic scenario, where exact GT exposure may not be available during inference, only the proposed LLPackNet gives proper restoration.}
\label{fig:main_visual_comparison}
\end{figure*}

We compare our network with Chen \etal \cite{chen2018learning}, Gu \etal \cite{gu2019self} and Maharjan \etal \cite{ICME19} on the SID dataset, see Table \ref{tab:main_comparison} and Fig. \ref{fig:main_visual_comparison}. These methods use the ratio of GT exposure to that of the input dark image, available in the SID dataset, to pre-amplify the images. The corresponding results are shown under the label `using GT exposure' in Table \ref{tab:main_comparison} and Fig. \ref{fig:main_visual_comparison}. But, since the GT information will not be readily available in a real-world setting, we additionally show results in the absence of GT information. This is shown under the heading `w/o GT exposure'. We also show results for `Chen \etal + Our Amplifier' in which our proposed amplifier is added to their algorithm. We have chosen Chen \etal because they have the least time and memory complexity, compared to the other existing methods. All the methods are appropriately retrained before evaluation.

\begin{figure*}[t!]
	
	\centering
	\scriptsize
	\setlength\tabcolsep{1pt}
	
    \begin{tabular}{ccc|ccc}
    
    {\scriptsize \textbf{PixelShuffle}} & {\scriptsize \textbf{Our UnPacking}} & {\scriptsize \textbf{GT}} & {\scriptsize \textbf{PixelShuffle}} & {\scriptsize \textbf{Our UnPacking}} & {\scriptsize \textbf{GT}}\\
    \includegraphics[width=0.16 \linewidth]{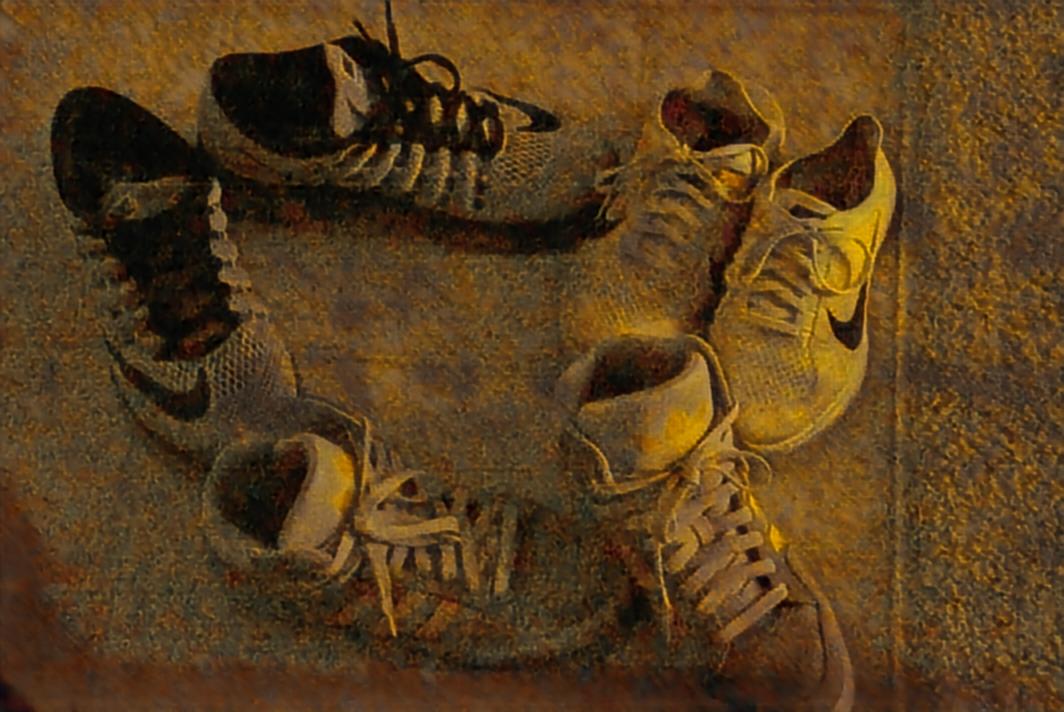}    & \includegraphics[width=0.16 \linewidth]{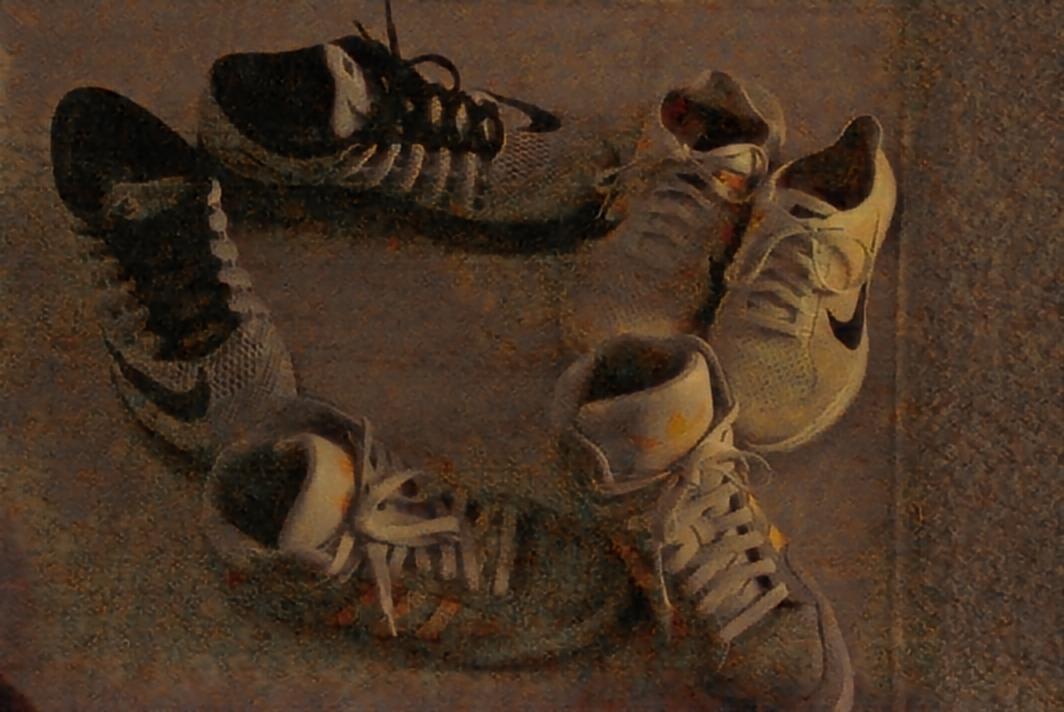}  &  
     \includegraphics[width=0.16 \linewidth]{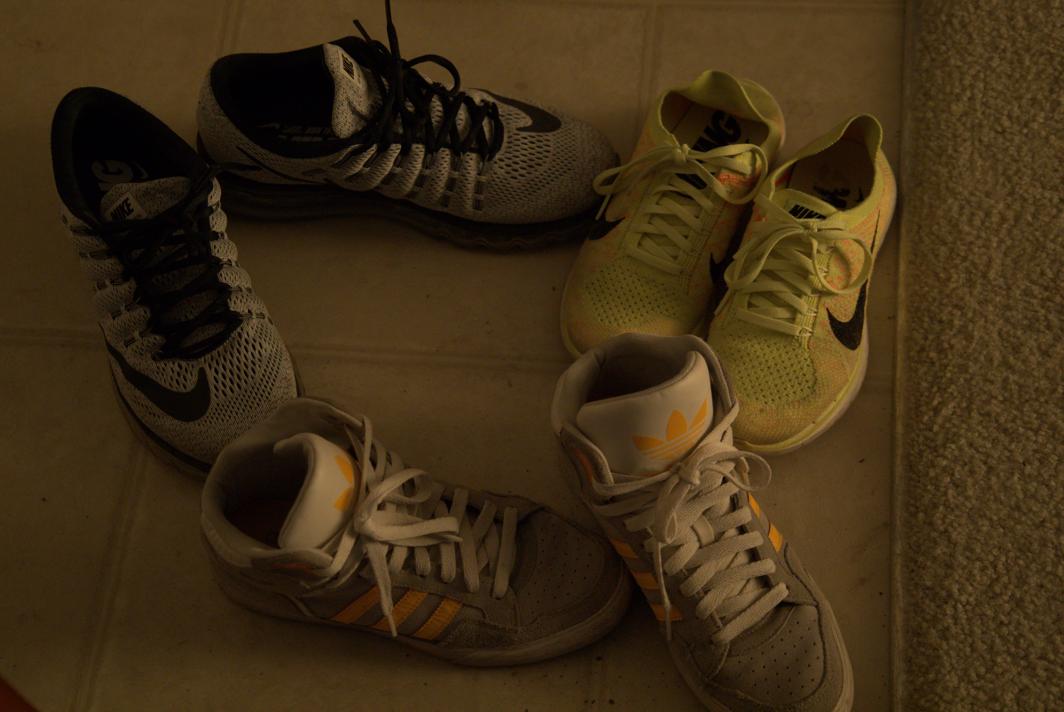} & 
     \includegraphics[width=0.16 \linewidth]{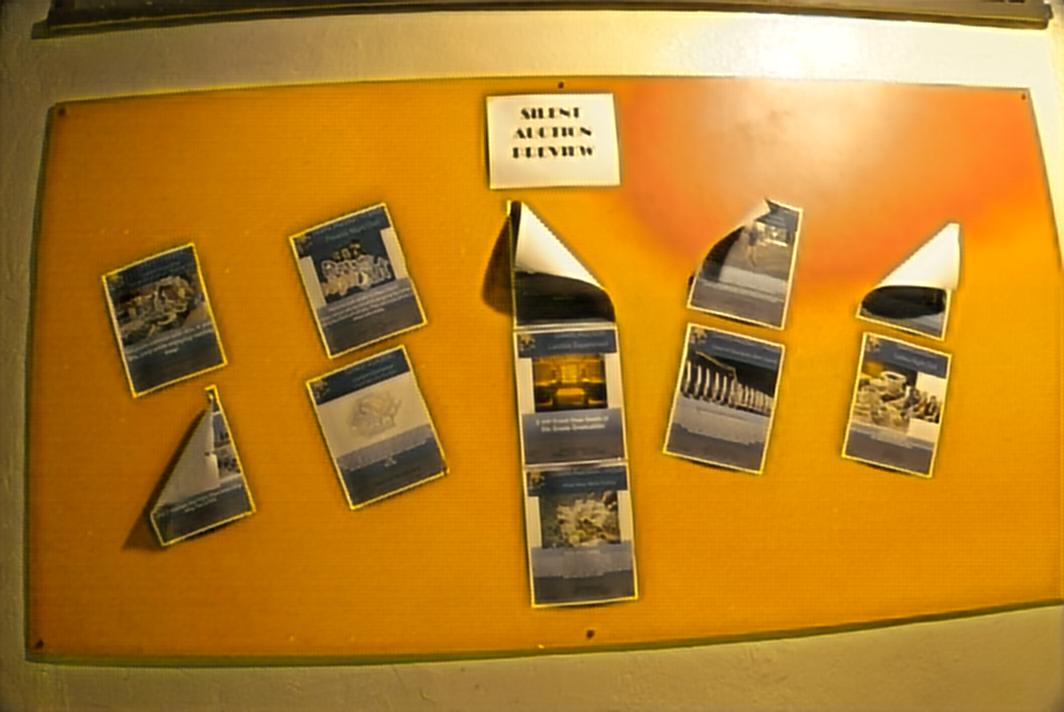}    & \includegraphics[width=0.16 \linewidth]{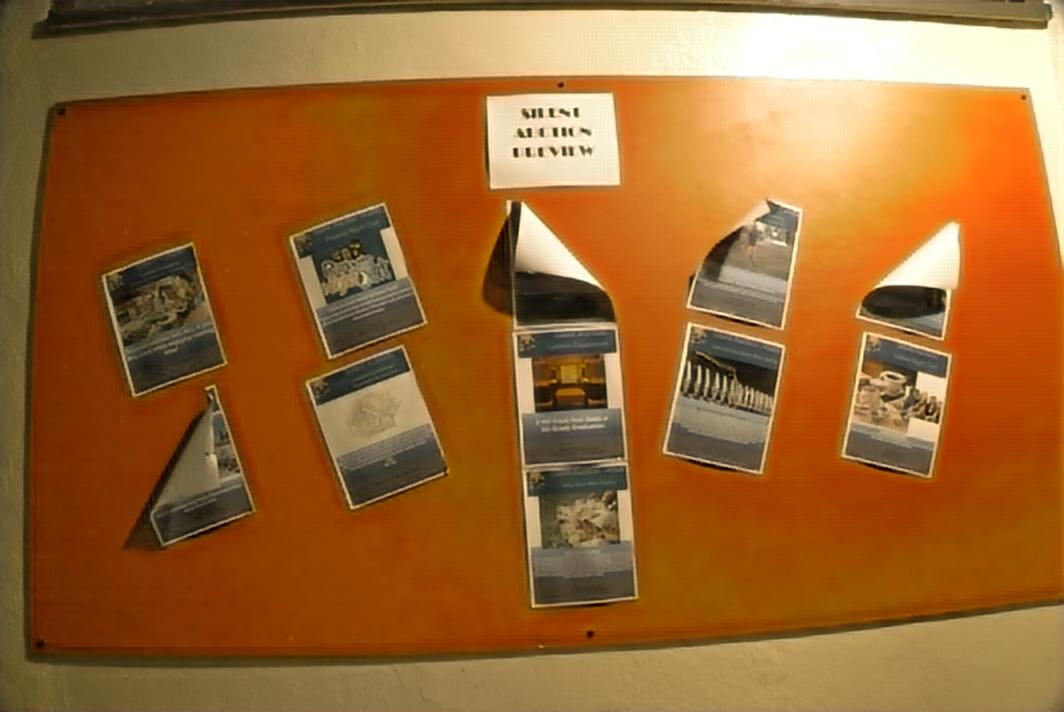}  &  
     \includegraphics[width=0.16 \linewidth]{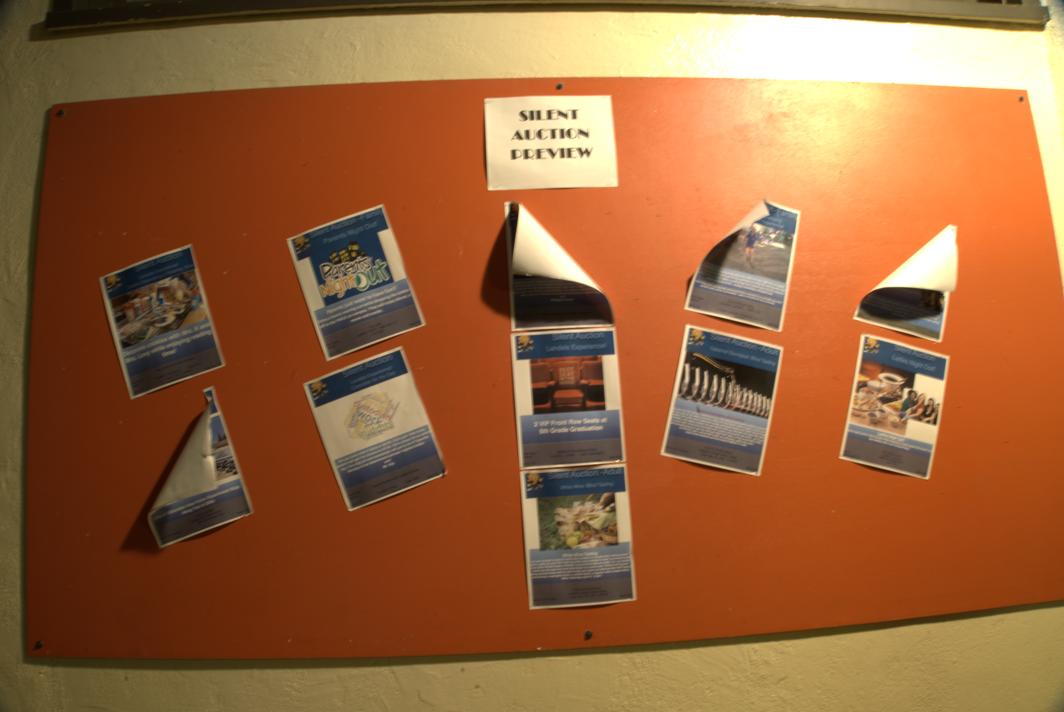}\\
     21.13/0.77 & \textbf{24.56}/\textbf{0.80} & & 15.85/0.79 & \textbf{21.01}/\textbf{0.88}\\
    \includegraphics[width=0.16 \linewidth]{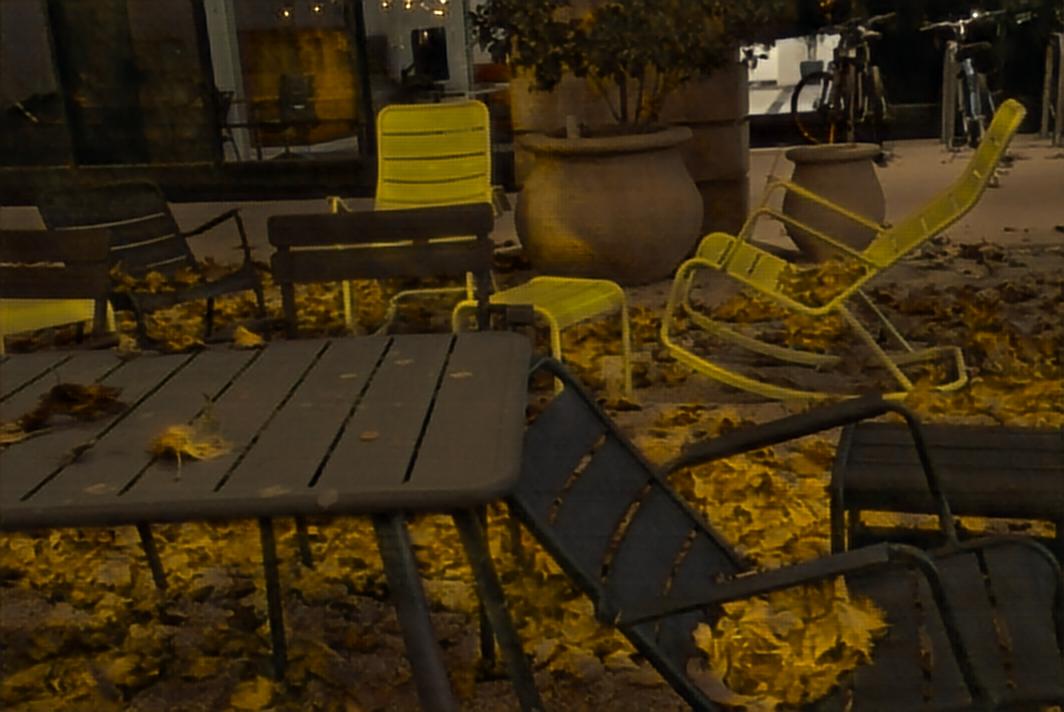}   &  \includegraphics[width=0.16 \linewidth]{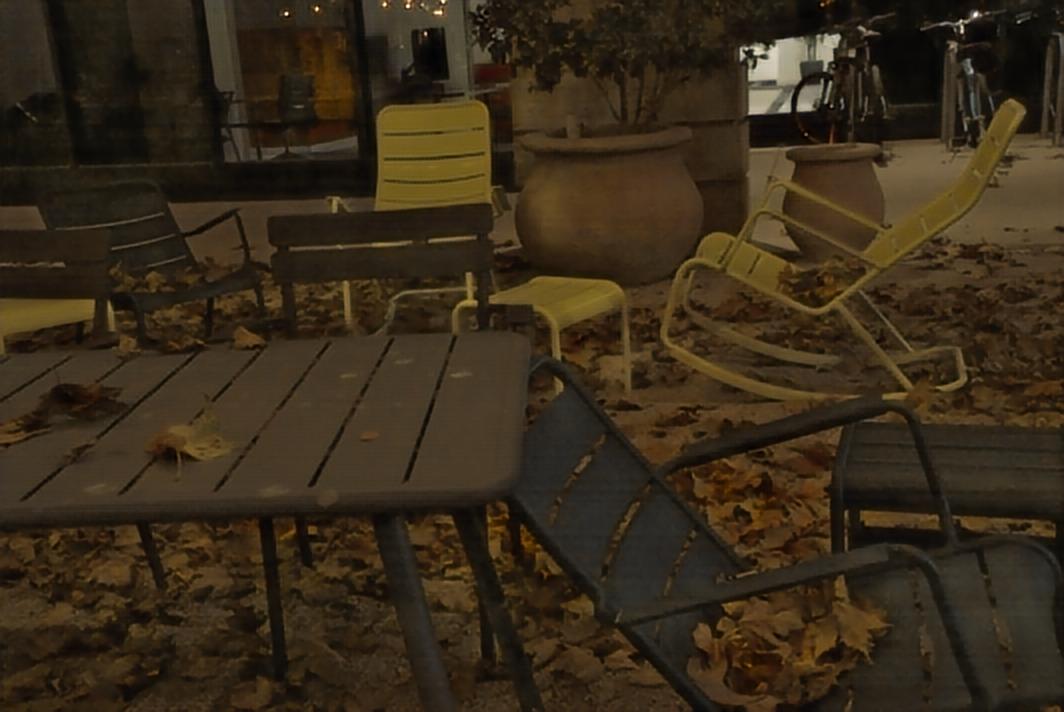} &  
     \includegraphics[width=0.16 \linewidth]{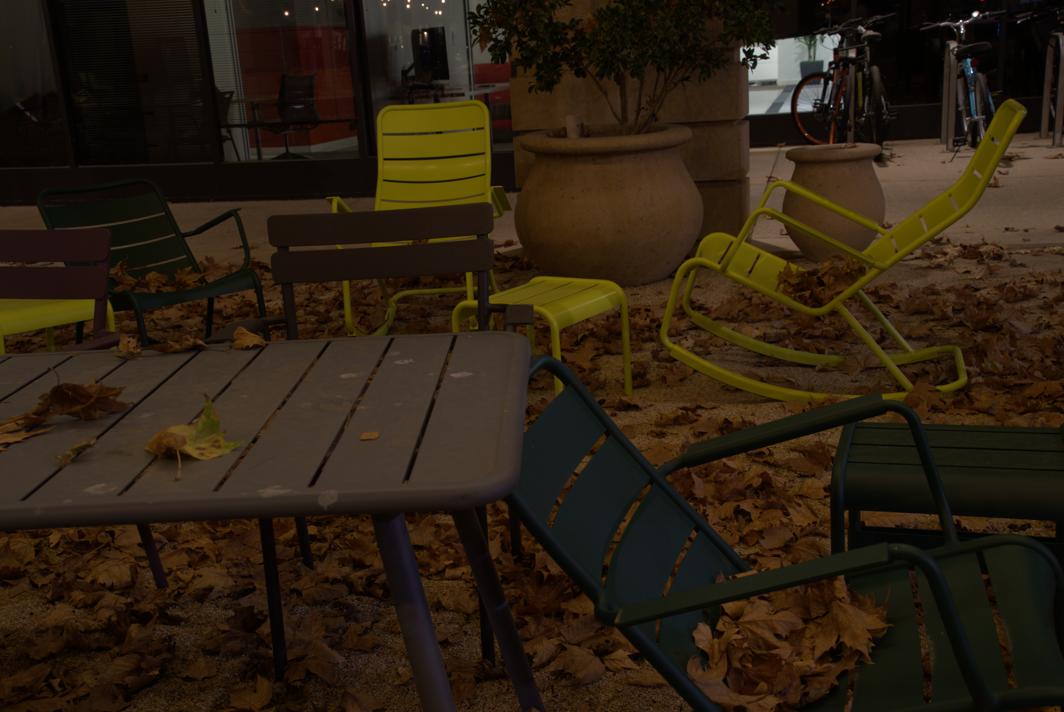} &
     \includegraphics[width=0.16 \linewidth]{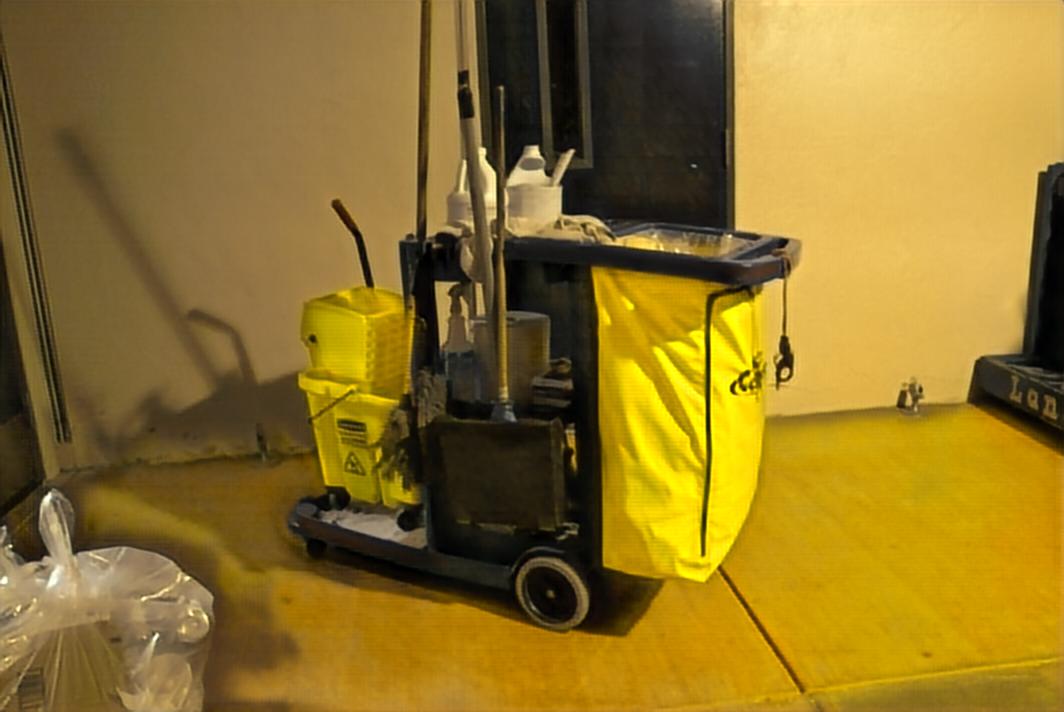}   & \includegraphics[width=0.16 \linewidth]{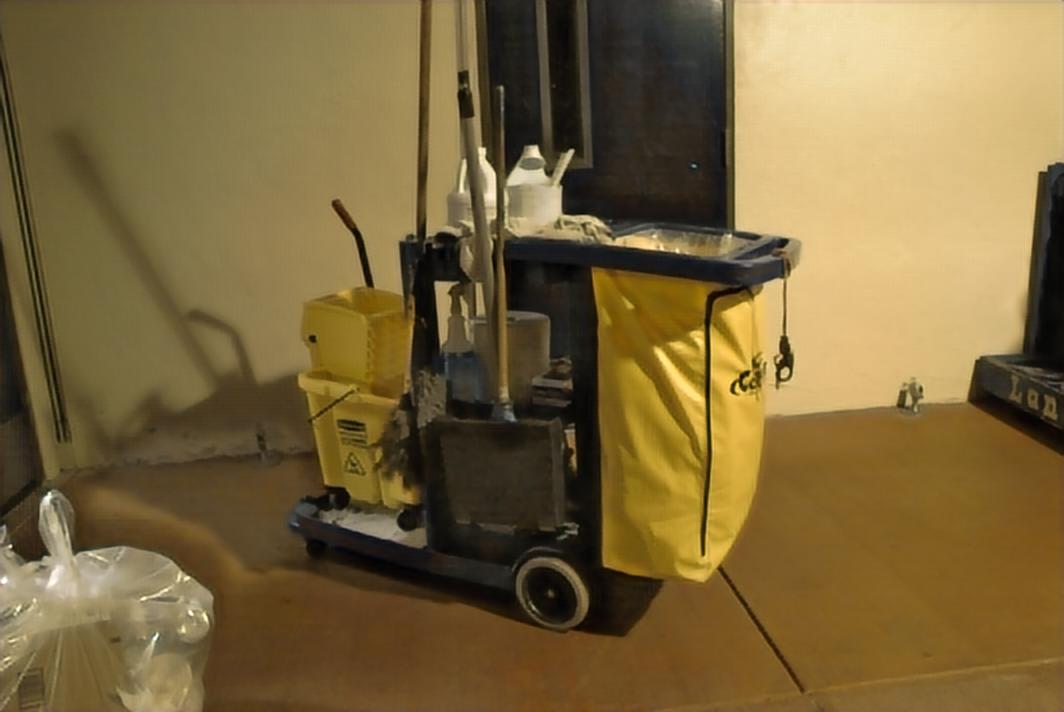} &  
     \includegraphics[width=0.16 \linewidth]{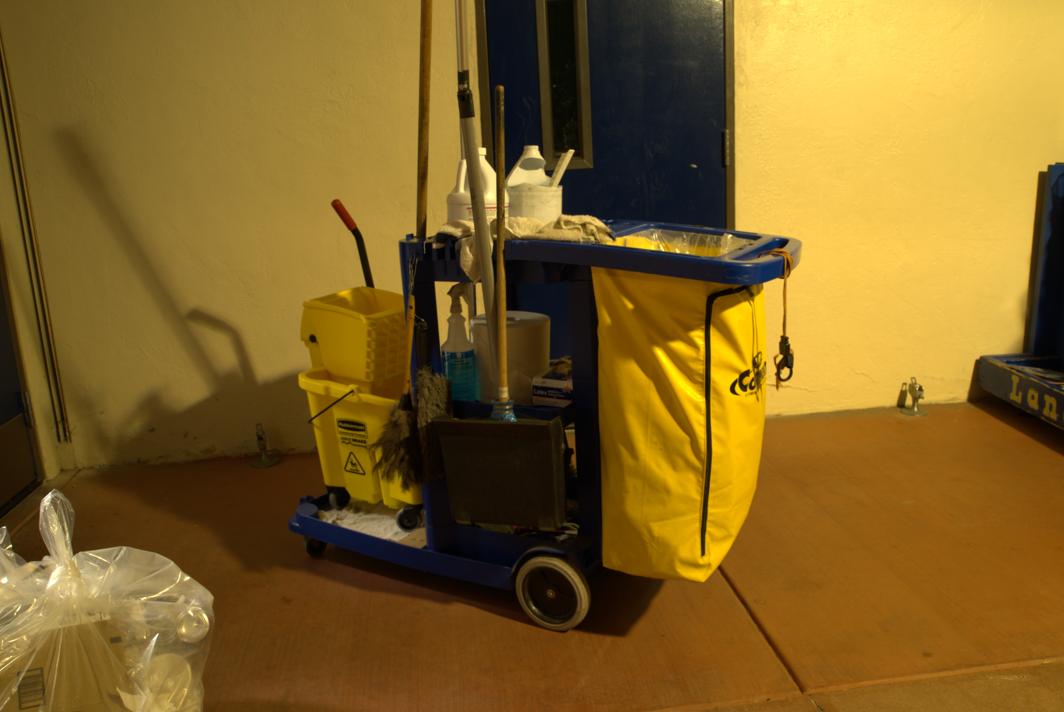}\\
     23.82/\textbf{0.80}& \textbf{25.14}/0.79& & 19.07/\textbf{0.84} & \textcolor{black}{\textbf{21.82}}/0.83
      
    \end{tabular}
    \caption{[View with good screen brightness] LLPackNet restoration using PixelShuffle and our proposed UnPack operation. The UnPack operation helps achieve better color restoration by containing color distortions such as color cast.}
\label{fig:packing_visual_comparison}

\end{figure*}

\begin{figure*}[ht!]
	
	\centering
	\scriptsize
	\setlength\tabcolsep{1pt}
    \begin{tabular}{ccc|ccc}
    
    {\scriptsize \textbf{No Amplifier}} & {\scriptsize \textbf{With Amplifier}} & {\scriptsize \textbf{GT}} & {\scriptsize \textbf{No Amplifier}} & {\scriptsize \textbf{With Amplifier}} & {\scriptsize \textbf{GT}}\\
    \includegraphics[width=0.16 \linewidth]{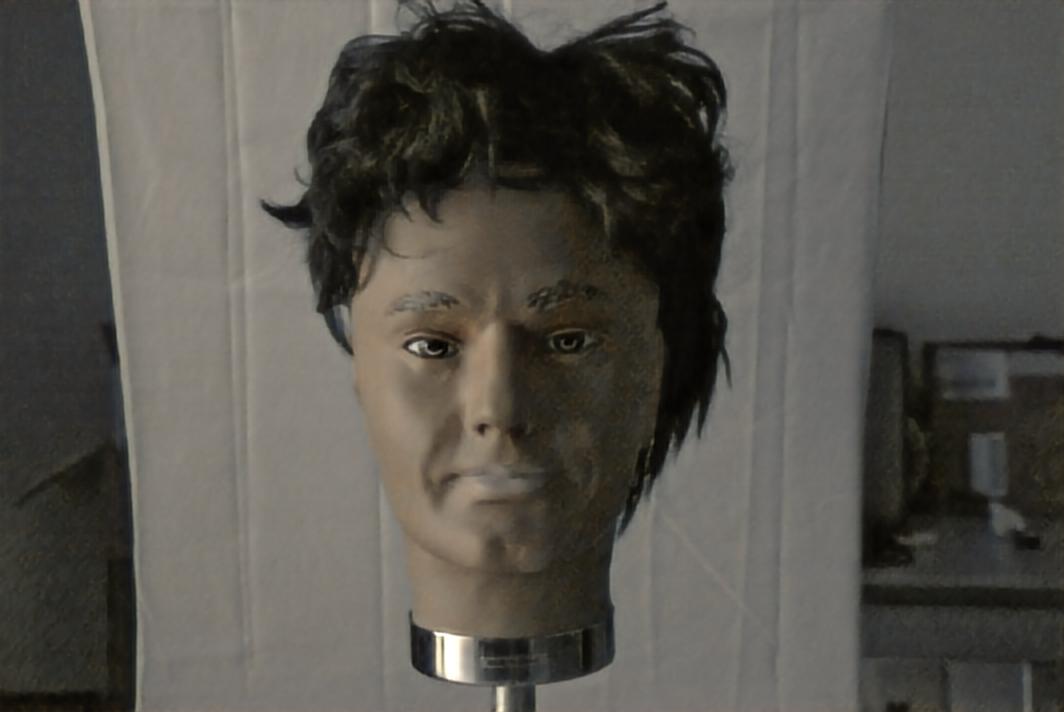}    & \includegraphics[width=0.16 \linewidth]{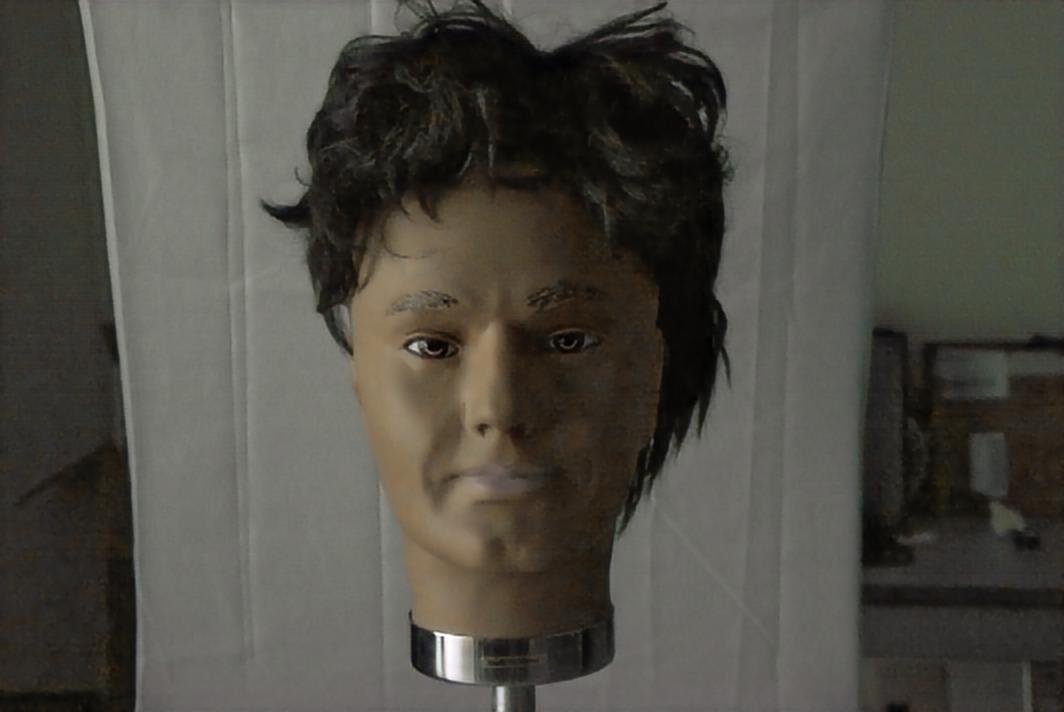} &  
     \includegraphics[width=0.16 \linewidth]{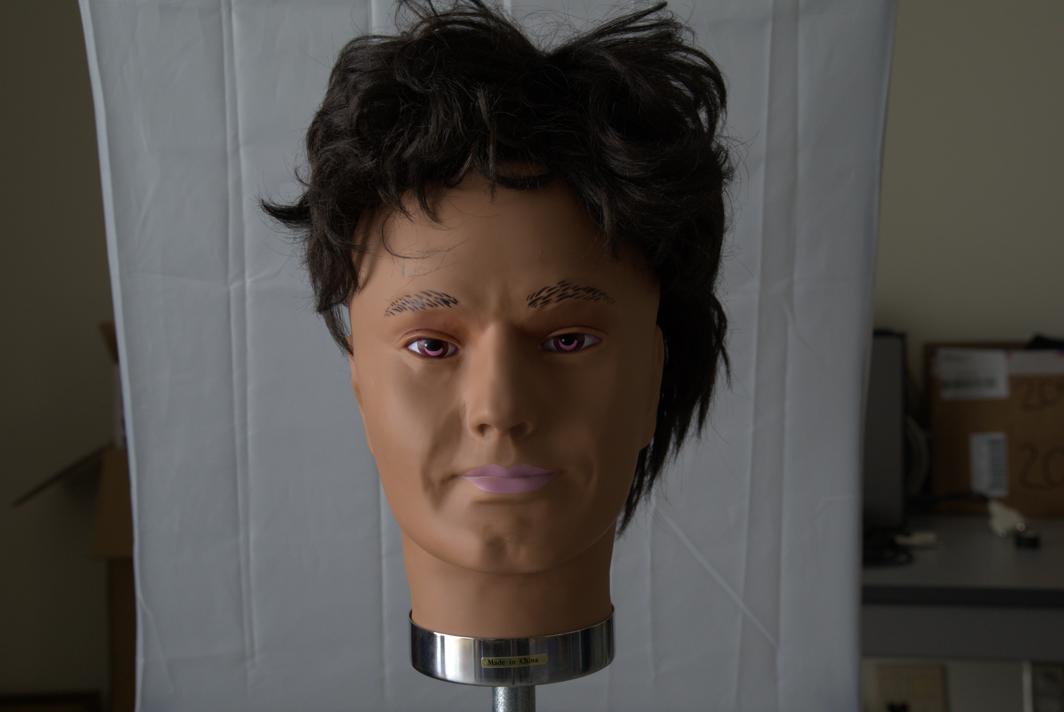} & 
     \includegraphics[width=0.16 \linewidth]{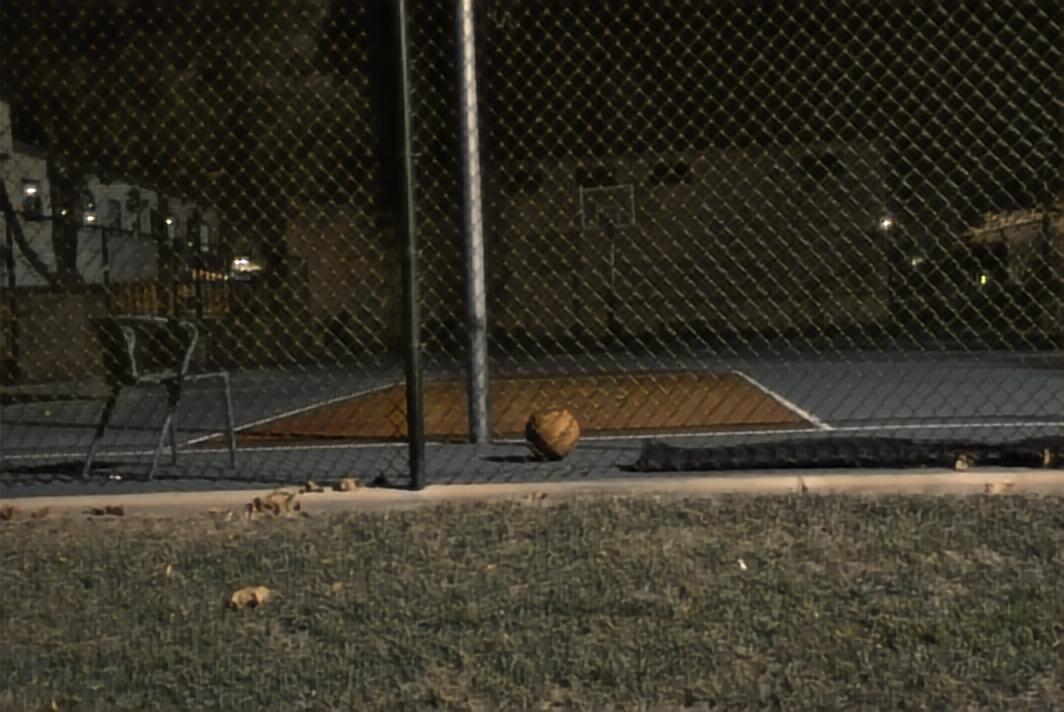}   & \includegraphics[width=0.16 \linewidth]{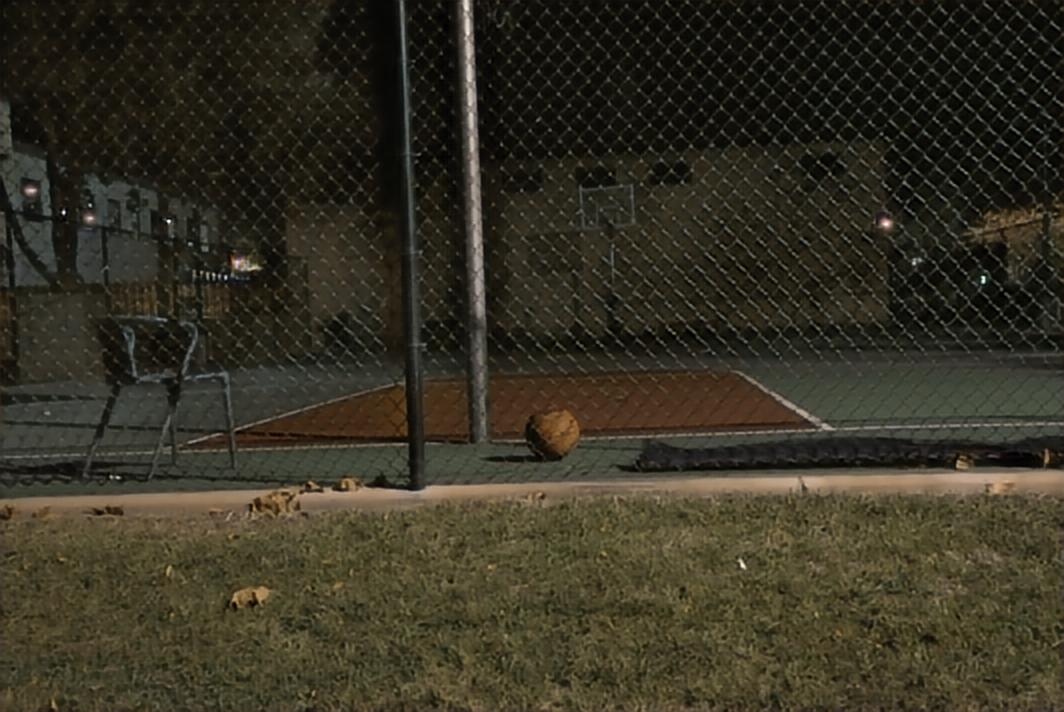} &  
     \includegraphics[width=0.16 \linewidth]{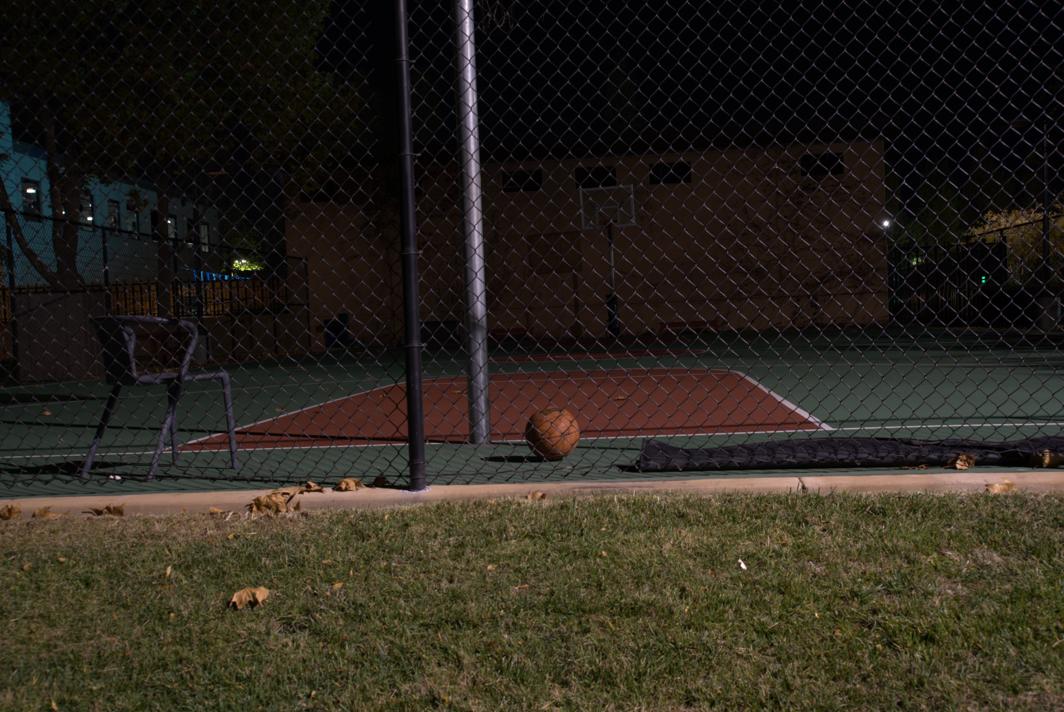}\\
     27.17/0.57 & \textbf{29.37}/\textbf{0.66} & & 23.78/0.60 & \textbf{27.30}/\textbf{0.64} \\
     
     \includegraphics[width=0.16 \linewidth]{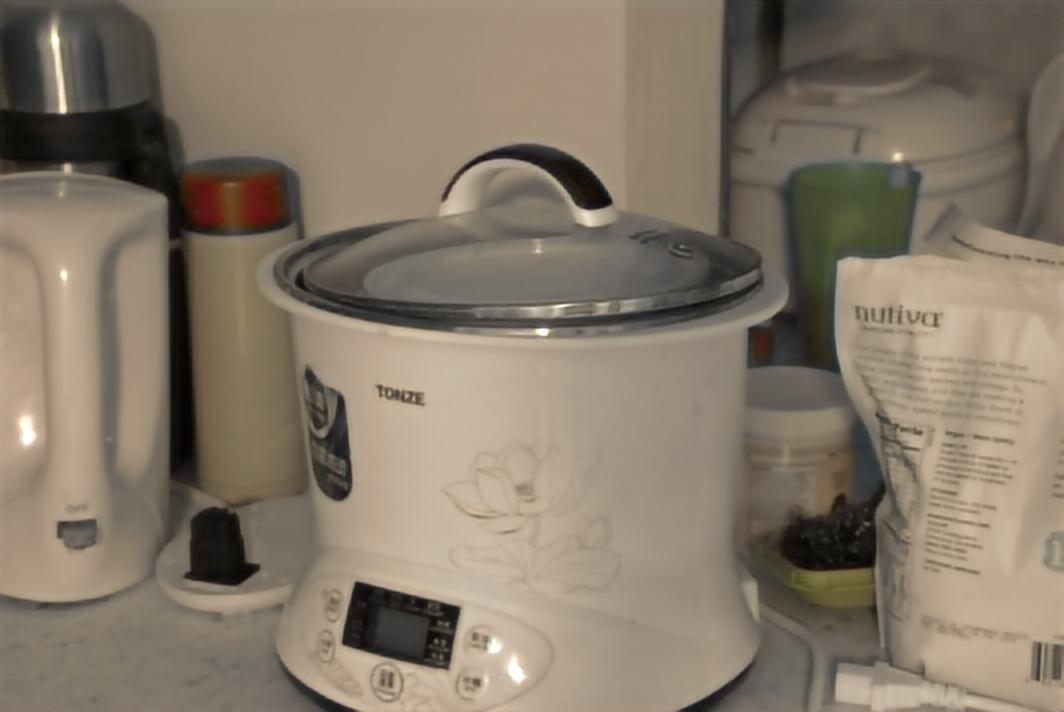}   & \includegraphics[width=0.16 \linewidth]{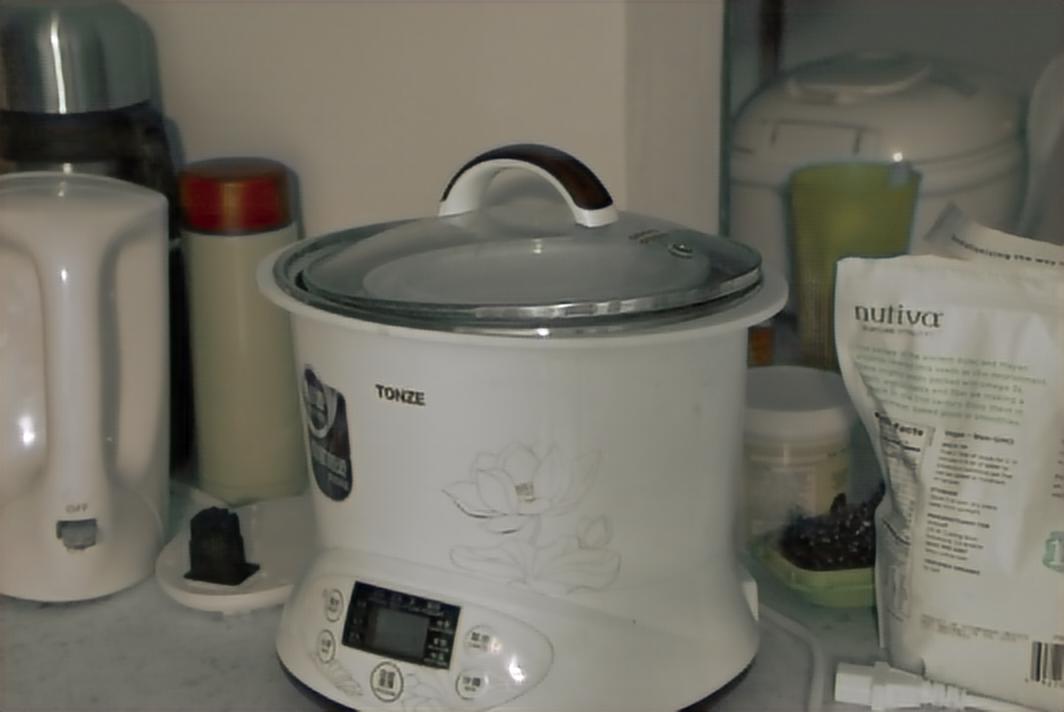} &  
     \includegraphics[width=0.16 \linewidth]{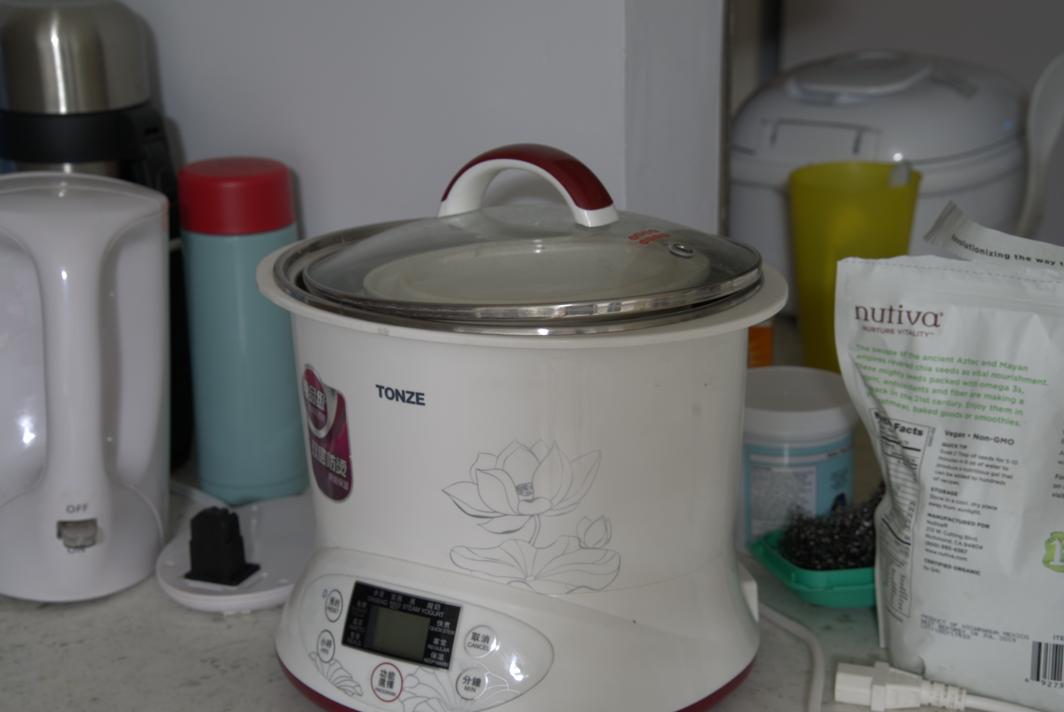} & 
     \includegraphics[width=0.16 \linewidth]{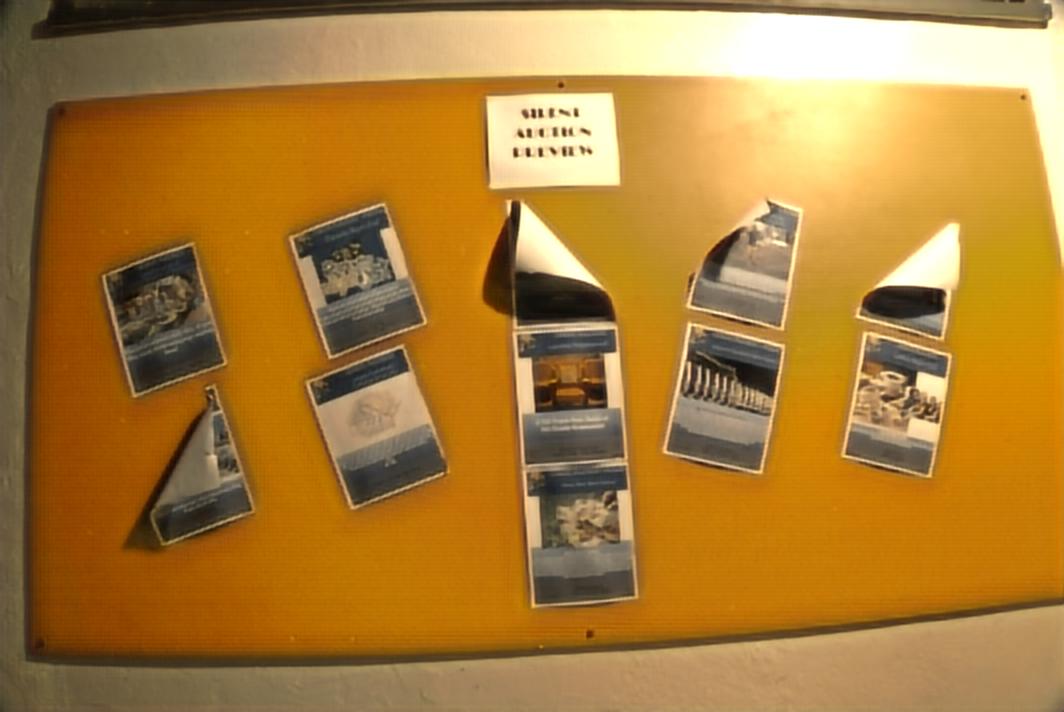}   & \includegraphics[width=0.16 \linewidth]{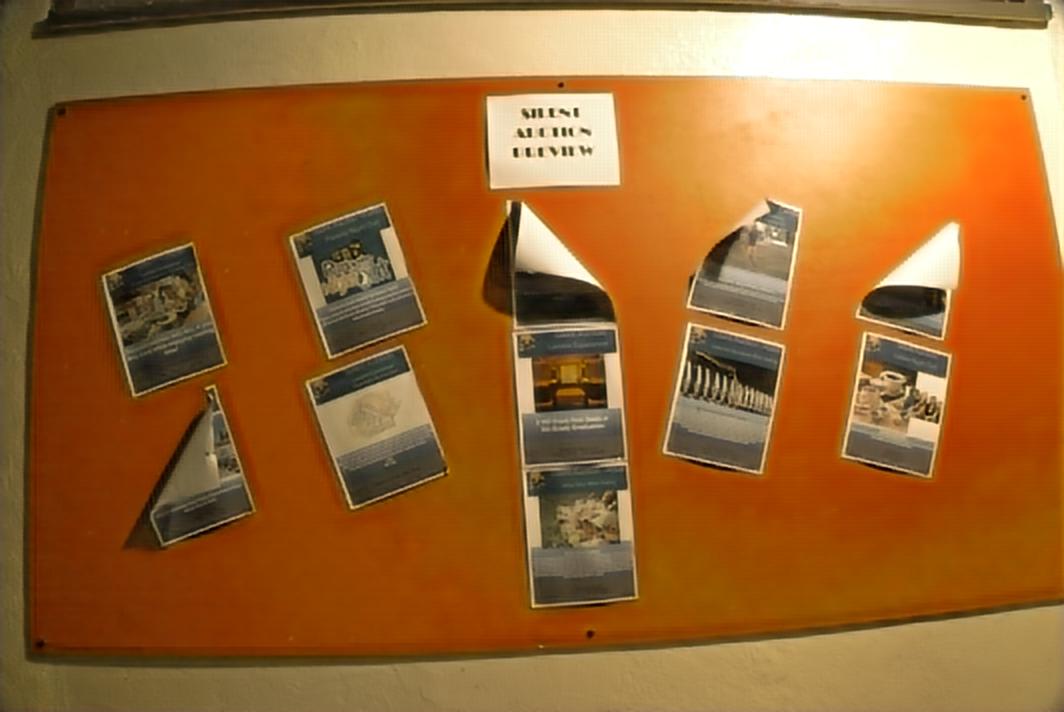} &  
     \includegraphics[width=0.16 \linewidth]{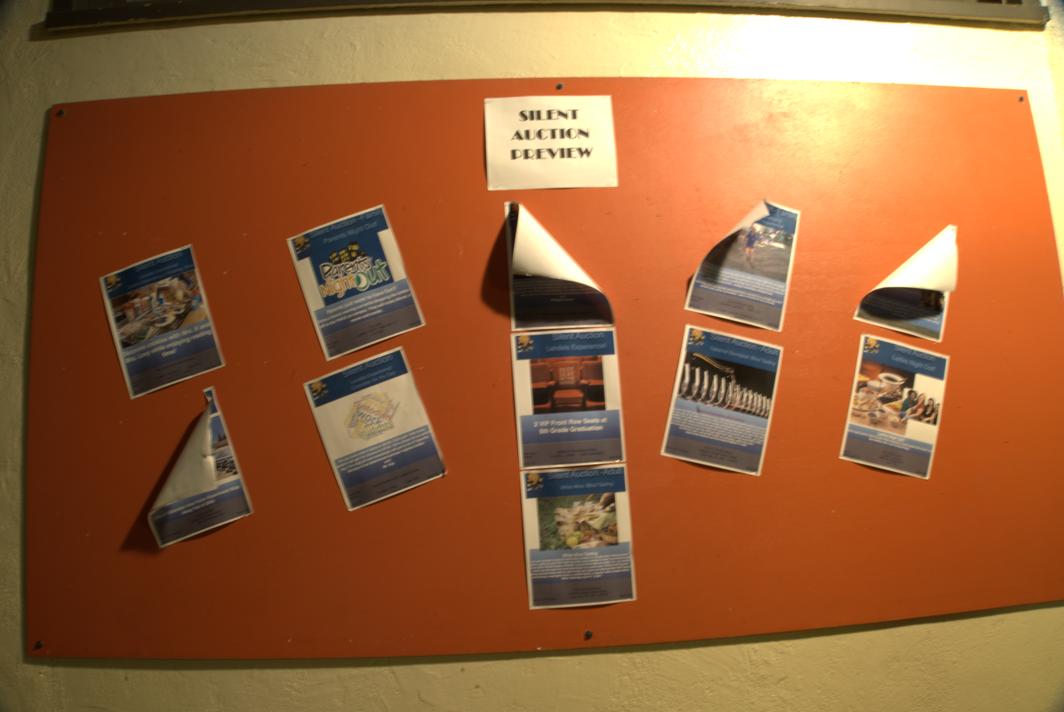}\\
     23.21/0.34 & \textbf{27.87}/\textbf{0.58} & &17.86/0.85 & \textbf{21.01}/\textbf{0.88}
      
    \end{tabular}
    \caption{[View with good screen brightness] Results using the proposed LLPackNet with and without amplification estimation. Without amplification, the colors are pale and tend to be monochromatic. }
\label{fig:amplifier_visual_comparison}
\end{figure*}

\textbf{Network speed and memory utilization: }
As shown in Table \ref{tab:main_comparison}, LLPackNet is $5-20 \times$ faster with $2-3 \times$ lower memory and $2-7 \times$ lesser model parameters.
We achieve this because we do the bulk of operations in 16$\times$ lower resolution. In contrast, Maharjan \etal \cite{ICME19} do not perform any downsampling operation and therefore, the feature maps propagating through their network are huge. This results in very high network latency and memory consumption. Gu \etal \cite{gu2019self} adopt a multi-scale approach that requires feature map propagation at 2$\times$ and 4$\times$ lower resolution. But this marginal downsampling is not sufficient to contain the network latency and memory consumption. Chen \etal \cite{chen2018learning} have relatively better metrics by performing up to 32$\times$ downsampling. But this is done only in steps of 2, requiring five downsampling and five upsampling operations. Further, four out of five upsampling operations are done using transposed convolution \cite{2016DeepLearningGuide}, which is much slower than the proposed UnPack operation. Thus, Chen \etal have a moderately high processing time and memory utilization. Check the supplementary for more details. 

\textbf{Restoration quality: }
 All methods perform notably well when the GT exposure is available. But in a practical setting when GT exposure is not readily available, except for our LLPackNet, the other methods struggle to restore proper colors.
  The results for this practical setting are also shown in Fig \ref{fig:abstract}.
Adding our amplifier module to Chen \etal improves their performance to some extent, but the restored images still exhibit noisy patches and color cast. This is because amplification is not the only deciding factor in improving the performance of a network. Rather, having a large receptive field, which provides more contextual information, and better correlation among the color channels is more important than the correct amplification factor. To further assess these claims, refer to the ablation studies in section \ref{sec:ablation}, which show that LLPackNet continues to give structurally consistent results even when the amplifier is removed. 
 
 \subsection{Ablation studies on LLPackNet}
 \label{sec:ablation}
 We now show ablation studies on LLPackNet to better understand the contribution of individual components. For each ablation study the network is appropriately retrained. 

 \textbf{UnPack $vs.\ $ PixelShuffle: }As a first ablation study, we replace the UnPack operation in the proposed LLPackNet with the PixelShuffle operation \cite{pixelshuffle} and the results are shown in Fig. \ref{fig:packing_visual_comparison}. We notice that the images restored using PixelShuffle are affected by heavy color cast. Using the UnPack operation in place of PixelShuffle improves the PSNR/SSIM from 22.72 dB/0.68 to 23.27 dB/0.69. Thus, the UnPack operation favors better color restoration.
 
 \textcolor{black}{In Section \ref{sec:unpack} we ascribed the better color restoration performance of Pack/UnPack over PixelShuffle to the fact that PixelShuffle breaks the RGB ordering in LR space, especially for large upsampling factors, whereas UnPack preserves the RGB ordering for any factor. To further test this hypothesis, we conducted an ablation study where we changed the image downsampling factor from 16$\times$ to 8$\times$, so that the final UnPack/PixelShuffle operation performs 4$\times$ upsampling instead of 8$\times$. This results in increased time and computational complexity for LLPackNet, but it reduces the separation between the Red and Blue channels in PixelShuffle. The performance of UnPack (23.29 dB) is almost the same as the 16$\times$ upsampling case, however the performance of PixelShuffle (23.28 dB) improves. In the case of 8$\times$ upsampling UnPack shows a gain of about 0.6 dB over PixelShuffle but for the case of 4$\times$ upsampling the difference reduces to 0.01dB. This confirms our hypothesis that Pack/UnPack has better performance because it preserves the RGB ordering for any upsampling factor.}
 
 \textbf{Estimating proper amplification: }Fig. \ref{fig:amplifier_visual_comparison} shows the restoration results using LLPackNet with and without the amplifier. Similar to the scotopic vision \cite{scotopic1,scotopic2}, without the amplifier, the restoration has faded colors. But the removal of the amplifier does not induce the annoying artifacts seen in restoration done using Chen \etal (see Fig. \ref{fig:main_visual_comparison} (B)). This can be attributed to the large receptive field due to the Pack operation. With the amplifier the performance of the network improves from 22.53dB/0.66 to 23.27dB/0.69.
 
Overall the combined effect of using the proposed Pack/UnPack operation over PixelShuffle, and estimating proper amplification, increases the average PSNR/SSIM from 21.35 dB / 0.60 to 23.27 dB / 0.69.

\begin{table}[t!]
\centering
\scriptsize
\bgroup
\def\arraystretch{1}
\begin{tabular}{c|c|c|c}
\hline \hline \textbf{Model} & \textbf{Processing Time} & \textbf{PSNR (dB)} & \textbf{SSIM} \\
\hline \hline
\textbf{Chen \etal \cite{chen2018learning}} & $0.21$ sec.  & $18.82$ & $0.73$ \\
\textbf{LIME \cite{2016lime}} & \textcolor{black}{0.19} sec.  & $16.94$ & $0.60$ \\
\textbf{Li \etal \cite{TIP2018structurePreserving}} &$17.89$ sec.  & $13.85$ & $0.65$ \\
\textbf{Gu \etal \cite{gu2019self}} &$0.41$ sec.  & $19.46$ & \textcolor{black}{\textbf{0.75}} \\\hline
\textbf{LLPackNet-8$\times$} \textbf{(Proposed)} & \textcolor{black}{\textbf{0.06}} sec. & \textcolor{black}{\textbf{19.61}} & $0.69$ \\
\textbf{LLPackNet-4$\times$} \textbf{(Proposed)} & $0.24$ sec. & \textcolor{black}{19.60} & \textcolor{black}{0.74} \\ \hline
\end{tabular}
\egroup
 \caption{Results on the LOL dataset \cite{LOL} for weakly illuminated compressed images having $400 \times 600$ VGA resolution. LLPackNet with 8$\times$ downsampling (LLPackNet-8$\times$) is very fast. But, since the image resolution is quite low, LLPackNet-4$\times$ opts for smaller downsampling to achieve better reconstruction as reflected in the SSIM value. 
}
 \label{tab:LOL}
 \vspace{-0.5cm}
\end{table}
\subsection{LLPackNet for low-resolution images}
The SID dataset contains high definition images, thereby, allowing us to chose a large downsampling factor of 16. This leads us to the question: Can LLPackNet also work for LR images? When LR images are downsampled using a large factor, the intra-channel correlation in the downsampled image is reduced, which negatively impacts the restoration. 
To investigate this, we conducted experiments on the LOL dataset \cite{LOL} containing weakly illuminated images at VGA resolution of 400$\times$600. As the images in the LOL dataset are already in the compressed PNG format, the 2$\times$ downsampling at the beginning of LLPackNet to separate out the Bayer pattern is not required. Thus, the effective downsampling is only 8$\times$ and we denote this network by LLPackNet-8$\times$. The results are shown in Table \ref{tab:LOL}. Once again, LLPackNet has the lowest processing time. We further observe that the large receptive field of LLPackNet enhances the denoising and color restoration capabilities. But, a slight blur is also introduced. To verify that the blur is because of large downsampling, we retrain LLPackNet on the LOL dataset with 4$\times$ downsampling, which we denote as LLPackNet-4$\times$. With LLPackNet-4$\times$, we obtain sharper results having higher SSIM values. 
\vspace{-0.4cm}

\section{Conclusion}
\vspace{-0.1cm}
In this paper, a fast and light-weight extreme low-light image enhancement network (LLPackNet) has been presented. LLPackNet performs aggressive down/up-sampling using the proposed Pack/UnPack operations to obtain a large receptive field and better color restoration. The network also uses a novel amplifier module that amplifies the input image without relying on ground-truth information. Overall, LLPackNet is 5--20$\times$ faster and  2--3$\times$ lighter, and yet maintains a competitive restoration quality compared to state-of-the-art algorithms.

 \clearpage

\section{Supplementary Material}
\begin{figure*}[h!]
	
	\centering
	\scriptsize	
    \begin{tabular}{ccccc}
    
    {\footnotesize \textbf{Maharjan \etal}} & {\footnotesize \textbf{Gu \etal}} & {\footnotesize \textbf{Chen \etal}} & {\footnotesize \textbf{Ours}} & {\footnotesize \textbf{GT}}\\
    \includegraphics[width=0.18 \linewidth]{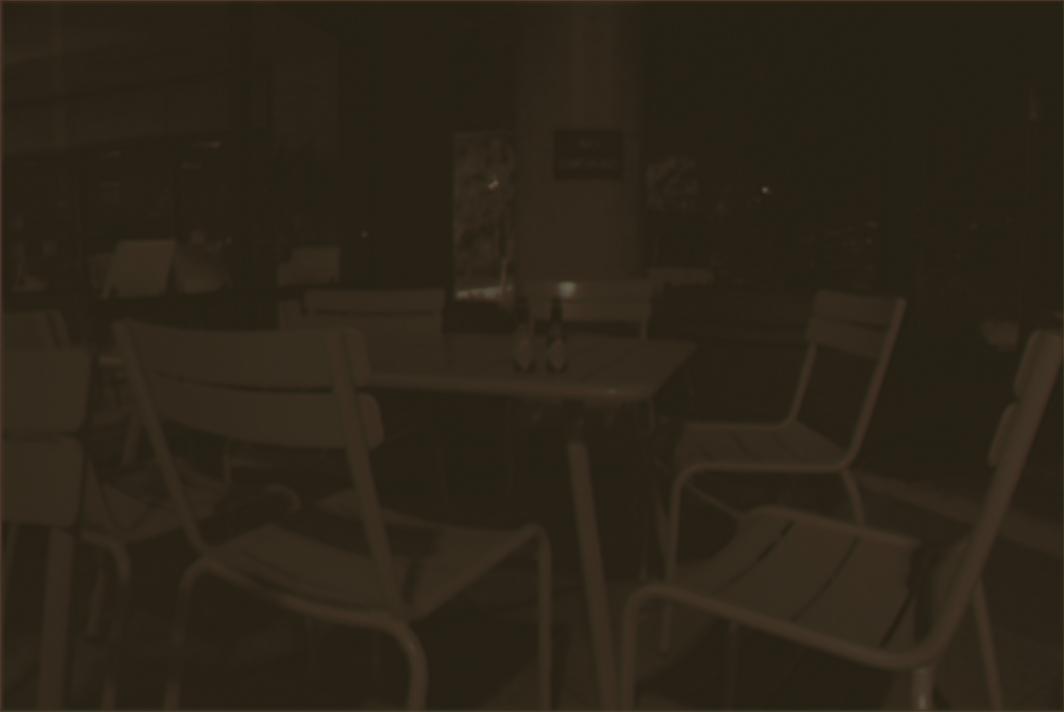} \hspace{-0.4cm}   & \includegraphics[width=0.18 \linewidth]{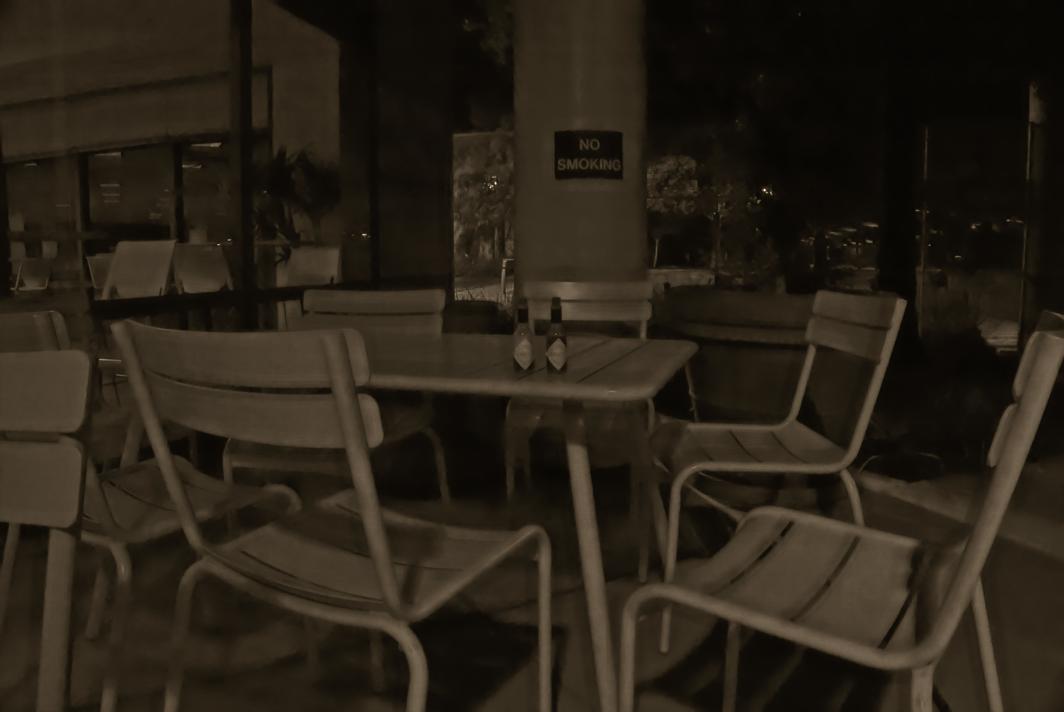} \hspace{-0.4cm} &  
     \includegraphics[width=0.18 \linewidth]{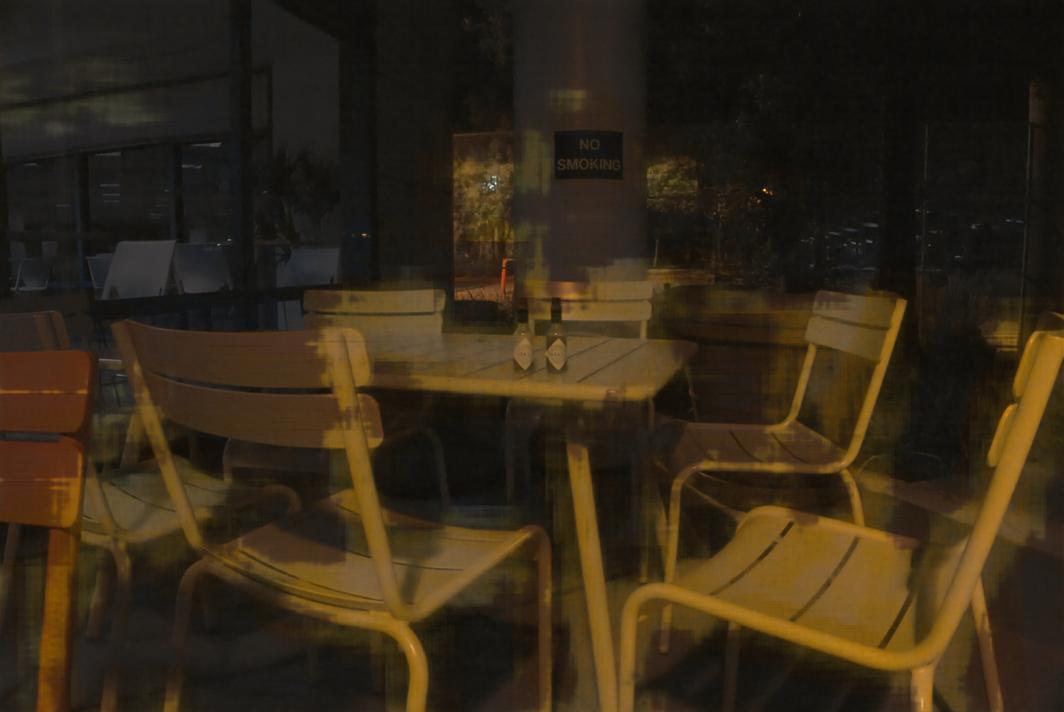} \hspace{-0.4cm} &  
     \includegraphics[width=0.18 \linewidth]{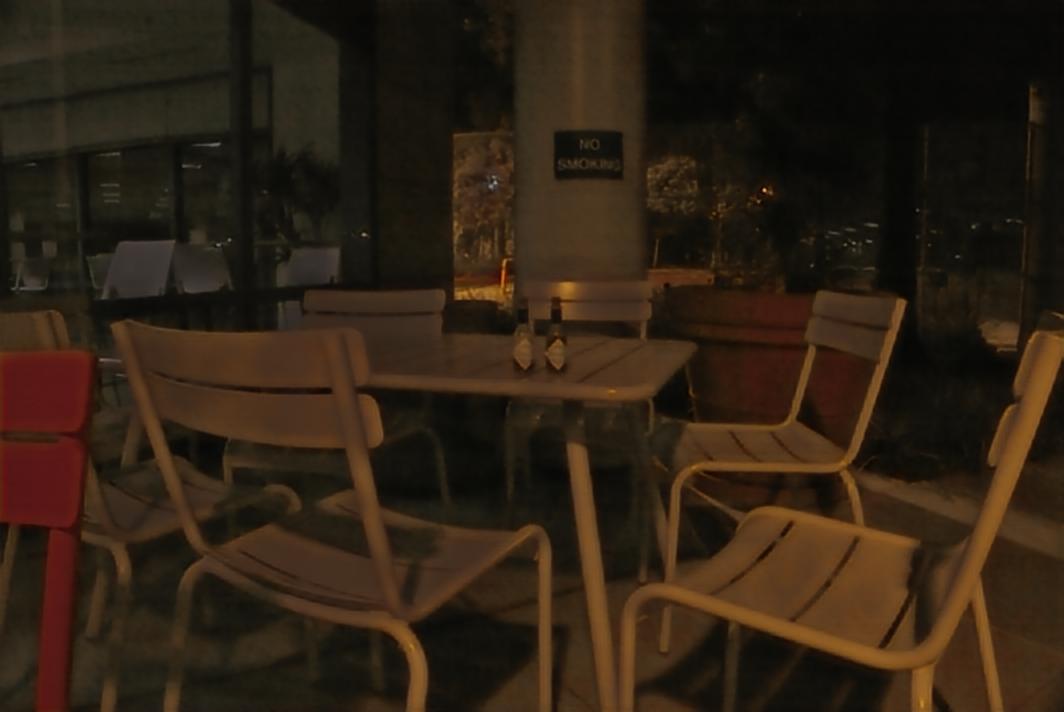} \hspace{-0.4cm} &  
     \includegraphics[width=0.18 \linewidth]{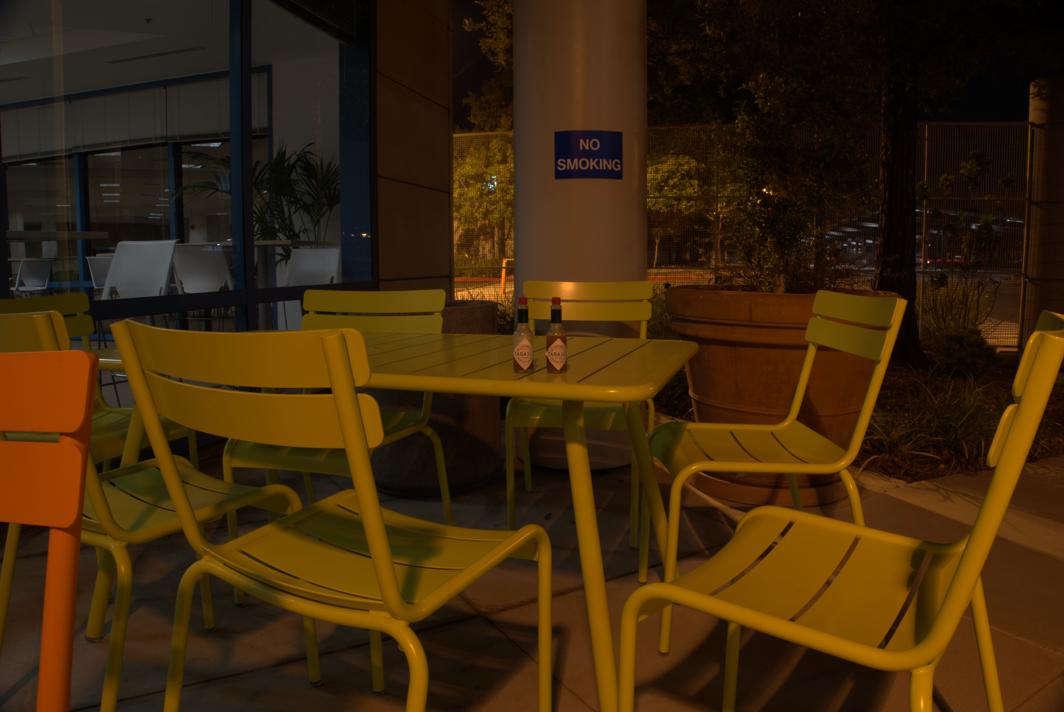}\\
     18.70/0.53 & 20.68/0.62 & \textcolor{black}{22.96}/\textcolor{black}{0.70} & \textbf{23.08}/\textbf{0.74} \\
     
     \includegraphics[width=0.18 \linewidth]{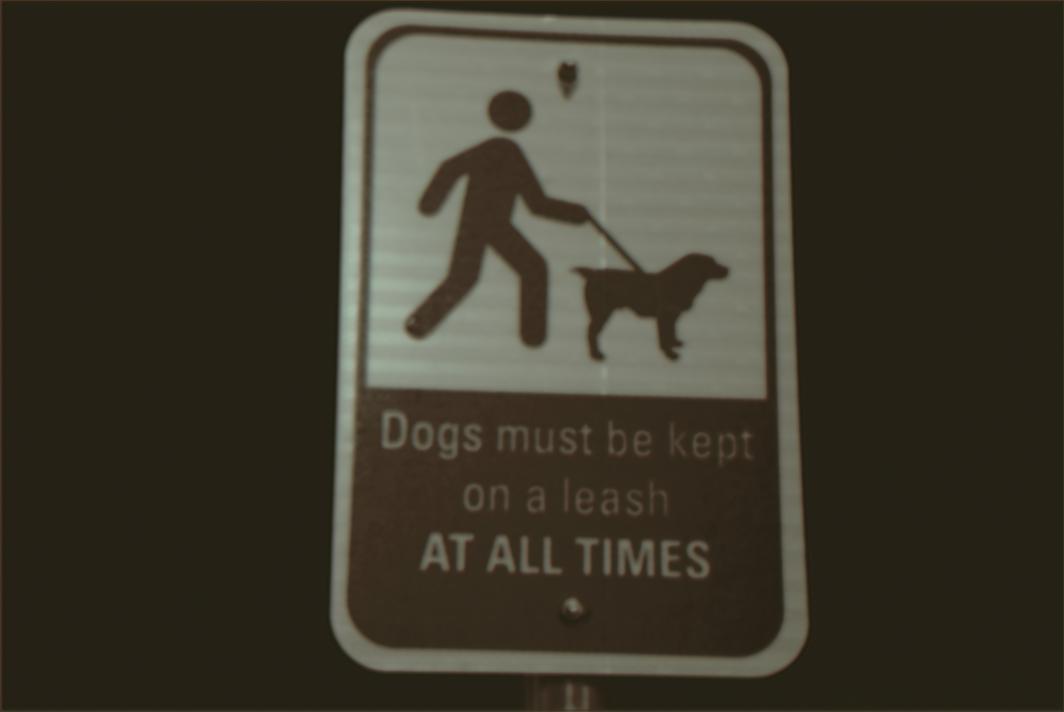} \hspace{-0.4cm}   & \includegraphics[width=0.18 \linewidth]{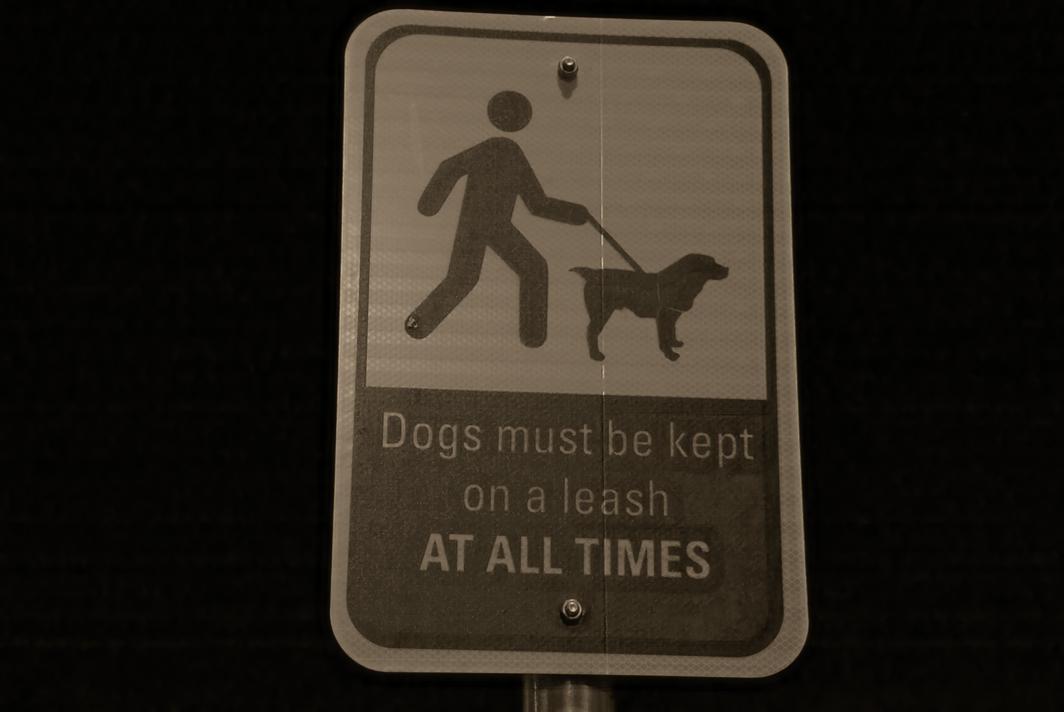} \hspace{-0.4cm} &  
     \includegraphics[width=0.18 \linewidth]{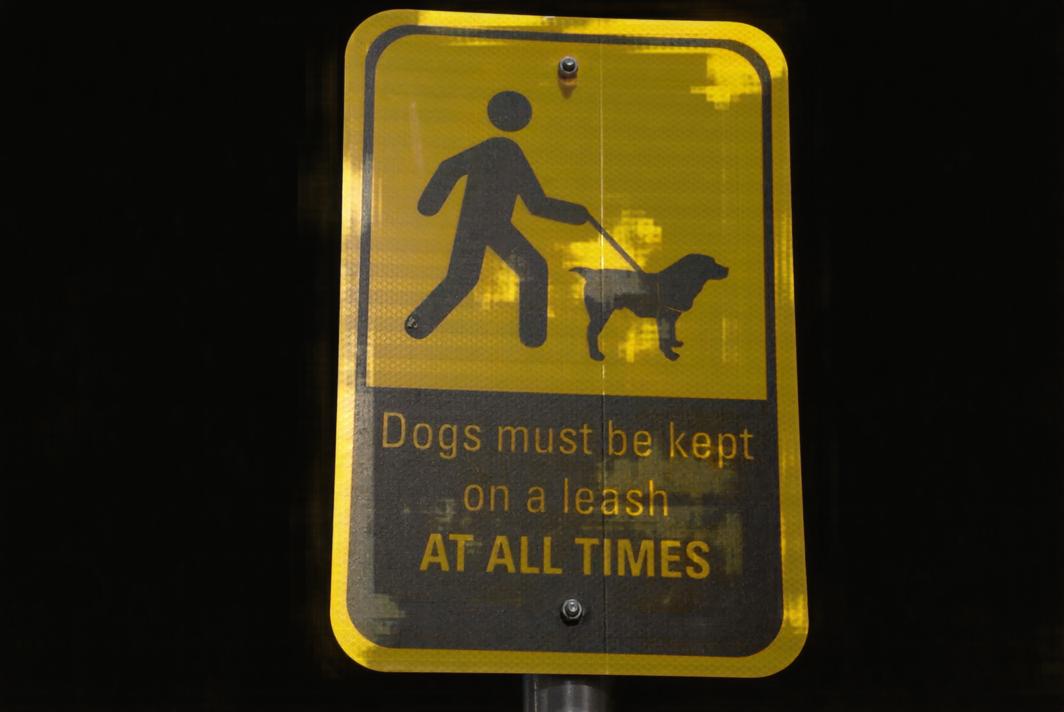} \hspace{-0.4cm} &  
     \includegraphics[width=0.18 \linewidth]{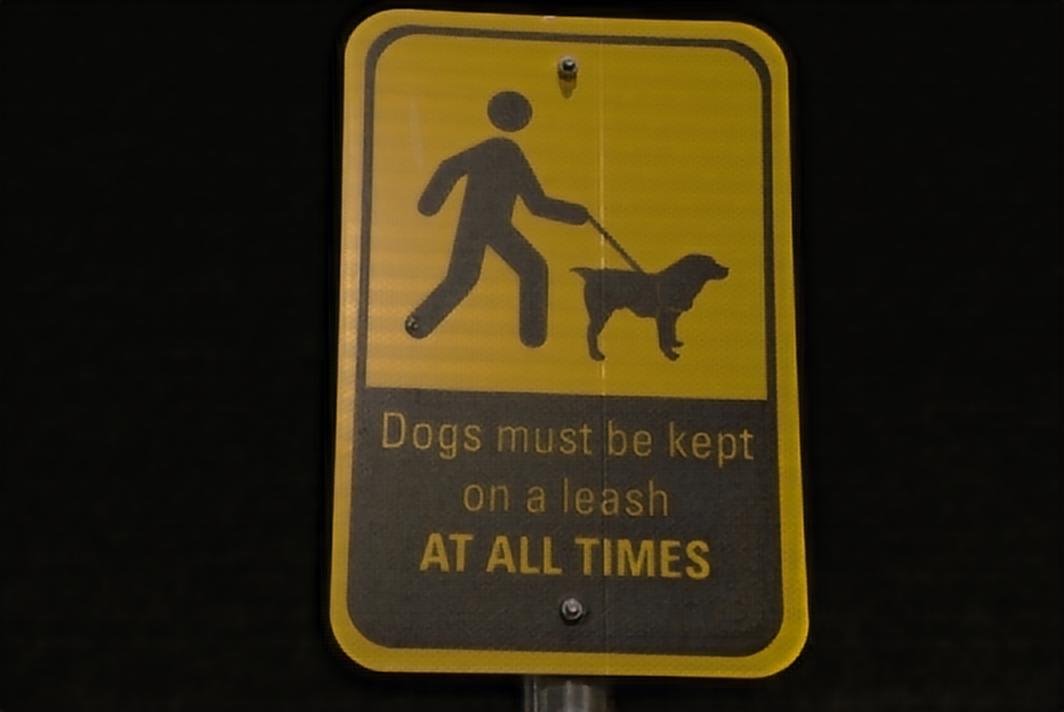} \hspace{-0.4cm} &  
     \includegraphics[width=0.18 \linewidth]{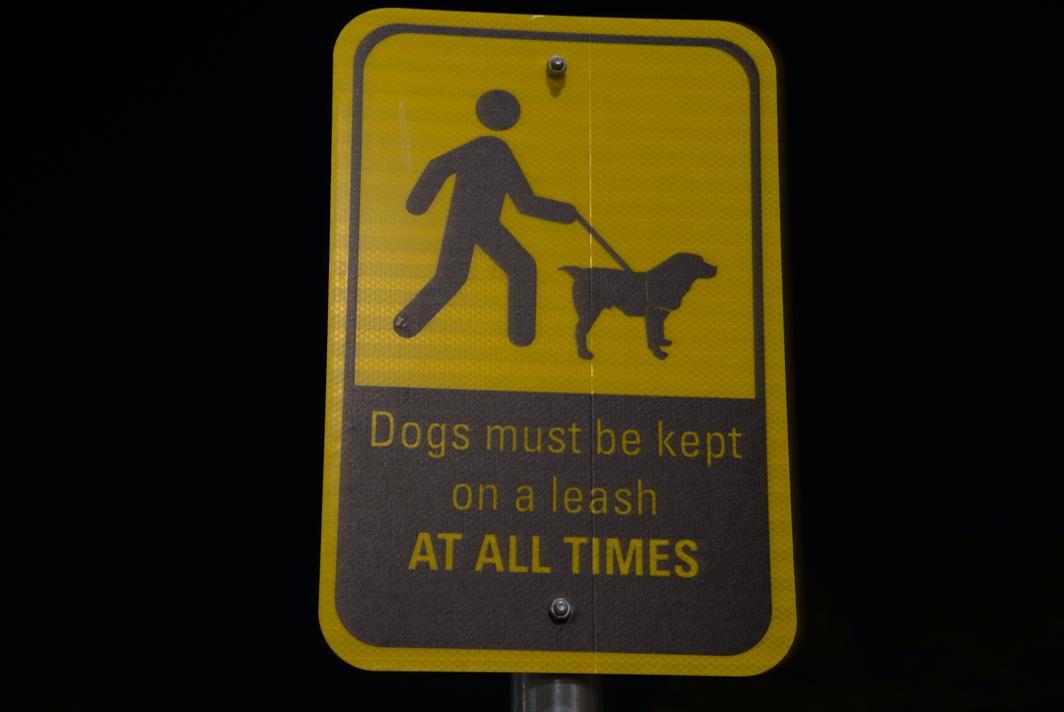}\\
     16.02/0.19 & 18.64/\textbf{0.65} & \textcolor{black}{19.46}/\textcolor{black}{0.62} & \textbf{20.85}/0.59 \\
     
     \includegraphics[width=0.18 \linewidth]{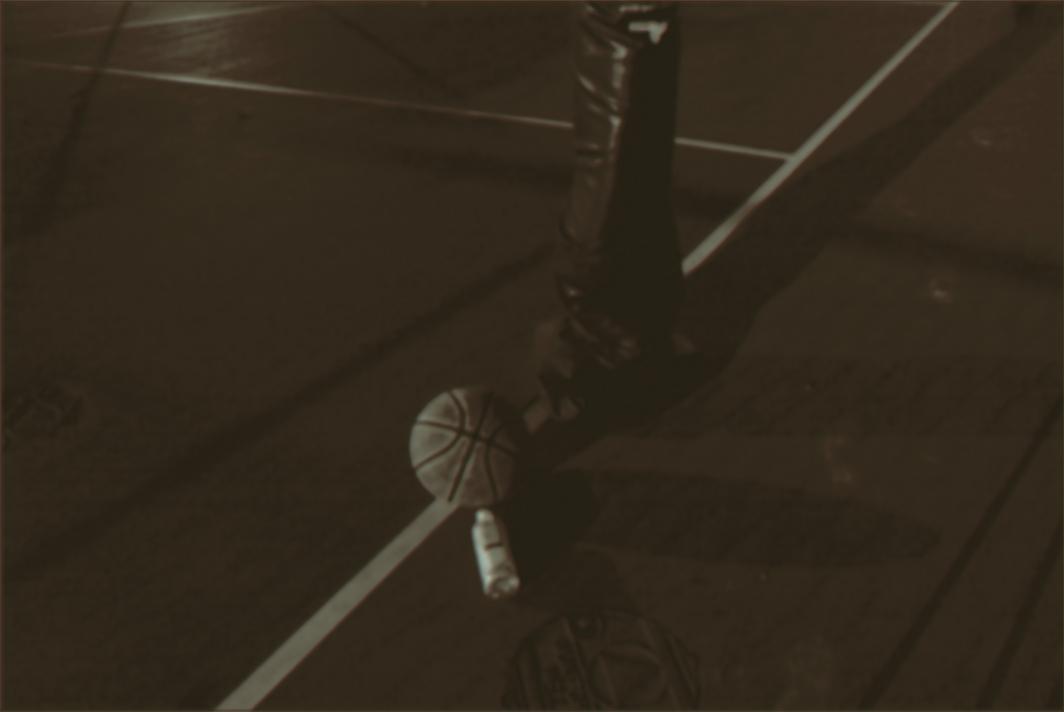} \hspace{-0.4cm}   & \includegraphics[width=0.18 \linewidth]{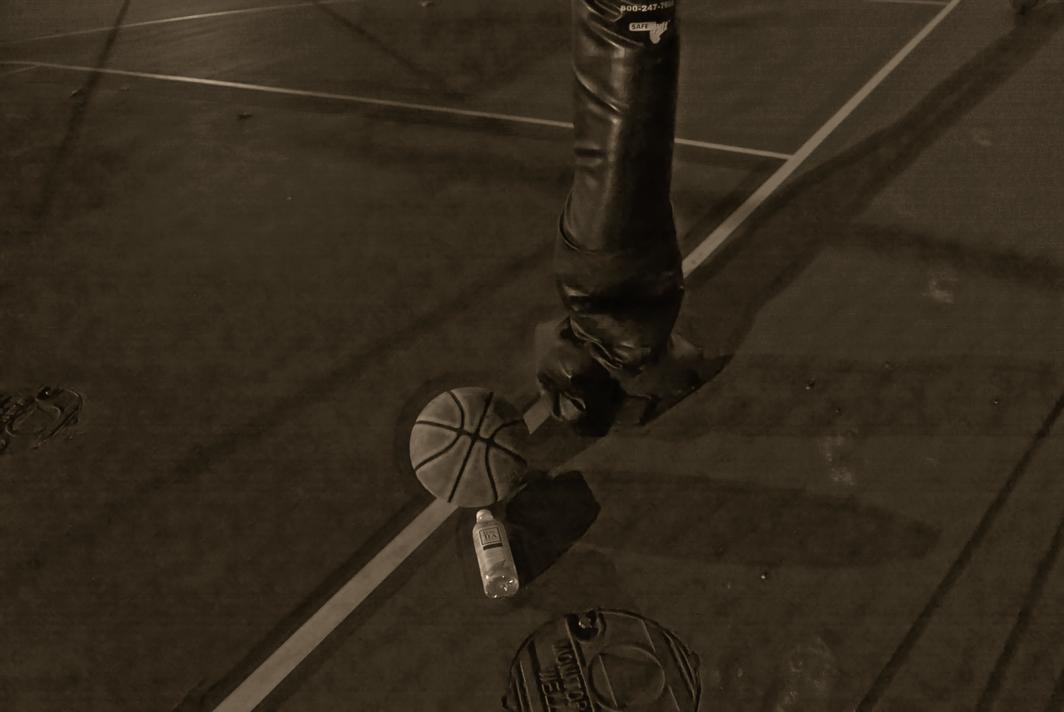} \hspace{-0.4cm} &  
     \includegraphics[width=0.18 \linewidth]{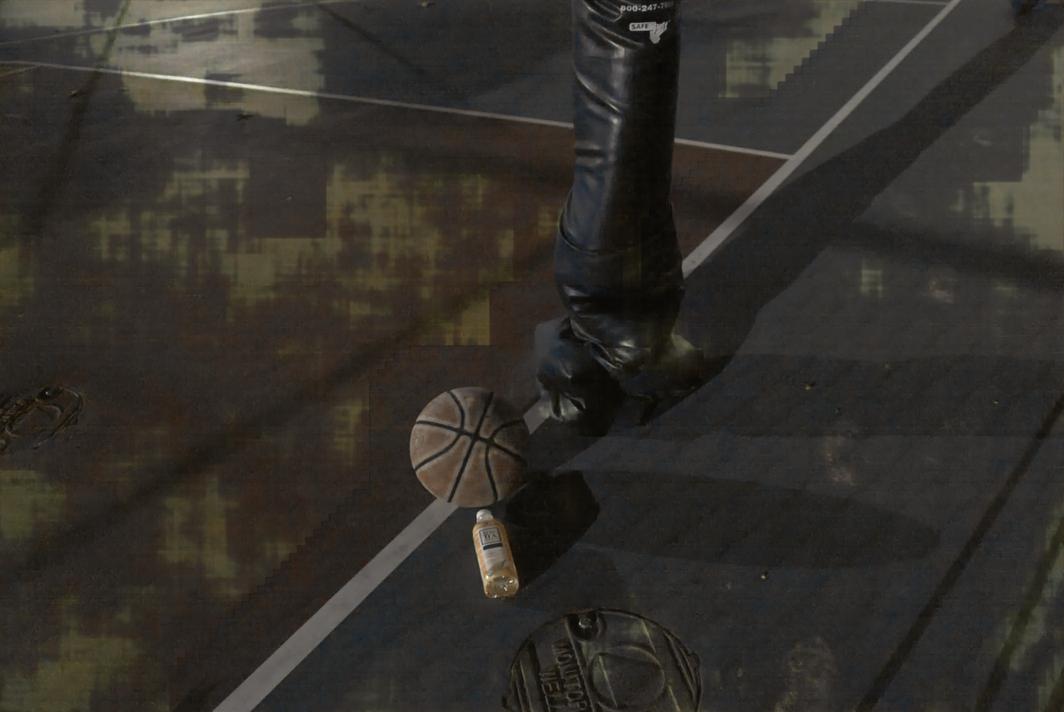} \hspace{-0.4cm} &  
     \includegraphics[width=0.18 \linewidth]{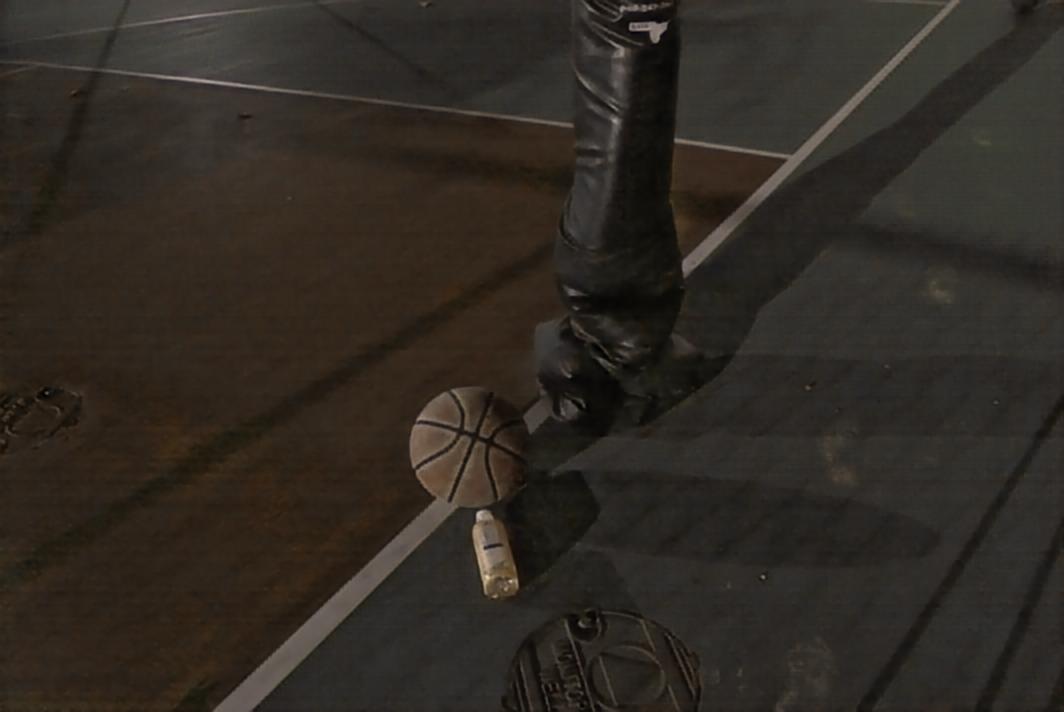} \hspace{-0.4cm} &  
     \includegraphics[width=0.18 \linewidth]{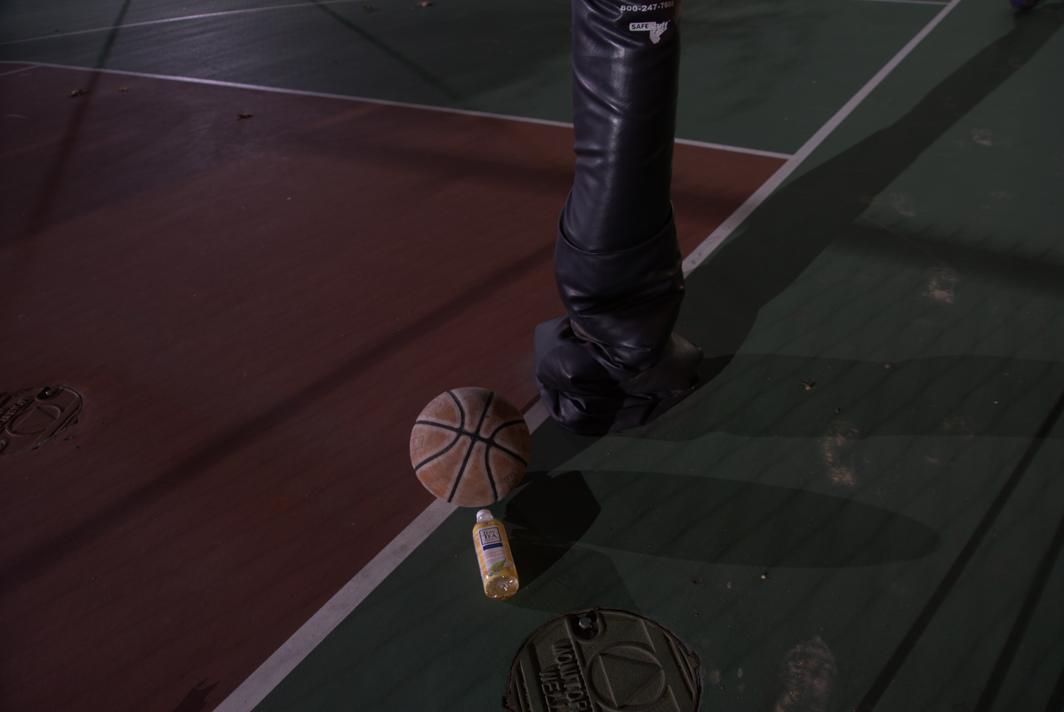}\\
     26.53/0.52 & \textcolor{black}{27.38}/0.55 & 24.57/\textcolor{black}{0.77} & \textbf{31.96}/\textbf{0.78} \\
     
     \includegraphics[width=0.18 \linewidth]{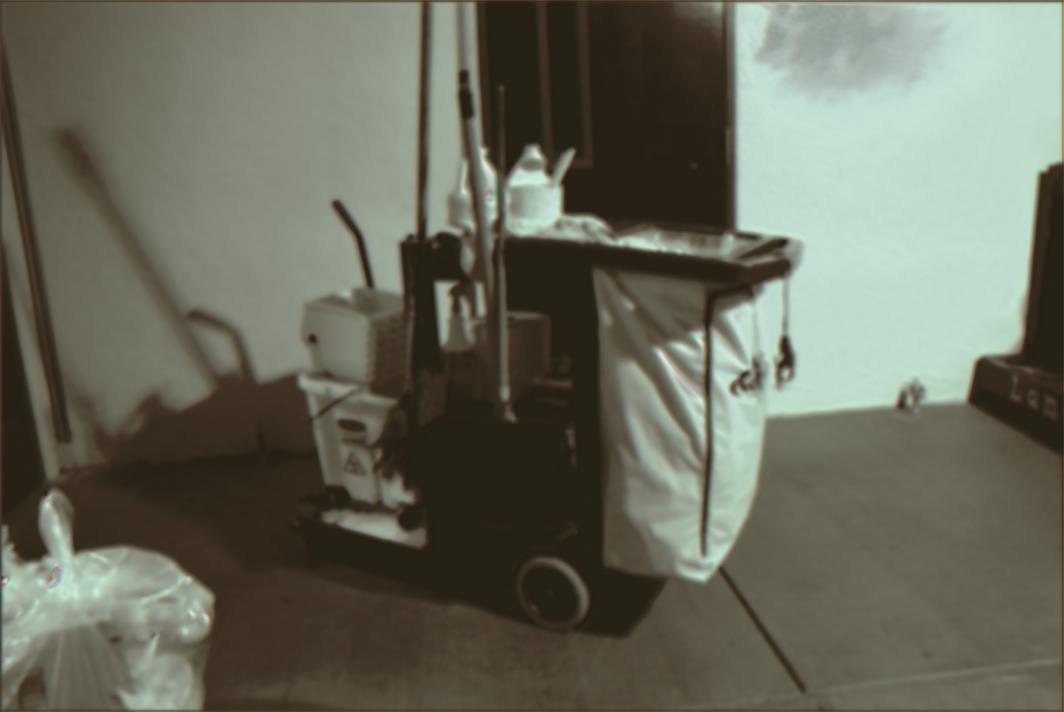} \hspace{-0.4cm}   & \includegraphics[width=0.18 \linewidth]{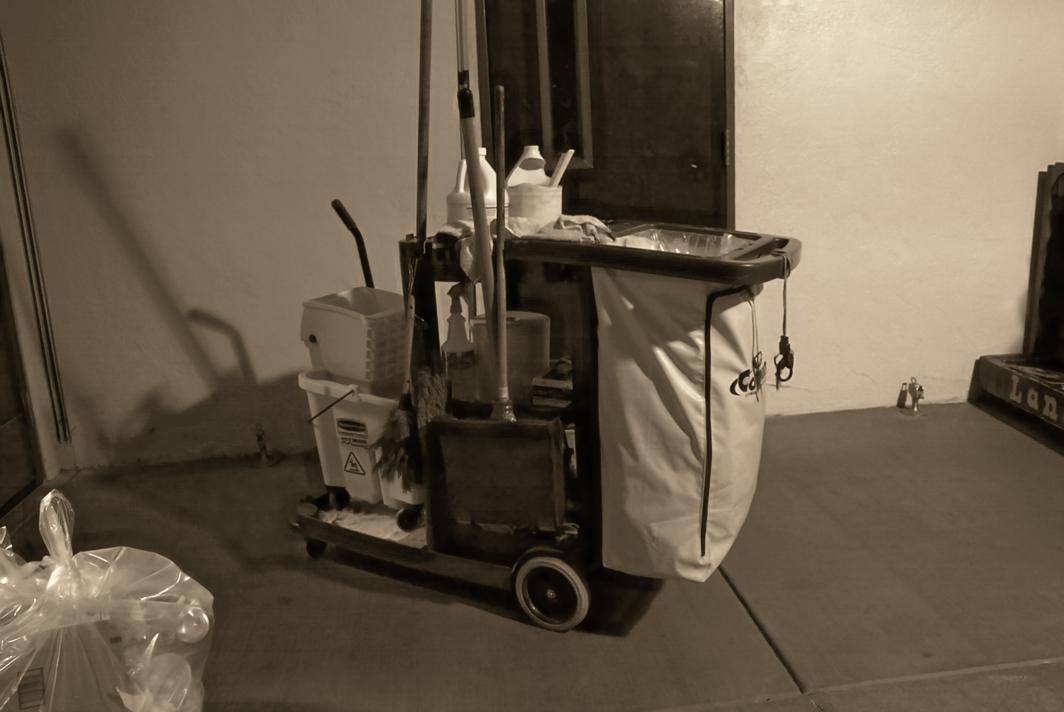} \hspace{-0.4cm} &  
     \includegraphics[width=0.18 \linewidth]{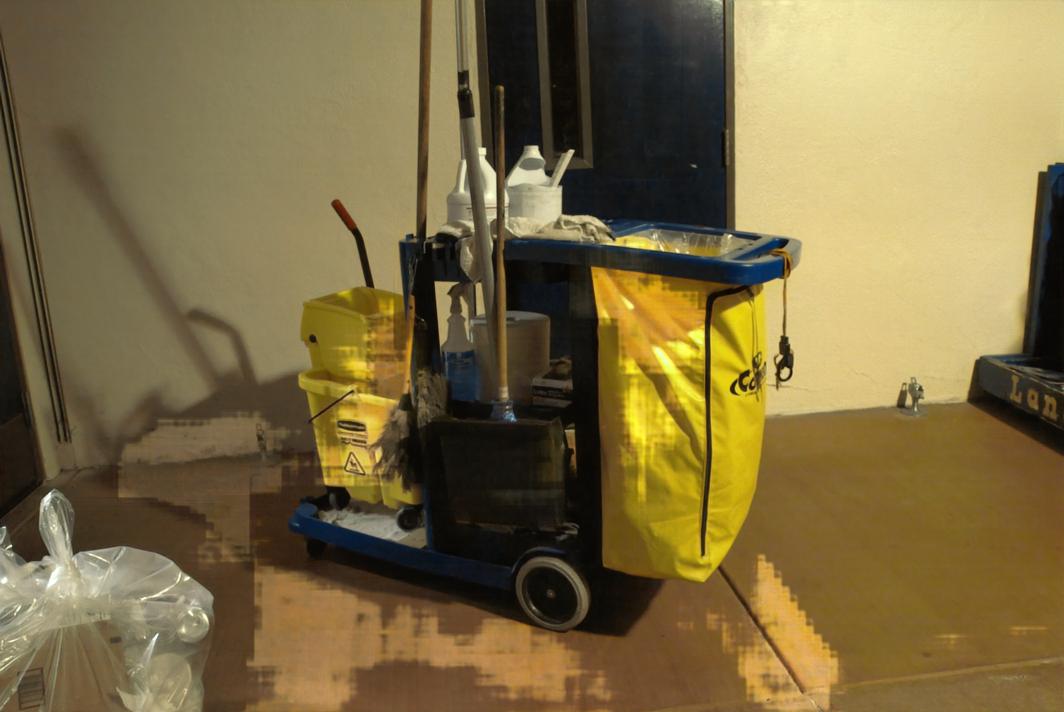} \hspace{-0.4cm} &  
     \includegraphics[width=0.18 \linewidth]{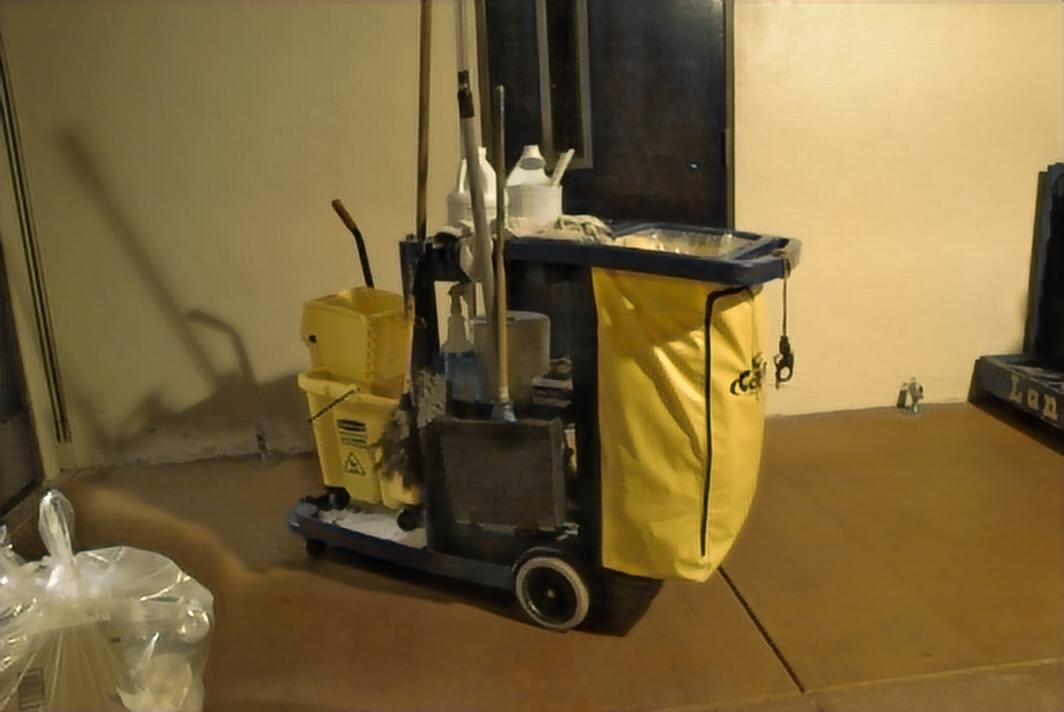} \hspace{-0.4cm} &  
     \includegraphics[width=0.18 \linewidth]{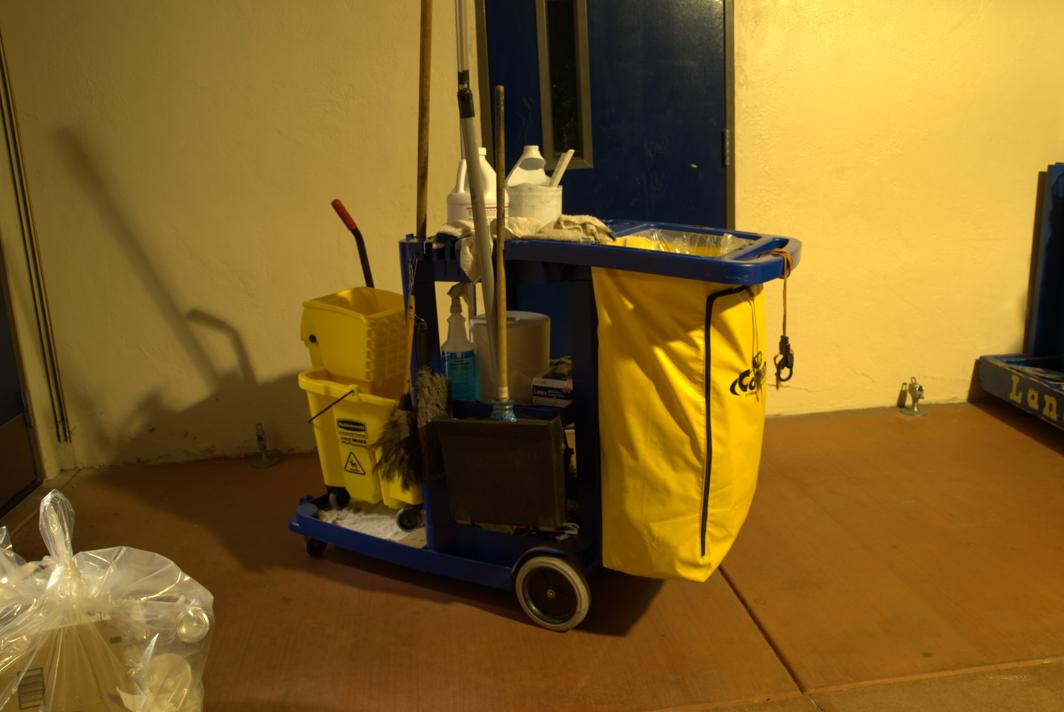}\\
     14.61/0.29 & 17.39/0.55 & \textcolor{black}{19.53}/\textcolor{black}{0.81} & \textbf{21.82}/\textbf{0.83} \\
     
     \includegraphics[width=0.18 \linewidth]{main/icme/31_IMG_PRED.jpg} \hspace{-0.4cm}   & \includegraphics[width=0.18 \linewidth]{main/iccv/31_IMG_PRED.jpg} \hspace{-0.4cm} &  
     \includegraphics[width=0.18 \linewidth]{main/cvpr/31_IMG_PRED.jpg} \hspace{-0.4cm} &  
     \includegraphics[width=0.18 \linewidth]{main/ours/31_IMG_PRED.jpg} \hspace{-0.4cm} &  
     \includegraphics[width=0.18 \linewidth]{main/ours/31_IMG_GT.jpg}\\
     \textcolor{black}{25.39}/\textcolor{black}{0.45} & 22.77/0.26 & 13.96/0.34 & \textbf{27.87}/\textbf{0.58} \\
     
     \includegraphics[width=0.18 \linewidth]{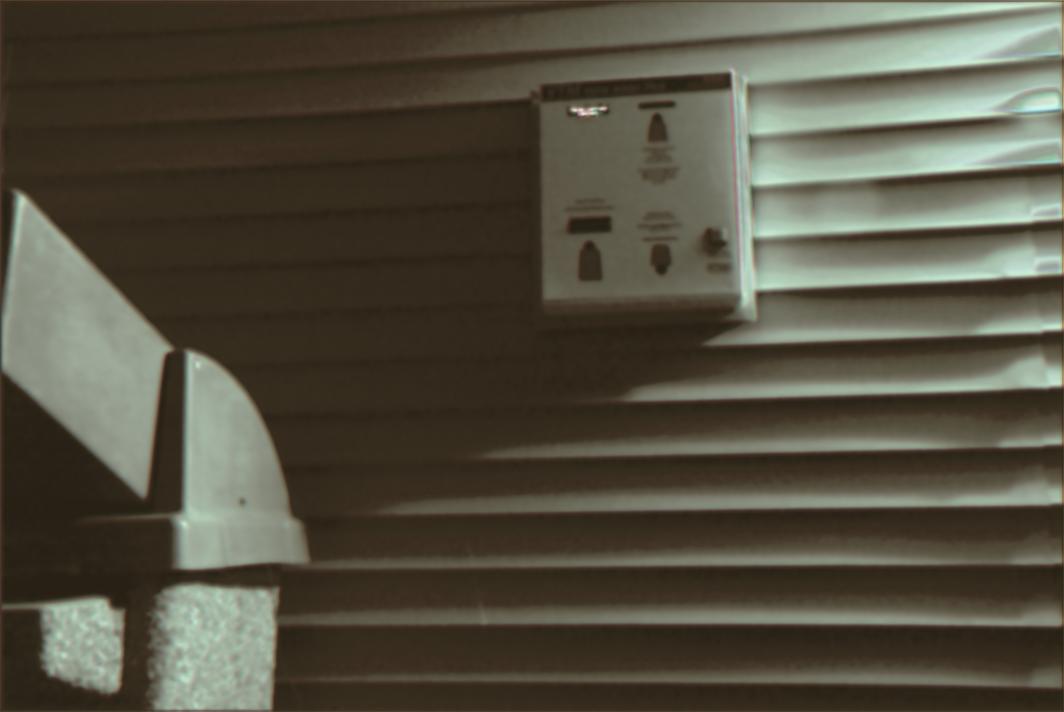} \hspace{-0.4cm}   & \includegraphics[width=0.18 \linewidth]{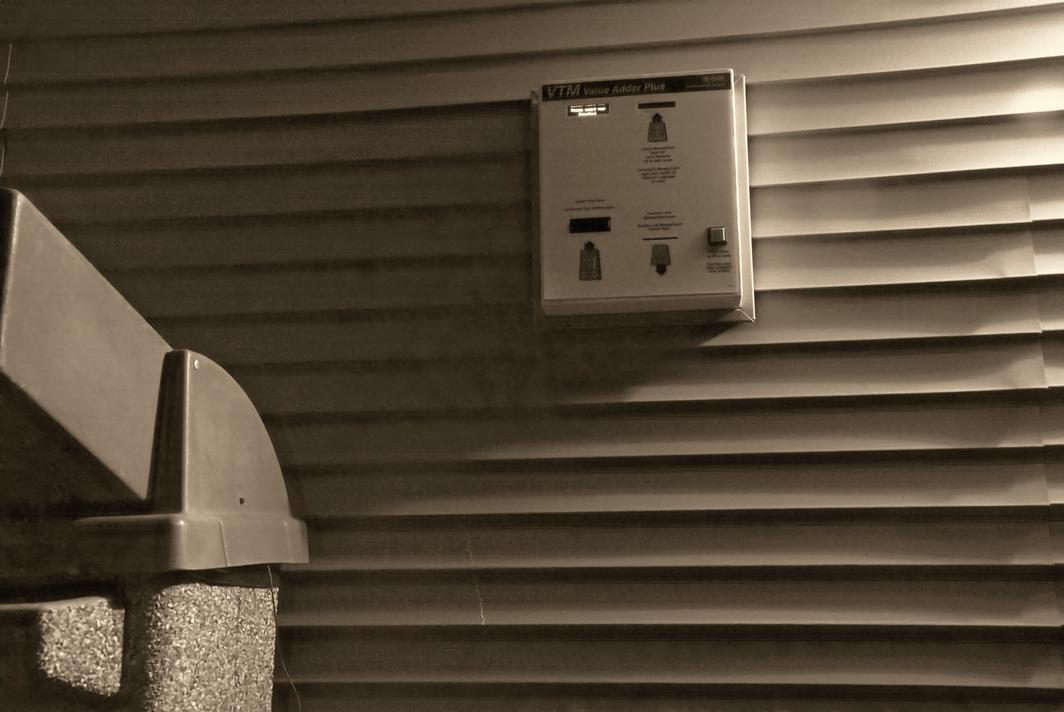} \hspace{-0.4cm} &  
     \includegraphics[width=0.18 \linewidth]{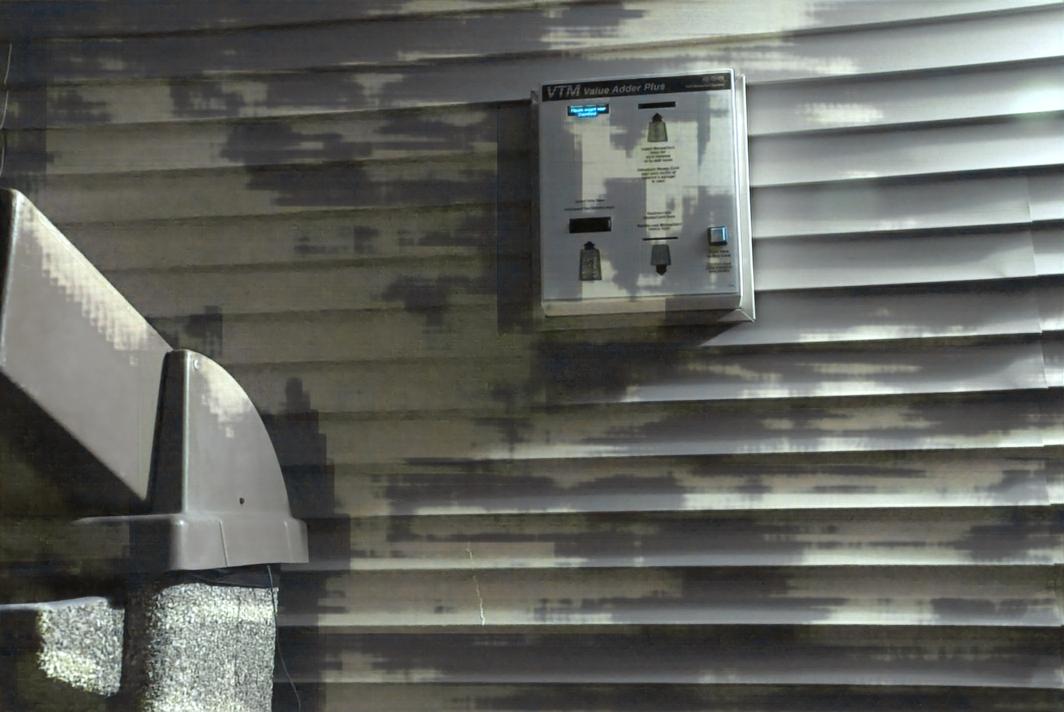} \hspace{-0.4cm} &  
     \includegraphics[width=0.18 \linewidth]{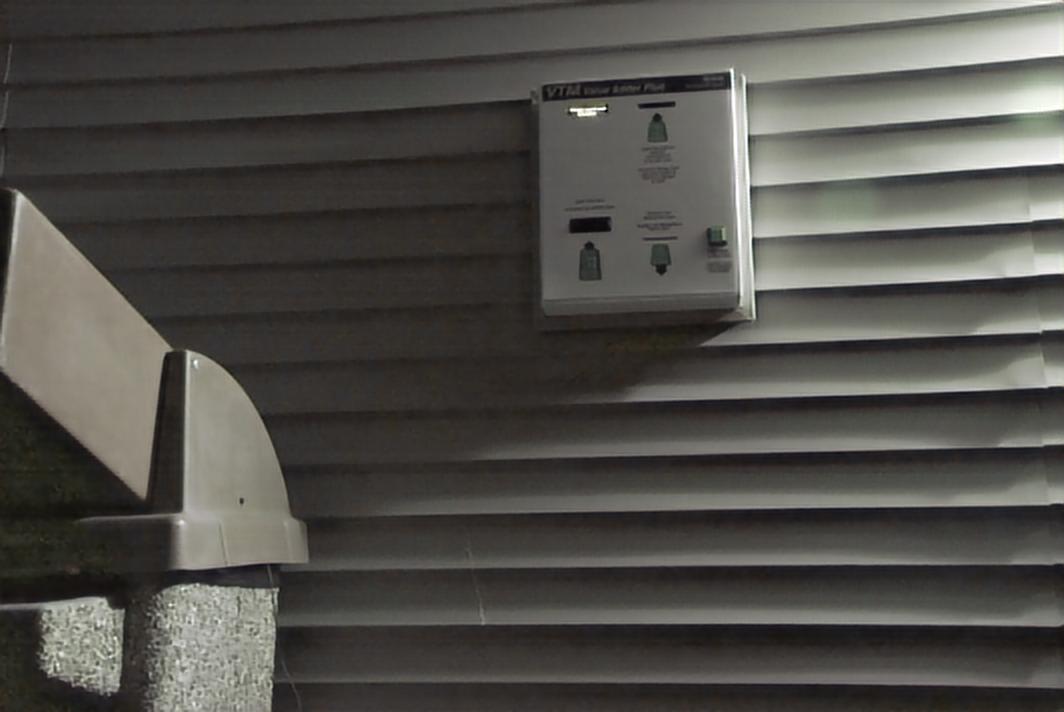} \hspace{-0.4cm} &  
     \includegraphics[width=0.18 \linewidth]{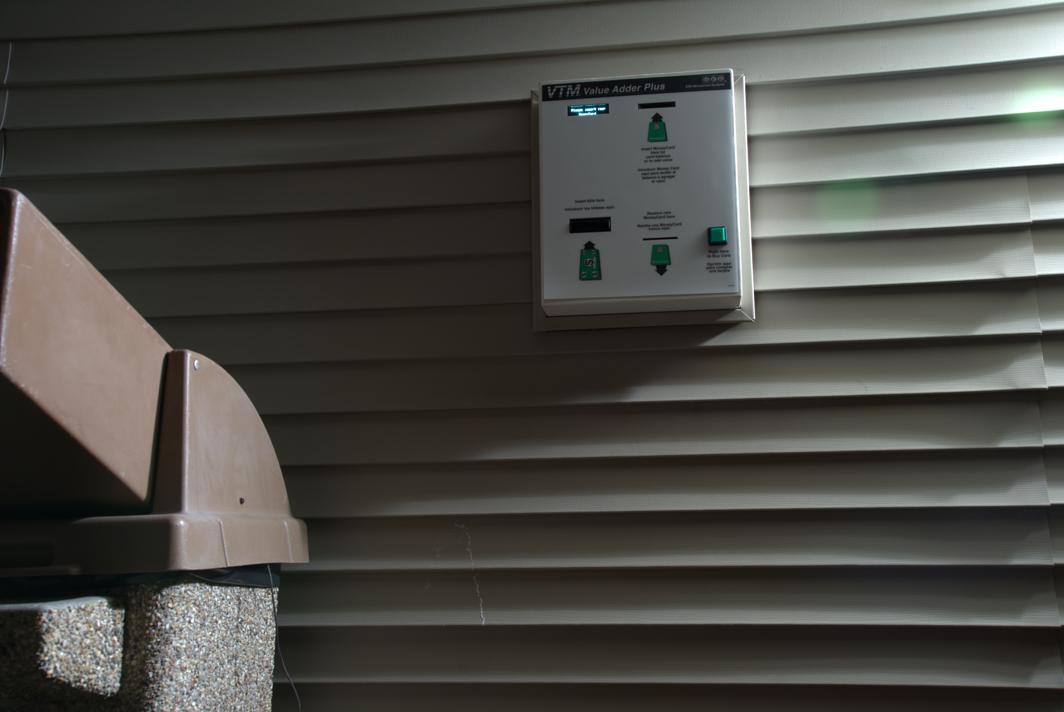}\\
     24.55/0.59 & \textcolor{black}{25.94}/0.57 & 17.84/\textcolor{black}{0.71} & \textbf{28.21}/\textbf{0.81} \\
     
     \includegraphics[width=0.18 \linewidth]{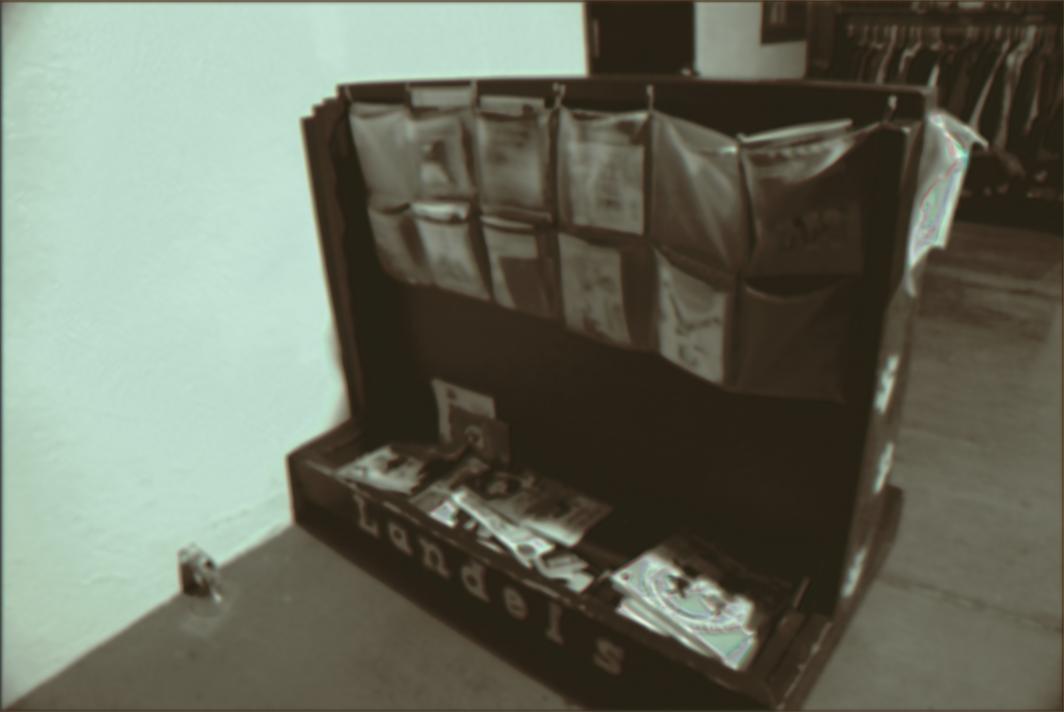} \hspace{-0.4cm}   & \includegraphics[width=0.18 \linewidth]{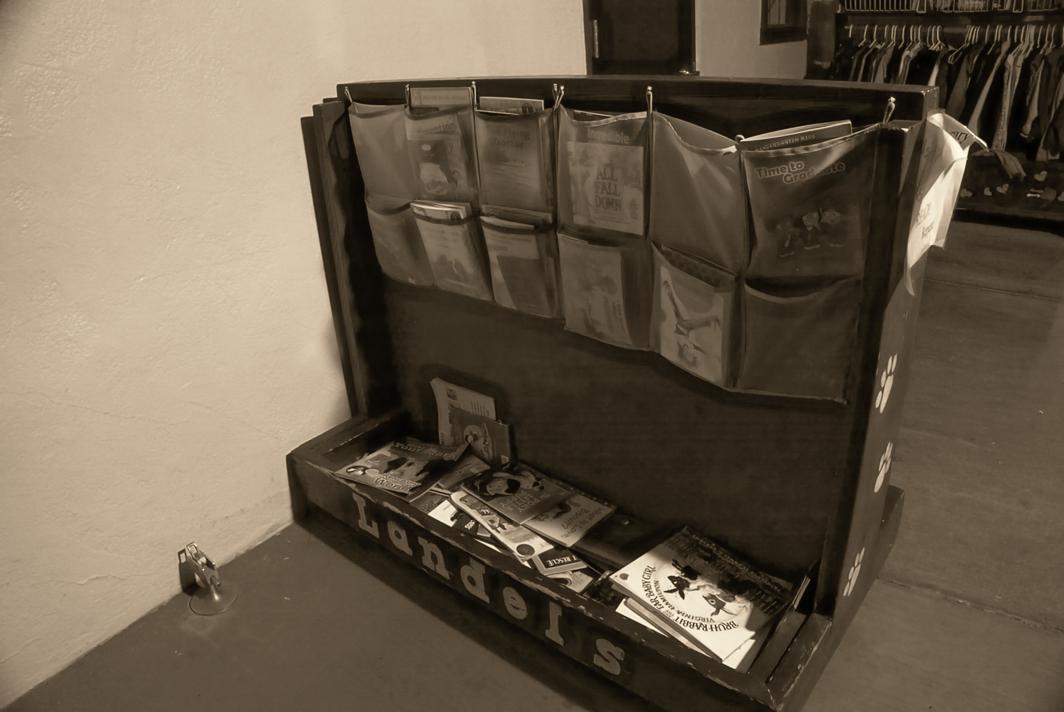} \hspace{-0.4cm} &  
     \includegraphics[width=0.18 \linewidth]{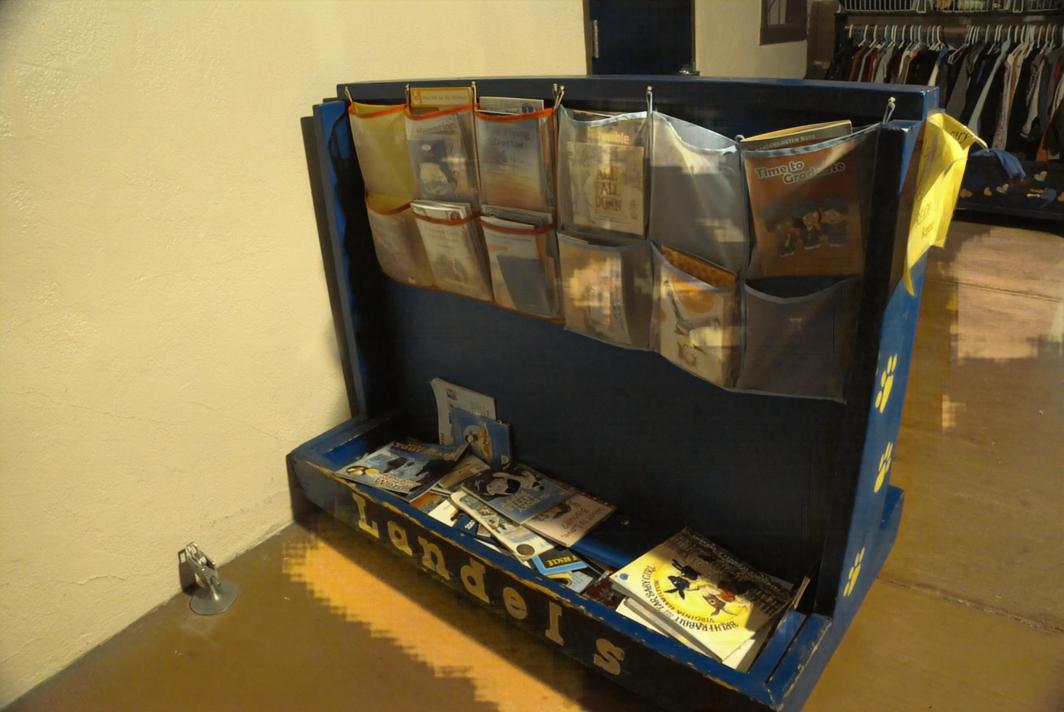} \hspace{-0.4cm} &  
     \includegraphics[width=0.18 \linewidth]{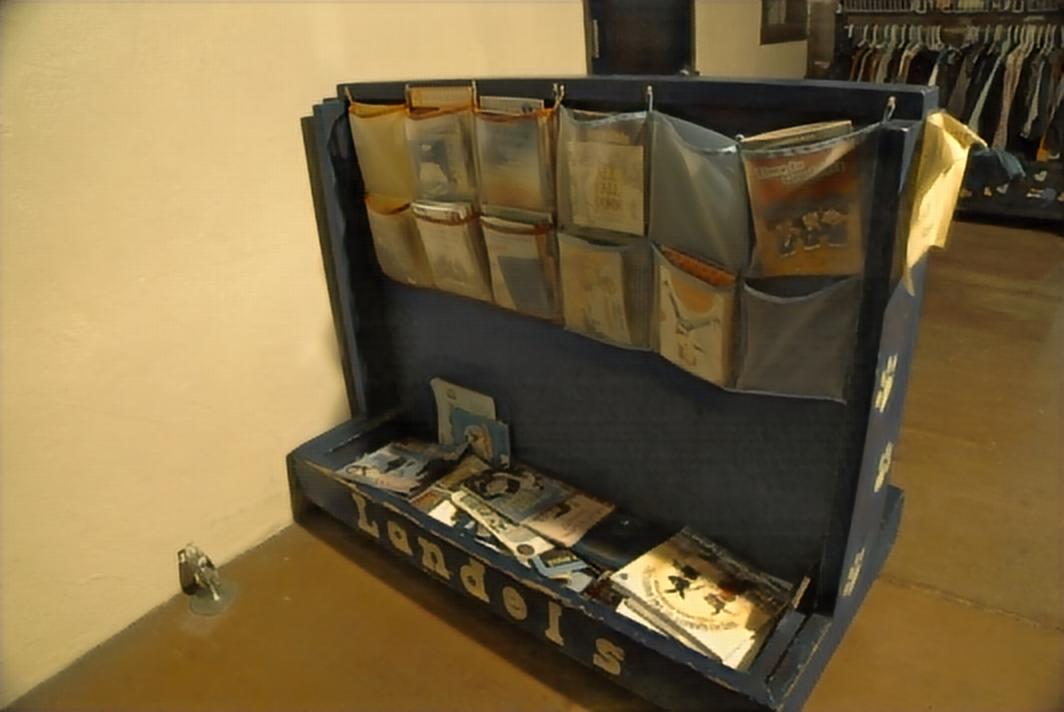} \hspace{-0.4cm} &  
     \includegraphics[width=0.18 \linewidth]{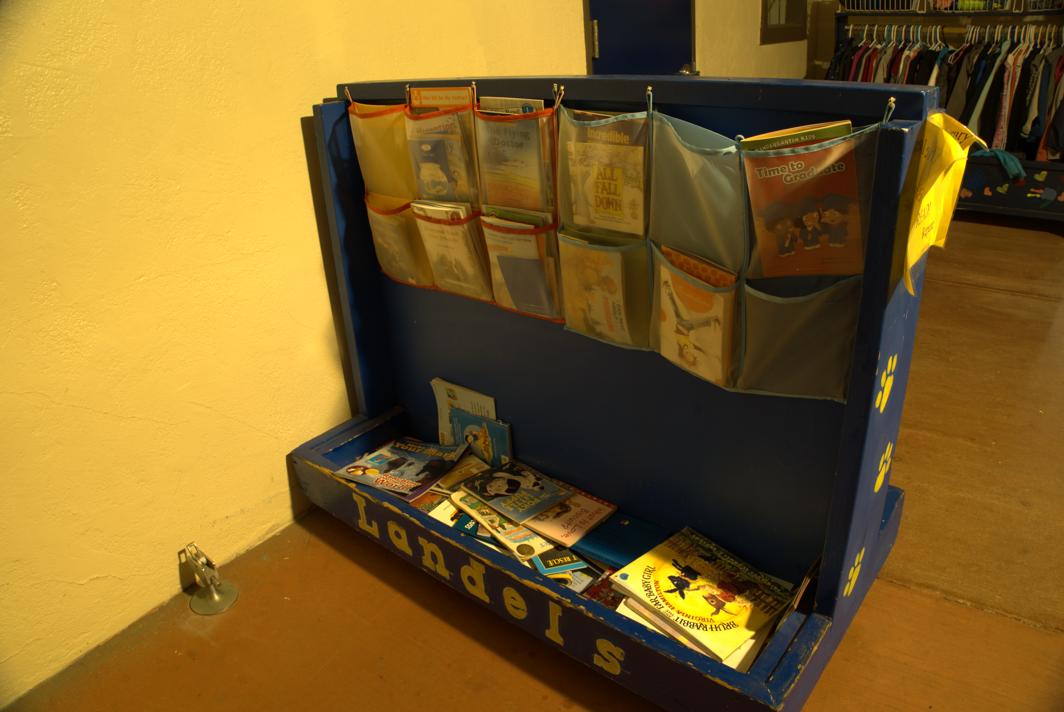}\\
     13.94/0.28 & 16.92/0.52 & \textcolor{black}{19.02}/\textbf{0.80} & \textbf{20.55}/\textbf{0.80} 
      
    \end{tabular}
    \caption{[Best viewed with maximum screen brightness] More visual comparisons corresponding to Table 1 of main paper for the practical case when GT exposure is not available.}
\label{fig:main_visual_comparison_supplementary}

\end{figure*}

\subsection{More qualitative results}

We show more qualitative results comparing the performance of the proposed method with existing methods. Fig. \ref{fig:main_visual_comparison_supplementary} shows results for enhancing extremely dark images for the practical scenario when the ratio of GT to input image exposure is not available. Refer to Fig. 4 (B) and Table 1 in main paper for more information about this setting.

\begin{figure*}[t!]
	
	\centering
	\scriptsize
	\setlength\tabcolsep{1pt}
    \begin{tabular}{cccccc}
    
    {\scriptsize \textbf{LIME}} & {\scriptsize \textbf{Chen \etal}} & {\scriptsize \textbf{Gu \etal}} & {\scriptsize \textbf{LLPackNet 8$\times$}} & {\scriptsize \textbf{LLPackNet 4$\times$}} & {\scriptsize \textbf{GT}}\\
    \includegraphics[width=0.16 \linewidth]{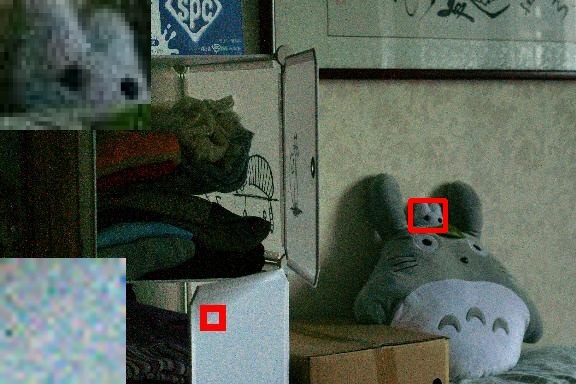}    & \includegraphics[width=0.16 \linewidth]{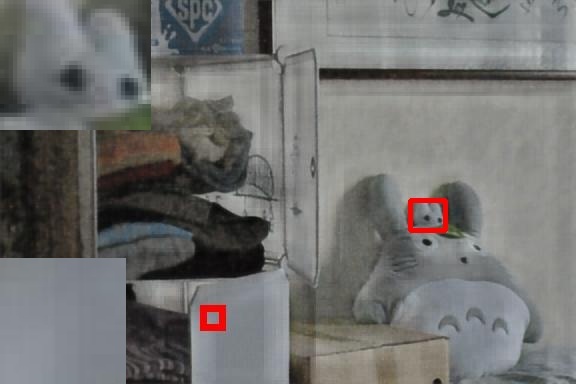} &  
     \includegraphics[width=0.16 \linewidth]{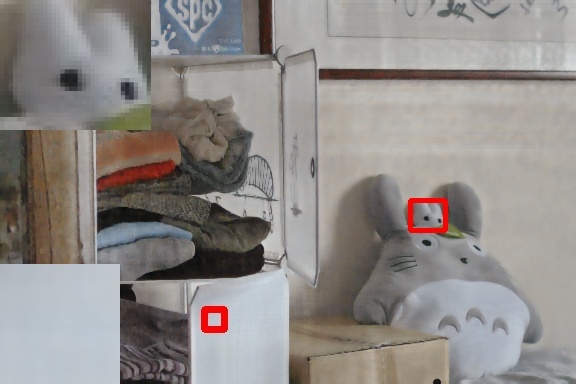} & 
     \includegraphics[width=0.16 \linewidth]{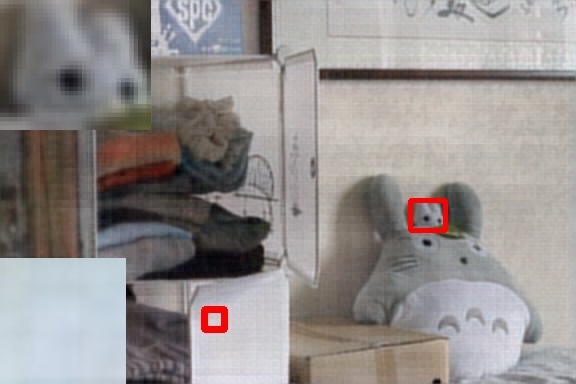}   & \includegraphics[width=0.16 \linewidth]{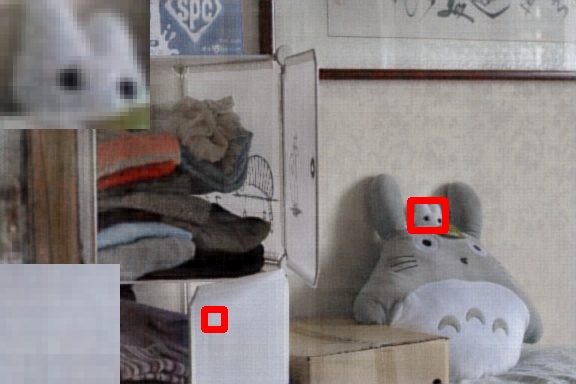} &  
     \includegraphics[width=0.16 \linewidth]{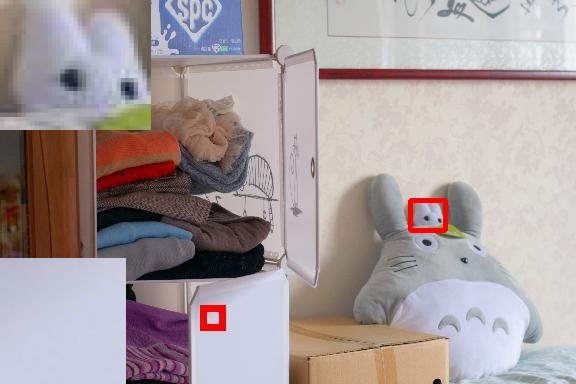}\\
     10.40/0.30 & 14.42/0.61 & 17.73/\textbf{0.69} & \textcolor{black}{17.77}/0.59 & \textbf{17.79}/\textcolor{black}{0.67}\\
     
     \includegraphics[width=0.16 \linewidth]{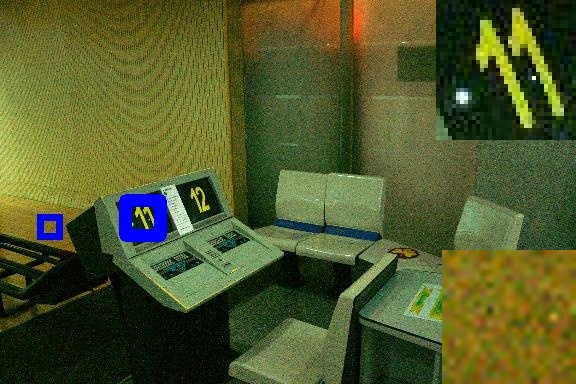}    & \includegraphics[width=0.16 \linewidth]{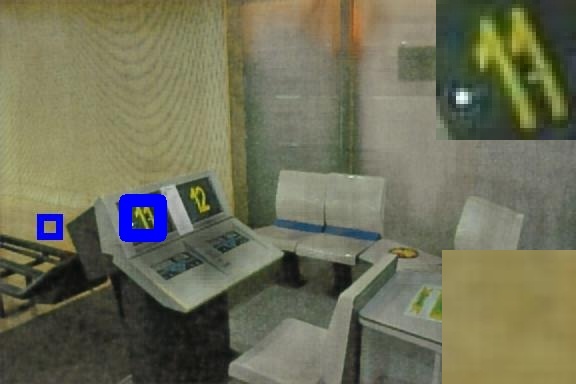} &  
     \includegraphics[width=0.16 \linewidth]{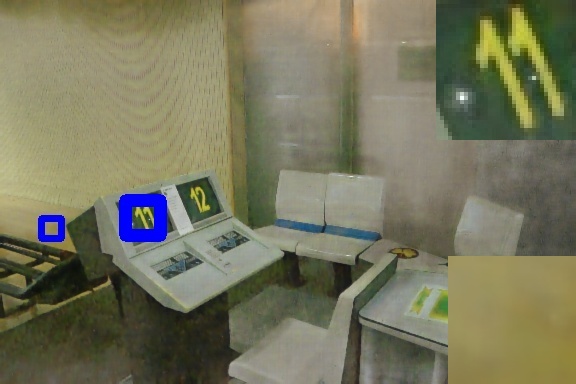} & 
     \includegraphics[width=0.16 \linewidth]{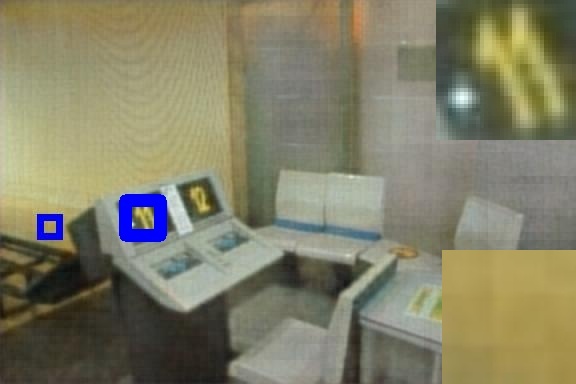}   & \includegraphics[width=0.16 \linewidth]{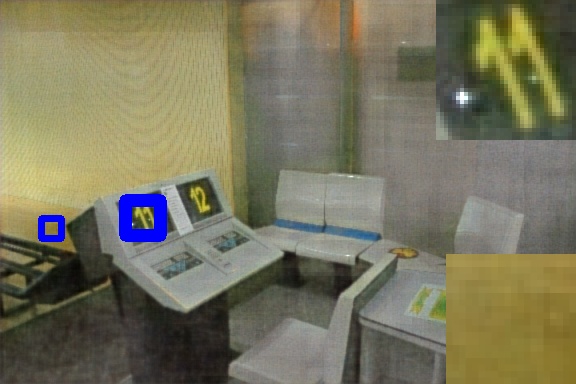} &  
     \includegraphics[width=0.16 \linewidth]{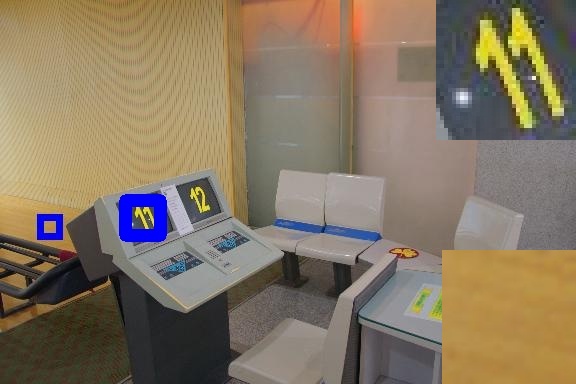}\\
     13.49/0.39 & 21.02/0.73 & 23.51/\textbf{0.77} & \textbf{24.31}/0.72 & \textcolor{black}{24.30}/\textcolor{black}{0.75}
    \end{tabular}
    \caption{Visual results on the LOL dataset corresponding to Table 2 in main paper. LLPackNet-8$\times$ is pretty fast with good color restoration but exhibits slight blurriness. This is rectified by LLPackNet-4$\times$ which chooses a lower downsampling factor befitting the low-resolution.}
\label{fig:LOL}
\end{figure*}

Fig. \ref{fig:LOL} shows qualitative results for the LOL dataset corresponding to Table 2 in the main paper.

\subsection{Worked out example of Pack/UnPack operation}

 Pack/UnPack operators perform intermixing of pixels for better color correlation. This intermixing is shown in Fig. 2 of the main paper. To further facilitate how the Pack$2 \times$/UnPack$2 \times$ do the shuffling in LR we display a worked-out example below. Consider an input tensor of $2 \times 2$ spatial resolution with $12$ channels as shown below.\\ ~ \\
{ \centering
    \begin{tabular}{c||cc||cc||cc||c||cc}
       \hline \textbf{Channel Count} & \multicolumn{2}{c||}{$1^{st}$ Channel } &  \multicolumn{2}{c||}{$2^{nd}$ Channel } &  \multicolumn{2}{c||}{$3^{rd}$ Channel }&  \multicolumn{1}{c||}{$\mathbf{\cdot \cdot \cdot}$ } &  \multicolumn{2}{c}{$12^{th}$ Channel } \\
       \hline \hline
       \textbf{Channel} & 1 & 2 & 5 & 6 & 9 & 10 & $\mathbf{\cdot \cdot \cdot}$ & 45 & 46 \\
       \textbf{Values} & 3 & 4 & 7 & 8 & 11 & 12 & $\mathbf{\cdot \cdot \cdot}$ & 47 & 48 \\ \hline
    \end{tabular}}
~ \\ ~ \\ Then, applying the UnPack $2\times$ operation we get a tensor of $4\times4$ spatial resolution with $3$ channels as shown below.
         
         
         
         
         
         
        
        
         
        
        

\begin{verbatim}
Red Channel or the first channel
          [ 1, 13,  2, 14]
          [25, 37, 26, 38]
          [ 3, 15,  4, 16]
          [27, 39, 28, 40],

Green Channel or the second channel
          [ 5, 17,  6, 18]
          [29, 41, 30, 42]
          [ 7, 19,  8, 20]
          [31, 43, 32, 44],
          
Blue Channel or the third channel
          [ 9, 21, 10, 22]
          [33, 45, 34, 46]
          [11, 23, 12, 24]
          [35, 47, 36, 48]
          
\end{verbatim}

\begin{table}[th!]
\centering
\scriptsize
\setlength\tabcolsep{1pt}
\begin{tabular}{c|cccc|cccc}
\hline
\hline
\textbf{H$\times$W; Channels} & \multicolumn{4}{c|}{\textbf{Execution Time in Seconds}} & \multicolumn{4}{c}{\textbf{Number of Learnable Parameters}} \\
& &
\multicolumn{1}{c}{\textbf{TransposeConv2D}} & \multicolumn{1}{c}{\textbf{UnPack}} & \multicolumn{1}{c|}{\textbf{Interpolation}} &  & \multicolumn{1}{c}{\textbf{TransposeConv2D}} & \multicolumn{1}{c}{\textbf{UnPack}} & \multicolumn{1}{c}{\textbf{Interpolation}} \\
\hline
\hline
$1024 \times 1024$; 32 -> $2048 \times 2048$; 8 &  & 0.18 & 0.05 & 0.13 &  & 1032 & $-$  & $-$ \\
$256 \times 256$; 128 -> $512 \times 512$; 32 &  & 0.04 & 0.01 & 0.04 &  & 16416 & $-$  & $-$ \\
$32 \times 32$; 512 -> $64 \times 64$; 128 &  & 0.0025 & 0.0006 & 0.0025 &  & 262272 & $-$  & $-$ \\ \hline 
\end{tabular}
\caption{We compare the execution time and learnable parameters required by Transposed Convolution (TransposeConv2D), Interpolation \cite{odena2016deconvolution} and the novel UnPack operation to perform upsampling by a factor of $2$. We use feature maps of different spatial resolutions and channel dimensions. Akin to modern deep methods we use fewer channels for large kernels and more channels for smaller kernels. To report these numbers we use the PyTorch framework on Intel Xeon E5-1620V4 @ 3.50 GHz CPU. The proposed UnPack is 3--4$\times$ faster than the popular techniques such as transposed convolution used by Chen \etal. }
\label{tab:execution_time}
\end{table}
\vspace{-0.5cm}
\subsection{Comparing Pack/UnPack with other popular down/up sampling operations}

\textbf{Downsampling: }Max-pooling is the most popular technique for downsampling feature maps. This has been used in many deep learning methods, including Chen \etal's network. But for a large downsampling it will cause huge loss of information. For example, when doing a 8$\times$ downsampling, max-pooling will choose only a single element from an $8\times8$ block.

Another popular downsampling technique is strided convolution, usually done with small kernels such as $3\times3$ or $5\times5$. But, for a large downsampling factor, say 8, a stride of 8 is required. However, with such small kernels it would lead to loss of information. To alleviate these issues, we used the novel Pack operation for downsampling feature maps without loss of information.

\textbf{Upsampling:} We have already shown the effectiveness of UnPack operation over the PixelShuffle operation in the main paper. Here, we compare with two other popular approaches -- Transposed convolution as used by Chen \etal and interpolation suggested by Odena \etal \cite{odena2016deconvolution}. The transposed convolution is very slow as compared to the UnPack operation because it has to iterate the convolution kernel over the entire feature map. Moreover it increases the parameter count of the network. On the other hand, the interpolation technique suggested by Odena \etal has no learnable parameters but is still a slower operation. This can be seen in Table \ref{tab:execution_time}.

\bibliography{egbib}
\end{document}